\documentclass[table]{ai2style/ai2}

\usepackage[utf8]{inputenc}
\usepackage{amsmath,amssymb,amsfonts}
\usepackage{url}
\usepackage{tabularx}
\usepackage{enumitem}
\usepackage{xspace}
\usepackage{bold-extra}
\usepackage{fdsymbol}   %

\definecolor{linkcolor}{RGB}{0, 0, 128}
\hypersetup{colorlinks=true, citecolor=linkcolor, linkcolor=linkcolor, urlcolor=linkcolor}

\setlist[itemize]{leftmargin=*,itemsep=0em,parsep=0.3em,topsep=0.3em}
\DeclareUnicodeCharacter{2212}{\ensuremath{-}}

\makeatletter
\providecommand\@LN[2]{}
\providecommand\@LN@col[1]{}
\makeatother

\newcolumntype{L}[1]{>{\raggedright\let\newline\\\arraybackslash\hspace{0pt}}m{#1}}
\newcolumntype{C}[1]{>{\centering\let\newline\\\arraybackslash\hspace{0pt}}m{#1}}
\newcolumntype{R}[1]{>{\raggedleft\let\newline\\\arraybackslash\hspace{0pt}}m{#1}}
\newcolumntype{P}[1]{>{\centering\let\newline\\\arraybackslash\columncolor{ai2lightpink}}m{#1}}

\newcommand{\github}{\raisebox{-1.5pt}{\includegraphics[height=1.05em]{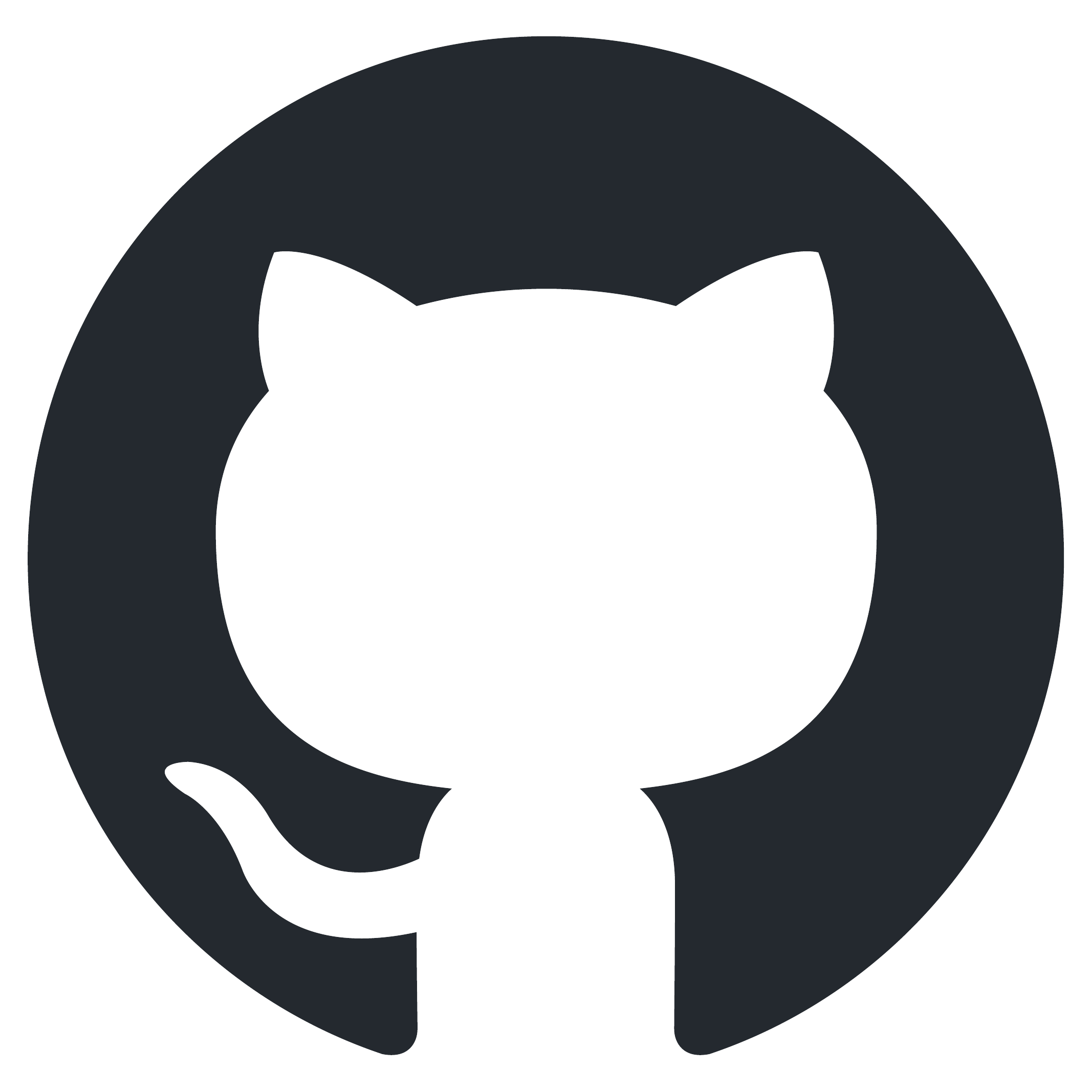}}\xspace}

\newcommand{\supp}{Appendix}               %
\newcommand{\tabfitwidth}{\width}          %
\newcommand{\citesupp}{\cite}              %
\newcommand{\suppreflist}{0}               %

\crefname{table}{Table}{Tables}\Crefname{table}{Table}{Tables}
\crefname{figure}{Figure}{Figures}\Crefname{figure}{Figure}{Figures}
\crefname{section}{Section}{Sections}\Crefname{section}{Section}{Sections}
\crefname{subsection}{Section}{Sections}\Crefname{subsection}{Section}{Sections}
\crefname{subsubsection}{Section}{Sections}\Crefname{subsubsection}{Section}{Sections}
\crefname{appendix}{Appendix}{Appendices}\Crefname{appendix}{Appendix}{Appendices}
\crefname{subappendix}{Appendix}{Appendices}\Crefname{subappendix}{Appendix}{Appendices}
\crefname{subsubappendix}{Appendix}{Appendices}\Crefname{subsubappendix}{Appendix}{Appendices}
\crefname{equation}{Eq.}{Eqs.}\Crefname{equation}{Equation}{Equations}

\newcommand{\aitwoplainbf}{\renewcommand{\textbf}[1]{{\bfseries ##1}}}
\AtBeginEnvironment{tabular}{\aitwoplainbf\aitwotabfont}
\AtBeginEnvironment{tabular*}{\aitwoplainbf\aitwotabfont}
\AtBeginEnvironment{tabularx}{\aitwoplainbf\aitwotabfont}

\makeatletter
\newcommand{\aitwotabfont}{}
\newcommand{\aitwotabshrink}[1]{\def\aitwotabfont{\fontsize{\strip@pt\dimexpr\f@size pt*#1/100\relax}%
  {\strip@pt\dimexpr\f@baselineskip*#1/100\relax}\selectfont\let\aitwotabfont\relax}}
\newcommand{\aitwotable}[1]{\csname aitwotab@#1\endcsname}
\@namedef{aitwotab@tab:extreme}{\aitwotabshrink{88}}           %
\@namedef{aitwotab@tab:station_networks}{\aitwotabshrink{90}}  %
\makeatother

\makeatletter
\NewCommandCopy{\aitwoincludegraphics}{\includegraphics}
\RenewDocumentCommand{\includegraphics}{s O{} m}{%
  \ifcsname aitwofig@#3\endcsname
    \scalebox{\csname aitwofig@#3\endcsname}{\IfBooleanTF{#1}{\aitwoincludegraphics*[#2]{#3}}{\aitwoincludegraphics[#2]{#3}}}%
  \else
    \IfBooleanTF{#1}{\aitwoincludegraphics*[#2]{#3}}{\aitwoincludegraphics[#2]{#3}}%
  \fi}
\newcommand{\aitwofigscale}[2]{\@namedef{aitwofig@#1}{#2}}
\makeatother
\aitwofigscale{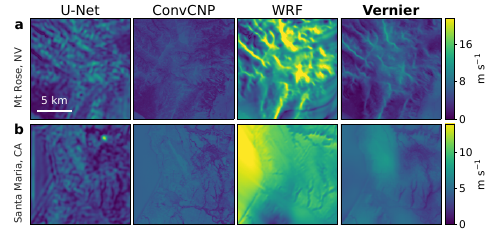}{0.75}         %
\aitwofigscale{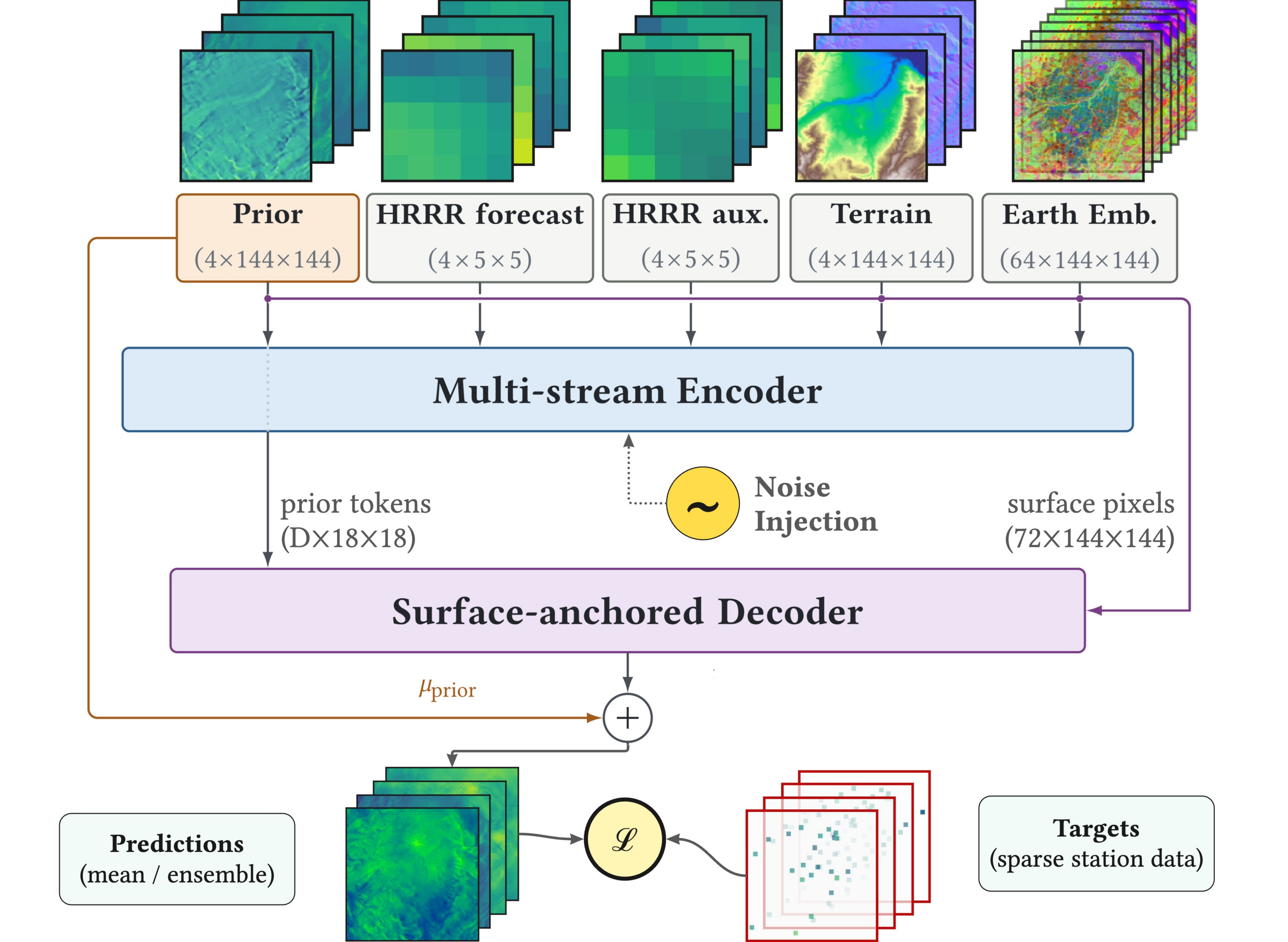}{0.75}                       %
\aitwofigscale{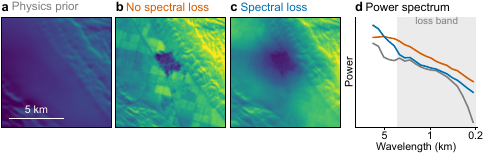}{0.8}           %
\aitwofigscale{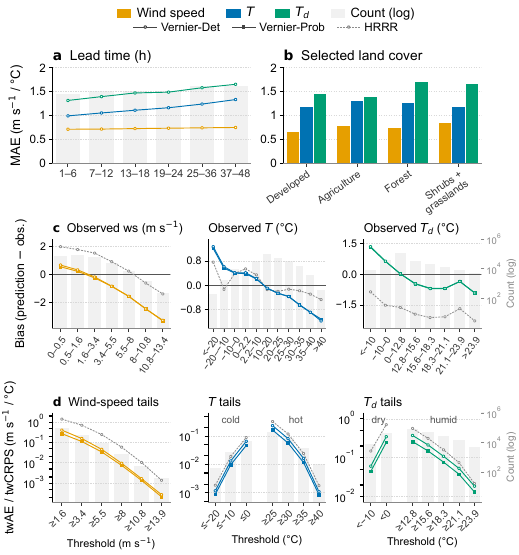}{0.8}          %
\aitwofigscale{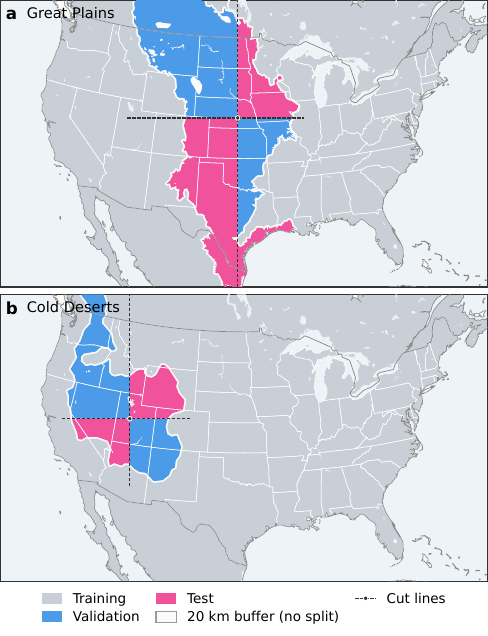}{0.65}      %
\aitwofigscale{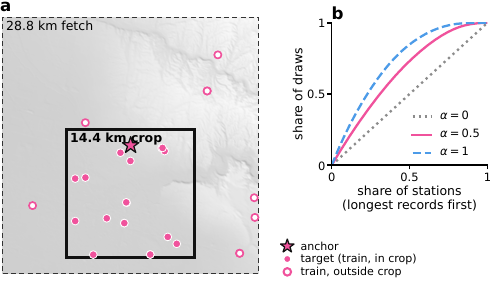}{0.75}      %
\aitwofigscale{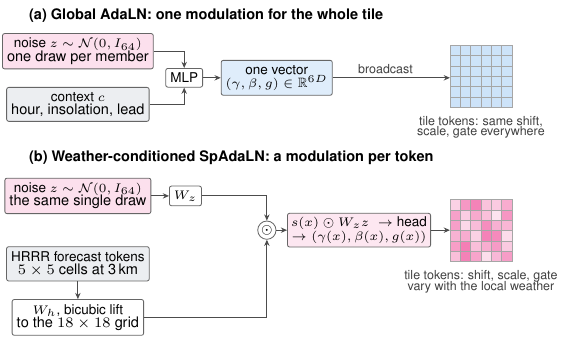}{0.75}      %
\aitwofigscale{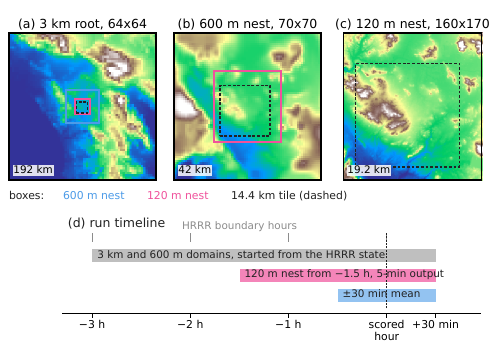}{0.85}          %
\aitwofigscale{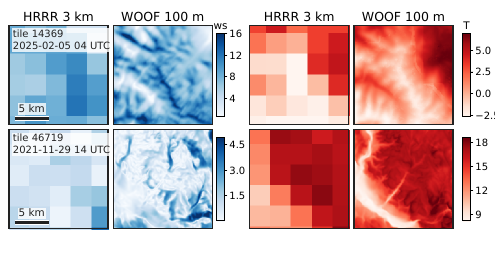}{0.75}         %
\aitwofigscale{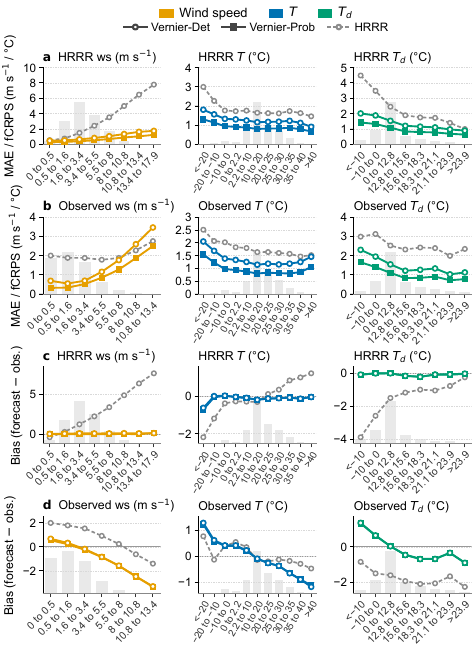}{0.65}  %

\expandafter\renewcommand\expandafter\section\expandafter{\expandafter\FloatBarrier\section}

\newcommand{\papertitle}{Vernier: Learning to Resolve Subkilometer Surface Weather from Observations}

\providecommand{\supp}{Supp.}                %
\providecommand{\tabfitwidth}{\columnwidth}  %
\providecommand{\aitwotable}[1]{}            %

\def\colordefault{0a3235}
\def\colorone{f0529c}
\def\colortwo{b11be8}
\def\colorthree{105257}
\def\colorfour{0fcb8c}
\def\colorfive{12cce5}
\def\colorsix{f65834}

\title{\huge\papertitle}
\DeclareRobustCommand{\aff}[1]{\raisebox{.28em}{\hspace{.02em}\scalebox{0.7}{\textbf{#1}}}}
\authorOne[\aff{1,2}]{Ando Shah*}
\authorOne[\aff{1}]{Yawen Zhang*}
\authorOne[\aff{1}]{Favyen Bastani}
\authorOne[\aff{1}]{Hadrien Sablon}
\authorOne[\aff{1}]{Piper Wolters}
\authorOne[\aff{1}]{Joseph Redmon}
\authorOne[\aff{3}]{Philipp Hess}
\authorOne[\aff{1}]{Troy Arcomano}
\authorOne[\aff{2}]{John Chuang}
\authorOne[\aff{4}]{Sherrie Wang}
\authorOne[\aff{1}]{Oliver Watt-Meyer}
\authorOne[\aff{1}]{Gabriel Tseng}
\renewcommand\authorlistTwo{\vspace{-0.3cm}}
\renewcommand\authorlistThree{\vspace{-0.3cm}}
\renewcommand\authorlistFour{\vspace{-0.3cm}}
\renewcommand\authorlistFive{\vspace{-0.3cm}}
\renewcommand\affiliation[2][]{\addtolist[#1]{#2}{\affiliationlist}{\affiliationformat}{\hskip 0pt plus 1fil\penalty0\hskip 2em plus -1fil\relax}}
\affiliation[\aff{1}]{\mbox{Allen Institute for Artificial Intelligence}}
\affiliation[\aff{2}]{\mbox{University of California, Berkeley}}
\affiliation[\aff{3}]{\mbox{Potsdam Institute for Climate Impact Research}}
\affiliation[\aff{4}]{\mbox{Massachusetts Institute of Technology}}
\abstract{\vspace{-3mm}\small\aitwoplainbf Wildfire response, agriculture, and urban heat planning need surface weather forecasts at finer scales than operational models provide, yet at these scales observations exist only at sparse weather stations.
We present \textbf{Vernier}, which bias-corrects and downscales 3\,km HRRR forecasts of 2\,m temperature, 2\,m dew point and 10\,m wind to a 100\,m grid, supervised only by these stations.
Vernier learns a residual correction to a physics-based prior that adjusts HRRR for local terrain.
A multi-stream encoder combines the forecast, prior and terrain with Earth-observation embeddings, and a surface-anchored decoder adds fine-scale detail. After deterministic training, we finetune Vernier with latent noise and dropout to generate ensembles.
At inference, \textbf{Searchlight} uses neighborhood attention to coordinate each member’s stochastic variations beyond the training tile, producing seamless fields in one pass per member without retraining.
At unseen stations in the held-out year 2026, Vernier's ensembles achieve lower continuous ranked probability score (CRPS) than all compared probabilistic baselines, and their members better reproduce observed differences between nearby stations.  
Unlike the compared 100\,m ensemble baselines, whose members are noisy or speckled, Vernier's members stay coherent.
Its ensemble mean reduces HRRR's mean absolute error (MAE) by 59\% for wind speed, 26\% for temperature and 38\% for dew point.
Coarsening Vernier’s forecasts to the HRRR grid barely changes MAE at stations, but erases much of the wind and temperature structure that lets high-resolution downscaling recover observed local contrasts.
Trained on all eligible historical stations, Vernier outperforms all compared operational models in wind-speed and temperature error (MAE and CRPS), while WeatherNext~3 retains a small advantage for dew point.
Vernier improves on station-specific calibration in probabilistic forecasts and better captures local weather contrasts between nearby stations.

\vspace{-4mm}

\par}
\newcommand{\aitwoteaser}{%
  \vspace{-0.5cm}%
  {\centering
  \noindent\includegraphics[width=0.98\textwidth]{figures/fig1.pdf}\par
  \captionsetup{font=footnotesize}
  \captionof{figure}{Vernier bias-corrects and downscales High-Resolution Rapid Refresh (HRRR) forecasts of surface wind, temperature and dew point to 100 m resolution for various applications. Searchlight generates the regional fields (`Vernier') by extending inference beyond the training tiles.}\label{fig:overview}\par}
  \vspace{0.15cm}}
\makeatletter %
\patchcmd{\mymaketitle}{\begin{tcolorbox}}{\aitwoteaser\begin{tcolorbox}}{}%
  {\PackageWarning{ai2_paper}{overview figure not placed: the class's \string\mymaketitle\space changed}}
\patchcmd{\mymaketitle}{\titlelist\par}{{\raggedright\titlelist\par}}{}%
  {\PackageWarning{ai2_paper}{title not set ragged right: the class's \string\mymaketitle\space changed}}
\makeatother
\begin{document}
\definecolor{colordefault}{HTML}{\colordefault}
\definecolor{colorone}{HTML}{\colorone}
\definecolor{colortwo}{HTML}{\colortwo}
\definecolor{colorthree}{HTML}{\colorthree}
\definecolor{colorfour}{HTML}{\colorfour}
\definecolor{colorfive}{HTML}{\colorfive}
\definecolor{colorsix}{HTML}{\colorsix}
\maketitle
\fancypagestyle{aitwofirstpage}{\fancyhf{}\renewcommand{\headrulewidth}{0pt}\renewcommand{\footrulewidth}{0pt}%
  \fancyfoot[C]{\thepage}\fancyfoot[L]{\footnotesize\textsuperscript{*}Denotes equal contribution}}
\thispagestyle{aitwofirstpage}

\section{Introduction}\label{sec:intro}

Many decisions depend on weather at the scale of a hillside or a city block: where a wildfire will spread~\cite{sharples2009overview,forthofer2014comparison}, which fields will frost~\cite{snyder2005frost}, and how hot a neighborhood will get~\cite{hoffman2020effects}. Yet operational forecasts are produced on grids of several kilometers. ECMWF's IFS runs at 9\,km globally~\cite{ecmwf2024ifs}, and NOAA's HRRR, among the finest operational models, runs at 3\,km over the contiguous United States and adjacent parts of Canada and Mexico~\cite{dowell2022hrrr}. AI weather models now rival numerical models on many benchmarks but still run at coarse resolution~\cite{lam2023graphcast,bi2023pangu,price2025gencast,alet2025fgn}. Within a single 3\,km grid cell, cold air pools in valleys, winds accelerate over ridges, and cities stay warmer than their surroundings~\cite{gustavsson1998,oke1982}. Capturing these local variations requires downscaling forecasts to sub-kilometer resolution, where terrain and land cover drive much of the variability~\cite{giezendanner2026partial,jin2025}.

Existing approaches fall into two groups. Physics-based methods include running a regional model such as WRF~\cite{skamarock2008wrf} inside the coarse forecast, which is very costly and physically difficult below 1\,km. WindNinja offers a cheaper alternative by adjusting winds to the terrain~\cite{forthofer2014comparison,windninja2025}, but it can inherit or amplify forecast bias~\cite{seto2025evaluation}. ML downscalers learn to map coarse fields to finer ones~\cite{vandal2017deepsd,stengel2020,miralles2022,mardani2025corrdiff,schillinger2026enscale,perkins2025hiro,pyrina2026joint,springenberg2026diffscale}. However, they require dense high-resolution targets, which are not observed at sub-kilometer scales.

\begin{figure}[h]
  \centering
  \includegraphics[width=\columnwidth]{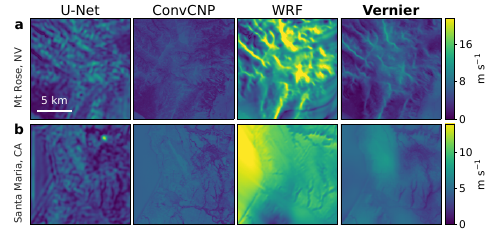}
  \captionsetup{font=footnotesize}
  \caption{\textbf{Downscaled fields can differ widely between stations.} 100\,m wind speed from two ML baselines (U-Net, ConvCNP), a 3-D physics-based model (WRF) and Vernier. The U-Net field shows speckle and the ConvCNP field shows sharp imprints of surface features, while Vernier's field resembles the terrain-driven structure of WRF.}
  \label{fig:station_fields}
\end{figure}

Weather stations measure surface weather directly and continuously: MADIS~\cite{miller2005madis} and HadISD~\cite{dunn2019hadisd} together report hourly from tens of thousands of stations across North America. Recent work has explored downscaling using such observations, judging models by their error at stations~\cite{scholz2023sim2real,vaughan2022convcnp,sousa2026earth}. But stations are sparse, noisy, and clustered in populated areas and at airports, and a model can match every station while producing unrealistic fields between them (\cref{fig:station_fields}). The challenge is to fit the stations while keeping the field realistic where there are no stations. What happens between stations depends strongly on terrain and land cover, which, unlike weather, are mapped everywhere at high resolution. This raises our central question: \textbf{how can we downscale coarse forecasts into realistic sub-kilometer fields using sparse station observations as supervision?}

We introduce \textbf{Vernier}, which learns surface weather at 100\,m from 3\,km HRRR forecasts, trained only on station observations. Rather than using static land-surface datasets like land cover or roughness, Vernier uses Earth-observation embeddings as dynamic surface descriptors. We test generalization to unseen stations in the held-out year 2026, and separately compare models trained on all eligible historical stations with operational forecasts. Our contributions are:

(1) \textbf{Physics-guided residual learning from sparse stations.} Our physics prior captures how terrain shapes local wind, temperature and dew point, giving physically plausible structure where no stations constrain the model. Vernier learns a residual correction to the physics prior from station observations.

(2) \textbf{Surface-anchored deterministic and probabilistic prediction.} A multi-stream encoder jointly attends over the coarse forecast, the prior and surface inputs at their native resolutions. A surface-anchored decoder then adds 100\,m detail by combining this context with the land-surface features at each pixel. The same architecture serves both deterministic and probabilistic training, and the probabilistic model is finetuned from the deterministic one with injected noise and dropout to generate ensembles. We also propose a spectral loss that mitigates land-cover imprinting in the wind field.

(3) \textbf{Scaling single-step ensembles beyond training tiles.}
We propose \textbf{Searchlight}, which uses neighborhood attention
at inference to coordinate each ensemble member spatially across
a larger domain. This optional mode generates continuous regional
weather fields in one pass per member without retraining, with
modestly reduced ensemble spread.

\section{Related Work}
\label{sec:related}

\paragraph{Bias correction and downscaling.}
Most existing methods address bias and resolution separately. Statistical post-processing corrects forecast biases at weather stations~\cite{glahn1972mos,gneiting2005emos}. Physical downscalers like WindNinja add terrain-driven detail to coarse winds~\cite{forthofer2014comparison,windninja2025}, and ML downscalers learn fine-scale fields with convolutional networks~\cite{vandal2017deepsd}, adversarial training~\cite{stengel2020,harris2022} or conditional diffusion~\cite{mardani2025corrdiff,perkins2025hiro}. Recent ML methods address both~\cite{pyrina2026joint,springenberg2026diffscale}, but they learn from dense high-resolution simulations, inheriting their biases and resolution, and such simulations rarely exist below 1\,km. Vernier addresses both, but learns from sparse stations alone.

\vspace{-4mm}

\paragraph{Earth embeddings for downscaling.}
Earth foundation models compress each year of satellite observations into an annual embedding of the land surface~\cite{brown2025alphaearth,tseng2026olmoearth,feng2026tessera}, a dynamic alternative to the static land cover and soil maps that physical models like WRF require~\cite{skamarock2008wrf}. Recent work uses these embeddings as surface descriptors for station-based weather downscaling~\cite{sousa2026earth,giezendanner2026partial} but trains and evaluates only at stations, leaving the fields between them unconstrained. Without such constraints, a model can imprint land-cover patterns onto the weather field (\cref{fig:spectral_loss}). Vernier constrains the fields between stations using a physics prior and a spectral loss, and measures their structure directly with station-pair correlations.

\vspace{-4mm}
\paragraph{Efficient generative ensembles.}
Diffusion models~\cite{price2025gencast,mardani2025corrdiff,springenberg2026diffscale} need
many network evaluations per member, whereas single-pass models
inject latent noise~\cite{alet2025fgn}, apply Monte Carlo
dropout~\cite{cachay2026ucast}, or combine
both~\cite{rasp2026weathernext}, and can be trained with energy
scores~\cite{schillinger2026enscale}.
On large domains, independently generated tiles can leave seams;
cBottle coordinates overlapping patches during diffusion to reduce
these artifacts~\cite{brenowitz2025cbottle}.
For single-step weather ensembles, Searchlight adapts neighborhood
attention~\cite{hassani2023neighborhood} at inference to coordinate
each member through overlapping, tile-sized neighborhoods,
generating regional fields in one pass without retraining.

\section{Method}\label{sec:method}

\subsection{Problem Formulation}
\label{sec:problem}

We downscale weather forecasts to 100\,m, conditioned on surface features and supervised by weather stations.

\vspace{-3mm}
\paragraph{Inputs and outputs.}
For a tile at a given time, the model receives three inputs:
(1) the coarse forecast $X$, the target variables together with auxiliary weather fields, 
(2) the surface inputs $S$ on the 100\,m grid, including terrain and Earth embeddings, and
(3) the context $c$: valid time, forecast lead time and location.
It predicts the 2\,m temperature $T$, the 2\,m dew point $T_d$, and the eastward and northward components $u$ and $v$ of the 10\,m wind at every 100\,m pixel $x$. Wind speed $\sqrt{u^2+v^2}$ and direction are computed from $(u,v)$. 

\vspace{-3mm}
\paragraph{Deterministic and probabilistic prediction.}
For a deterministic model, it returns a single estimate of the field,
\begin{equation}
    \widehat{Y}=f_\theta(X,S,c)
    \label{eq:det}
    \vspace{-2mm}
\end{equation}
A probabilistic model returns an ensemble of $K$ members,
\begin{equation}
    \widehat{Y}^{(k)}=f_\theta\big(X,S,c,\epsilon^{(k)}\big)\qquad k=1,\dots,K
    \vspace{-2mm}
    \label{eq:prob}
\end{equation}
where $\epsilon^{(k)}$ is a random draw that differs between members (\cref{sec:stochastic_prediction}). Each member is a complete 100\,m field, and the spread of the members expresses the forecast uncertainty. The ensemble mean serves as a point estimate.

\vspace{-3mm}
\paragraph{Supervision from weather station observations.}
Supervision comes from the weather stations inside a tile: station $j$ at pixel $x_j$ reports observations $y_j$ for a subset $V_j\subseteq\{T,T_d,u,v\}$ of the variables. The model is compared with the observations only at those pixels and for those variables,
\begin{equation}
    \mathcal{L}_{\mathrm{station}}=\sum_{j}\sum_{q\in V_j}\ell_q\big(\widehat{Y}_q(x_j),\,y_{j,q}\big)
    \vspace{-2mm}
    \label{eq:station_loss}
\end{equation}
where $\ell_q$ is a per-variable loss, defined for the deterministic and probabilistic models in \cref{sec:training}.

\subsection{Physics-Guided Residual Learning}
\label{sec:physical_prior}

Rather than predicting the 100\,m field from scratch, Vernier starts from a physics-based 100\,m estimate of the same field and learns to correct it~\cite{mardani2025corrdiff}. A fixed physical downscaler $\mathcal{P}$ turns the coarse forecast and the terrain $S_t$, with the solar geometry from the context $c$, into a prior field $P=\mathcal{P}(X,S_t,c)$, and the network predicts the residual:
\begin{equation}
    \widehat{Y}=P+\Delta_\theta(X,S,c,P)
    \vspace{-1mm}
    \label{eq:residual}
\end{equation}
Each probabilistic member adds a correction generated from its random input $\epsilon^{(k)}$ to the same prior. The prior adjusts the coarse forecast to the 100\,m terrain, e.g., flow around terrain and cooling with elevation. From station observations, the network learns a correction for what the prior misses: forecast biases, prior modeling errors, and effects of the land surface. It uses embeddings derived from satellite observations, which capture surface features such as land cover and roughness that affect all three variables. Unlike stations, the embeddings are available at every pixel, so the correction learned at stations can be applied between them.

\vspace{-3mm}
\paragraph{Physics prior.}
For wind $u$ and $v$, we follow WindNinja's mass-consistent solver~\cite{windninja2025}, reimplemented in PyTorch: the coarse wind is interpolated bicubically to 100\,m and projected onto a divergence-free flow over a terrain-following mesh, and a fitted factor corrects the forecast's wind-speed bias. A stability parameter, predicted per pixel by a small network from the forecast, terrain slope and solar zenith angle, sets how much air flows over rather than around terrain. For temperature and dew point, a lapse-rate correction adjusts the forecast for the height difference between the 100\,m terrain and its local mean over 3\,km, with a stability adjustment, a bounded moisture increase in valleys, and a constraint $T_d\le T$. The prior's parameters are fitted to training-station observations and then fixed, and the prior is fast enough to compute on the GPU for every training sample (\supp~\ref{supp:prior}).

\begin{figure}[t]
  \centering
  \includegraphics[width=\columnwidth]{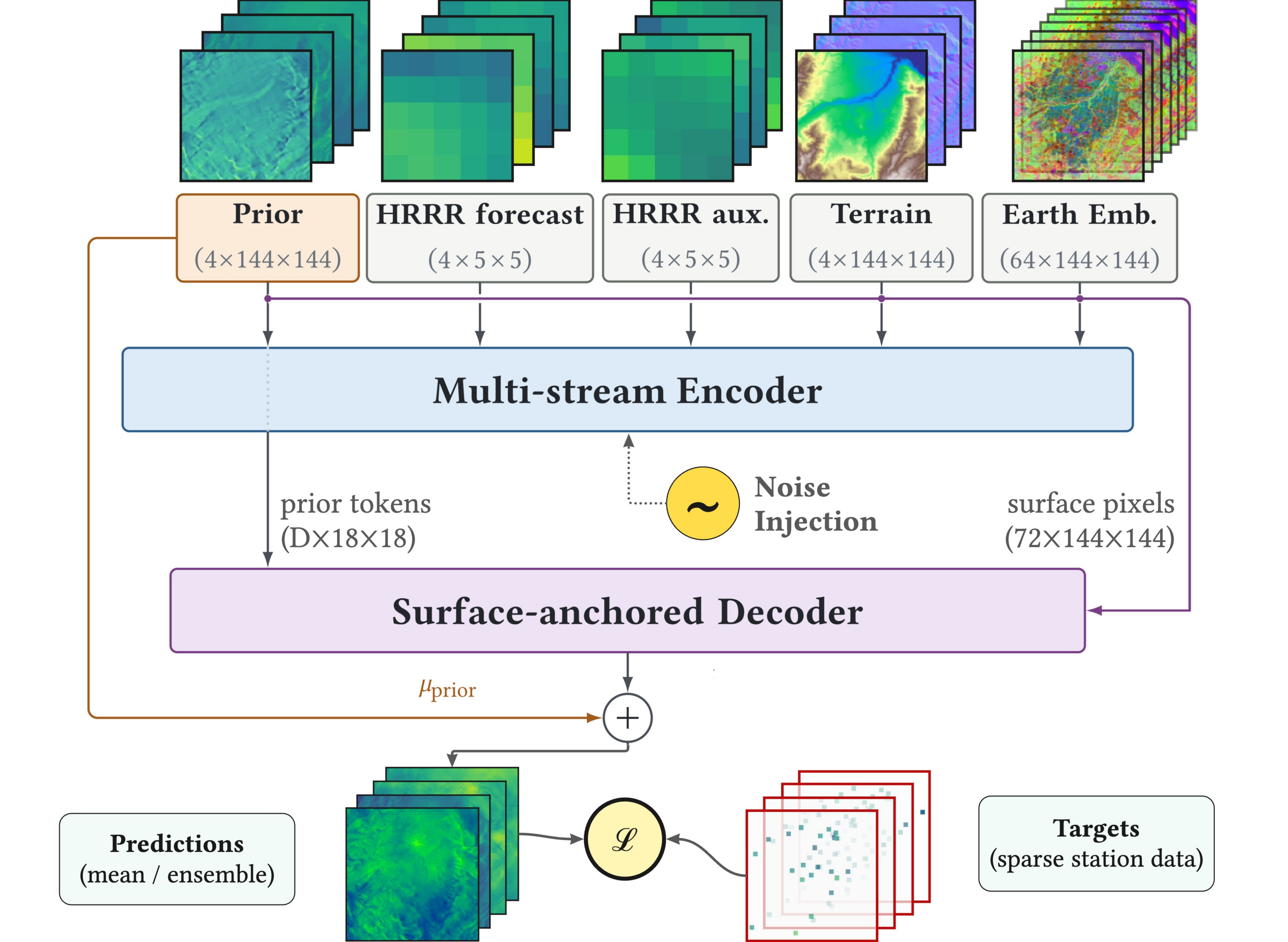}
  \captionsetup{font=footnotesize}
  \caption{\textbf{Vernier model architecture.} Each input (coarse forecast, physics prior, terrain, Earth embeddings) is split into tokens at its own resolution. The multi-stream encoder attends jointly over all tokens, and the surface-anchored decoder combines its output with per-pixel surface features to correct the prior at 100\,m. The model is trained on sparse station observations only. The probabilistic version adds noise and dropout to produce  ensemble members.}
  \label{fig:architecture}
\end{figure}

\subsection{Model Architecture}
\label{sec:architecture}

Vernier has a multi-stream encoder and a surface-anchored decoder (\cref{fig:architecture}). The encoder jointly attends over all input tokens across the tile, and the decoder anchors the output to 100\,m surface inputs at every pixel.

\paragraph{Multi-stream encoder.}\label{sec:encoder}
Inspired by MMDiT~\cite{esser2024mmdit}, the encoder is a multi-stream Transformer. The weather forecast, terrain, embeddings and prior each enter as their own token stream, and joint attention lets every token attend to every other. Each data stream keeps its own resolution and number of tokens: the high-resolution inputs are split into patches, and each coarse forecast cell becomes one token. We use 2D rotary position embeddings (RoPE)~\cite{heo2024ropevit} based on each token's location on the ground, so tokens from different resolutions share one coordinate frame. The context $c$ conditions each block through adaptive layer normalization (AdaLN). \supp~\ref{supp:encoder} gives the full configuration.

\paragraph{Surface-anchored decoder.}\label{sec:decoder}
Each token covers $p\times p$ pixels at 100\,m ($p$ is the patch size), which is too coarse to carry all details, so the decoder re-injects that detail from the 100\,m inputs. Small pointwise networks map each high-resolution input $g\in\{\text{terrain},\text{embedding},\text{prior}\}$ to features $\phi_g(x)$ at each pixel $x$. Linear heads on the encoder's output tokens predict coefficients $b$ and $W_g$, which are bicubically upsampled to 100\,m. The correction to the prior at pixel $x$ is
\begin{equation}
    \Delta(x)=b(x)+\sum_{g}W_g(x)\,\phi_g(x)
    \vspace{-2mm}
    \label{eq:surface_decoder}
\end{equation}
The coefficients depend on the weather and set how strongly each feature applies; the features supply the 100\,m pattern. In the probabilistic model, all members share the same features and differ only in their coefficients. \supp~\ref{supp:decoder} gives further details.

\subsection{Learning from Sparse Observations}
\label{sec:training}

\paragraph{Deterministic training.}
All variables are trained jointly with a weighted sum of per-variable losses,
\begin{equation}
\ell_{\mathrm{det}}=\lambda_T\,\ell_T+\lambda_{T_d}\,\ell_{T_d}+\lambda_{\mathrm{wind}}\,\ell_{\mathrm{wind}}
    \vspace{-1mm}
    \label{eq:deterministic_loss}
\end{equation}
where weights $\lambda$ balance the three losses. Temperature and dew point use an L2 loss. For wind, the L2 loss on the $(u, v)$ components can be written as $(\widehat{s}-s)^2+2\,\widehat{s}\,s\,(1-\cos\Delta\vartheta)$, where $\widehat{s}$ and $s$ are the predicted and observed wind speeds and $\Delta\vartheta$ is the angle between the wind vectors. The direction term shrinks when the predicted speed drops, which pushes the model toward weak winds. We replace $\widehat{s}\,s$ with $s^2$, giving $\ell_{\mathrm{wind}}=(\widehat{s}-s)^2+2\,s^2(1-\cos\Delta\vartheta)$, so the direction penalty no longer depends on the prediction. Without this change, the model is negatively biased in wind speed and noticeably underestimates winds (\cref{tab:ablation}). \supp~\ref{supp:deterministic} gives details.

\paragraph{Ensemble generation.}\label{sec:stochastic_prediction}
Each ensemble member draws a 64-dimensional Gaussian noise vector, which enters the network through AdaLN in one of two ways: (1) \emph{global modulation}~\cite{alet2025fgn}, where the noise sets shifts, scales and gates that are the same everywhere in the tile, or (2) \emph{weather modulation}, where the noise is multiplied by local embeddings of the coarse forecast, so a single draw produces perturbations that vary with the local weather across the tile (shown in~\cref{fig:supp_adaln_vs_spadaln}). This lets the ensemble spread vary with the weather regime. Following WeatherNext~3~\cite{rasp2026weathernext}, dropout ($p=0.1$) adds a second source of randomness: in training, members share dropout masks but draw independent latents; at inference, both are sampled independently. We train one model with each modulation type, and Vernier-Prob combines 32 members from each (\supp~\ref{supp:gen-ens}).

\paragraph{Probabilistic training.}
The probabilistic model is trained with the continuous ranked probability score (CRPS), which compares the distribution of its $K$ members $a^{(1)},\dots,a^{(K)}$ with the observation $y$. We use its \emph{fair} version~\cite{ferro2014fair}, which corrects for the small number of members. For temperature, dew point and wind speed, the fCRPS is:
\begin{equation}
    C=\frac{1}{K}\sum_{k=1}^{K}\big|a^{(k)}-y\big|-\frac{1}{K(K-1)}\sum_{k<k'}\big|a^{(k)}-a^{(k')}\big|
    \vspace{-1mm}
    \label{eq:fair_score}
\end{equation}
The first term keeps members close to the observation, and the second penalizes them for all giving the same answer. For wind $(u,v)$ we use the fair energy score~\cite{gneiting2007strictly}, which has the same form with $|\cdot|$ replaced by the Euclidean distance. The probabilistic loss is the weighted sum of four losses,
\begin{equation}
    \ell_{\mathrm{prob}}=\lambda_T C_T+\lambda_{T_d}C_{T_d}+\lambda_{s}C_{s}+\lambda_{uv}E_{uv}.
    \label{eq:probabilistic_loss}
\end{equation}
Probabilistic training starts from the deterministic model for faster convergence and uses $K=2$ members per sample.

\begin{figure}[h]
  \centering
  \includegraphics[width=\columnwidth]{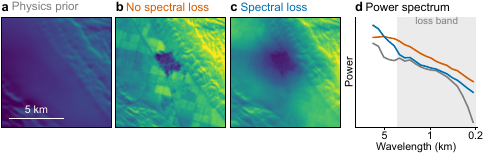}
  \captionsetup{font=footnotesize}
  \caption{\textbf{Effect of the spectral loss on generated 100\,m wind speed.} (a) Physics prior. (b) Without the spectral loss, field boundaries from the embedding imprint onto the wind. (c) With it, these boundaries are suppressed while the prior's terrain-driven structure is preserved. (d) Power spectra of a--c: in the shaded 200\,m--3.2\,km band, the model with the spectral loss stays near the prior's power, while the model without it exceeds the prior.}
  \label{fig:spectral_loss}
\end{figure}

\paragraph{Surface artifact mitigation.}\label{sec:imprint}
Land surface features carry real fine-scale signal, but away from stations the model can also copy land-cover texture that has no clear effect on the weather, e.g., field boundaries in the wind field (\cref{fig:spectral_loss}b). We add a one-sided spectral loss $\mathcal{L}_{\mathrm{spec}}$ that uses the wind prior as a reference: it penalizes predicted wind power that exceeds $\tau$ times the prior's in any fine-scale band $b\in\mathcal{B}$, so detail below this limit is kept. The training loss becomes $\mathcal{L}=\mathcal{L}_{\mathrm{station}}+\lambda_{\mathrm{spec}}\mathcal{L}_{\mathrm{spec}}$. The loss suppresses surface artifacts (\cref{fig:spectral_loss}) at a small cost in wind skill (\cref{tab:ablation}). \supp~\ref{supp:spectral} studies how $\tau$ and $\mathcal{B}$ affect this trade-off.

\subsection{Searchlight for Large-Domain Inference}
\label{sec:searchlight}
For domains larger than a training tile, Vernier runs inference either tile by tile or with \emph{Searchlight}, which produces one seamless field.
In Searchlight, each token attends to a tile-sized window centered on its location in every input stream, so the attention extent matches training while the whole domain is processed at once, without retraining.
The prior is assembled from overlapping tile-sized solves, keeping each solve's center crop without averaging, and each member draws its noise once over the whole domain.
Compared with tile-by-tile inference, station scores change little and the ensemble spread is modestly lower (\cref{tab:operational}). These differences stay roughly constant as the domain grows. Further details including implementation specifics and ablations are available in~\cref{supp:searchlight}.

\section{Experiments}
\label{sec:experiments}

\subsection{Datasets and Splits}
\label{sec:data}

We construct a dataset combining operational weather forecasts, weather station observations, and high-resolution surface features, which spans January 2017 to September 2026.

\begin{figure*}[tb]
  \centering
  \includegraphics[width=\textwidth]{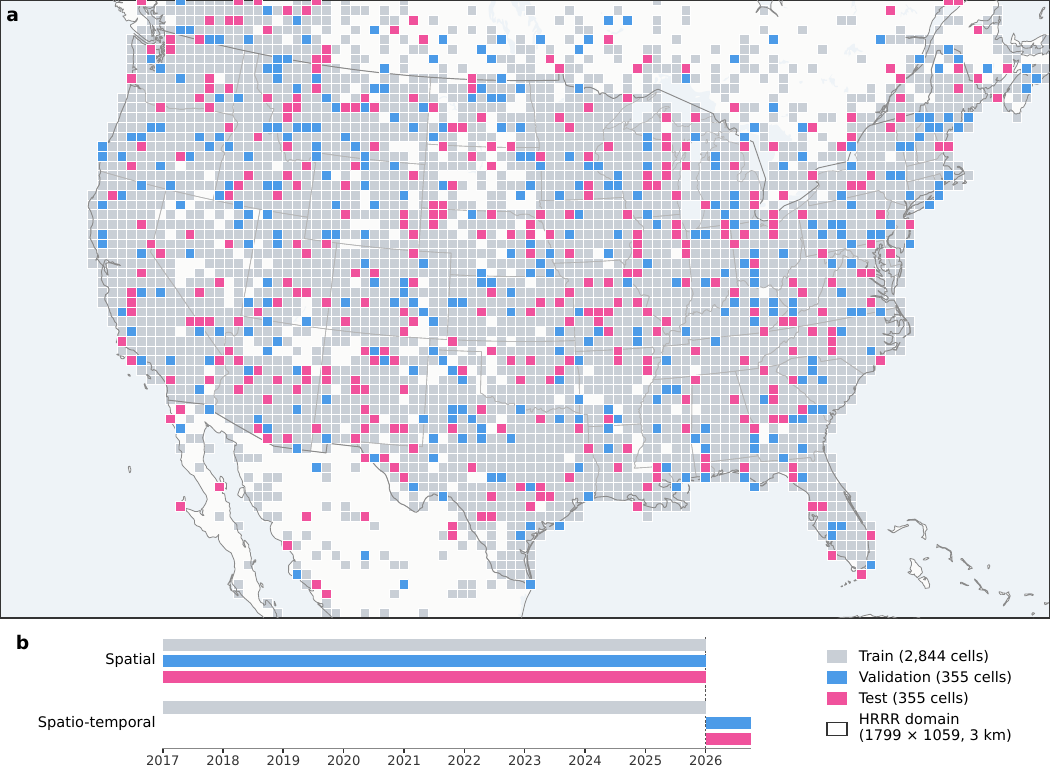}
  \caption{\textbf{HRRR domain and spatial split.} \textbf{a}, The 3,554 eligible $50\times50$\,km cells, assigned to training (gray, 2,844), validation (blue, 355) and test (pink, 355) with seed 9 and stratified by station availability, terrain relief and surface-embedding diversity. The map uses HRRR's Lambert conformal projection, and its frame is the native $1799\times1059$ grid at 3\,km. Blank areas are water or cells without valid surface embeddings or a station with sufficient wind coverage; eligible cells extend into southern Canada and northern Mexico. \textbf{b}, The same cells serve two holdouts. In the spatial holdout, all three partitions cover 2017--2025, so only the validation and test stations are unseen. In the spatio-temporal holdout, the validation and test cells are also held out in time and cover January--September 2026.}
  \label{fig:supp_split_map}
\end{figure*}

\paragraph{Operational weather forecasts.}
We use NOAA's High-Resolution Rapid Refresh (HRRR)~\cite{dowell2022hrrr}, which provides 3\,km forecasts at hourly leads up to 48\,h over CONUS and adjacent parts of Canada and Mexico (\cref{fig:supp_split_map}; \supp~\ref{supp:hrrr}). The primary HRRR inputs include 2\,m temperature ($T$), dew point ($T_d$), 10\,m winds ($u, v$), supplemented by cloud cover, skin temperature, surface pressure, and soil moisture. We use hourly HRRR analyses for station quality control.

\begin{figure*}[t]
  \centering
  \includegraphics[width=\textwidth]{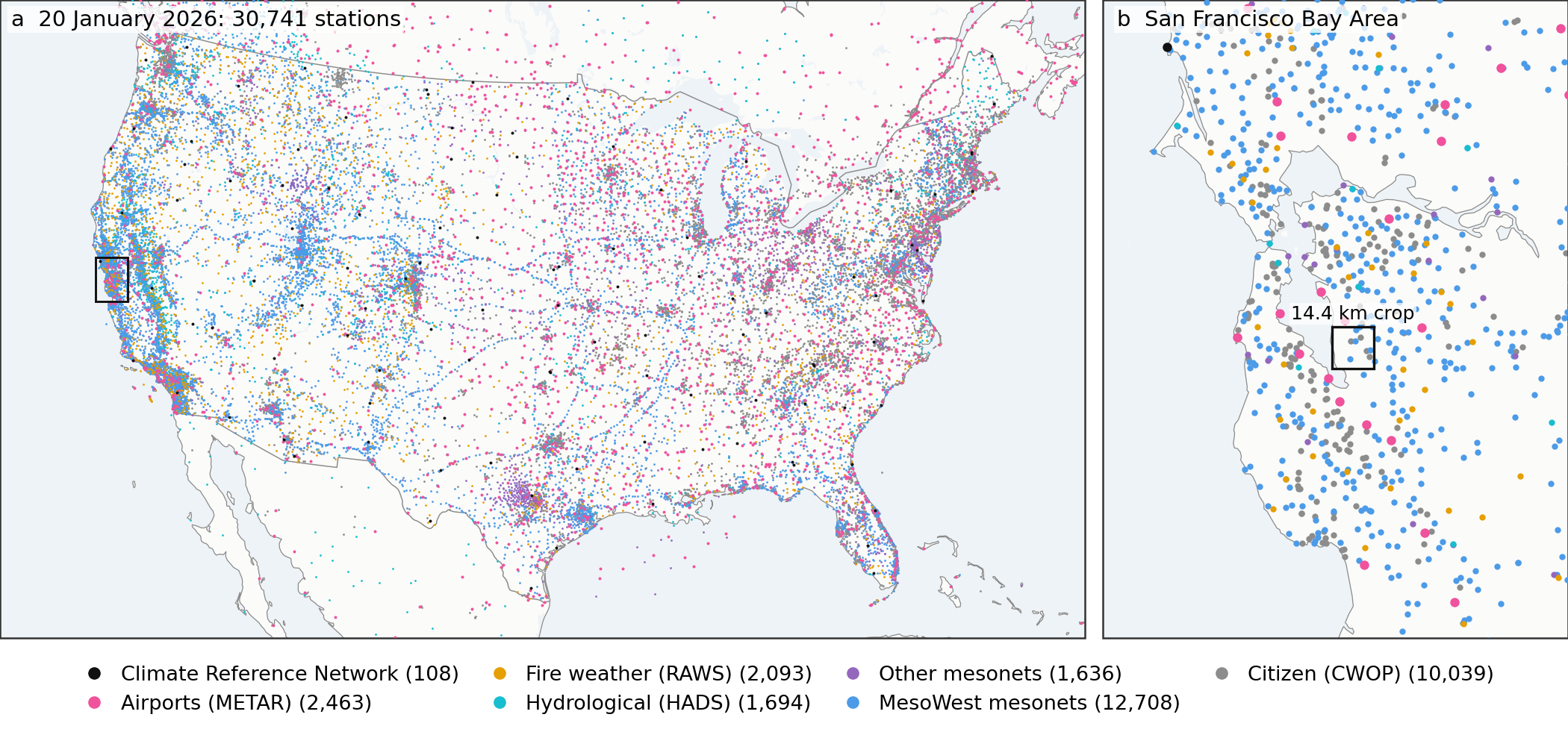}
  \caption{\textbf{Available stations on 20 January 2026.} \textbf{a}, The 30,741 land stations in the HRRR domain with at least one hour passing quality control that day, colored by network (counts in parentheses). \textbf{b}, The San Francisco Bay Area (box in \textbf{a}), with one $14.4\times14.4$\,km crop for scale. Dense volunteer and regional mesonets complement the sparser airport network. HadISD has no records in 2026.}
  \label{fig:supp_station_snapshot}
\end{figure*}

\paragraph{Weather station observations.}
We combine MADIS~\cite{madis} and HadISD~\cite{dunn2016hadisd} observations from 77,460 stations over the HRRR domain. To obtain hourly observations, we average reports within $\pm$30\,min of each hour, and then apply quality control to remove implausible values, stuck and sheltered sensors, duplicate records, and large discrepancies from HRRR analyses ($>6\sigma$ of the analysis--observation differences).
After quality control, 59,535 stations remain; \cref{tab:supp_station_stats} gives their station-hours by network and partition, and \cref{fig:supp_station_snapshot} shows the stations reporting on one day of the test period.

\begin{table}[t]
  \centering
  \footnotesize
  \setlength{\tabcolsep}{4pt}
  \begin{tabular}{@{}l r rrr@{}}
    \toprule
     & & \multicolumn{3}{c}{Station-hours (millions)} \\
    \cmidrule(l){3-5}
    & Stations & $T$ & $T_d$ & Wind \\
    \midrule
    \multicolumn{5}{@{}l}{\emph{By network, January 2017--September 2026}} \\
    Citizen (CWOP) & 21,655 & 750.7 & 730.9 & 634.6 \\
    MesoWest mesonets & 18,557 & 629.6 & 495.3 & 517.0 \\
    Fire weather (RAWS) & 6,883 & 159.7 & 164.7 & 149.1 \\
    Other mesonets & 5,051 & 158.0 & 82.2 & 118.4 \\
    Hydrological (HADS) & 4,387 & 92.2 & 7.8 & 37.1 \\
    Airports (METAR) & 2,688 & 202.8 & 200.6 & 195.4 \\
    HadISD & 199 & 9.1 & 4.4 & 7.8 \\
    Climate Reference Network & 115 & 4.0 & 2.4 & $<$0.1 \\
    \textbf{Total} & \textbf{59,535} & \textbf{2,006.1} & \textbf{1,688.5} & \textbf{1,659.3} \\
    \midrule
    \multicolumn{5}{@{}l}{\emph{By spatial partition}} \\
    Train, 2017--2025 & 45,657 & 1,436.6 & 1,212.8 & 1,187.9 \\
    Validation, 2017--2025 & 5,793 & 185.7 & 156.7 & 153.8 \\
    Test, 2017--2025 & 5,530 & 180.8 & 149.3 & 151.2 \\
    Validation, 2026 & 3,475 & 18.5 & 15.9 & 15.4 \\
    Test, 2026 & 3,482 & 18.4 & 15.4 & 15.4 \\
    \bottomrule
  \end{tabular}
  \caption{Station observations after quality control, for land stations in the HRRR domain. A station-hour counts when the variable passes every check; wind requires both components. Stations have at least one valid hour. CWOP: Citizen Weather Observer Program; RAWS: Remote Automatic Weather Stations; HADS: Hydrometeorological Automated Data System; METAR: routine airport weather reports. Partitions are the spatial split cells of \cref{fig:supp_split_map}; 2026 covers January--September.}
  \label{tab:supp_station_stats}
\end{table}

\paragraph{Station and tile sampling.}
Each training sample is drawn in three steps.
First, a station $s$ is chosen with probability $p(s)\propto n_s^{0.5}$, where $n_s$ counts its eligible prediction hours, and the square root limits the weight of stations with long records.
Second, an eligible hour is drawn uniformly, and then an HRRR forecast valid at that hour, uniformly over its available lead times.
Third, a $14.4\times14.4$\,km crop (144$\times$144 pixels at 100\,m) is placed around the station, randomly offset in training and centered in evaluation.
All valid observations from stations of the same partition inside the crop serve as targets and are averaged per variable within each 100\,m pixel.

\paragraph{Land-surface features.}
Terrain features include elevation from Copernicus GLO-30~\cite{copernicus2021glo30}, resampled from 30\,m to 100\,m, and derived surface normals.
We use OlmoEarth v1.3 embeddings~\cite{tseng2026olmoearth}, retaining the first 64 of 128 dimensions, averaged from 10\,m to 100\,m.
We also evaluate 64-dimensional AlphaEarth embeddings~\cite{brown2025alphaearth} in our ablation study (\cref{tab:ablation}).

\paragraph{Spatial split and temporal holdout.}
We divide the HRRR domain into $50\times50$\,km cells and assign the 3,554 eligible ones to train, validation and test (80/10/10\%; see \cref{fig:supp_split_map}), stratified by station density, terrain and embedding heterogeneity; each station belongs to exactly one partition.
Models are trained on the training stations over 2017--2025.
The ablations (\cref{tab:ablation}) are scored on the validation stations over the same years.
The main comparison (\cref{tab:results}) is scored on the test stations in January--September 2026, holding out both stations and time. Each evaluation set consists of 50k fixed $14.4\times14.4$\,km tiles, and on average, only 0.02\% of the pixels in a tile have a valid observation.
Ecoregion holdouts, which withhold entire regions from training, are reported in \supp~\ref{supp:ecoregion_splits_results}.

\paragraph{Operational setting.}
For operational comparison (\cref{tab:operational}), we train Vernier-Det and Vernier-Prob once over 2017--2025, with every station as a target and none held out; samples are anchored at 41,155 eligible stations (\supp~\ref{supp:sampling}).
We test it on January--September 2026.
The operational comparison covers two domains: the full HRRR domain and the Pacific Northwest (\cref{fig:supp_wrf_domain}).
In the Pacific Northwest, we evaluate the same models without retraining, using only forecasts initialized at 00 and 12\,UTC with 12--24\,h leads to match the UW-WRF~\cite{mass2003regional,uwwrf} baseline (\supp~\ref{supp:pnw}).
Within each domain, all models are scored on the same 50k tiles and forecast times.

\subsection{Evaluations}
\label{sec:evaluations}

\paragraph{Station accuracy.}
We report MAE and bias (forecast $-$ observation) at station locations for wind speed ($ws$), wind direction ($dir$), 2\,m temperature ($T$) and 2\,m dew point ($T_d$), with direction scored only for observed speeds $\geq 1.6$\,m\,s$^{-1}$.

\paragraph{Spatial structure, $R_{\mathrm{pair}}$.}
We evaluate sub-grid structure in the 100\,m outputs using the predicted and observed differences between station pairs in the same tile and hour.
Pair differences cancel shared forecast biases, so they test whether local
variations track observations.
We cap separations at HRRR's 3\,km grid spacing to focus on the local structure that downscaling aims to recover.
$R_{\mathrm{pair}}$ is the Pearson correlation between observed and predicted pair differences, averaged over the 0.1--1, 1--2 and 2--3\,km separation bands:
\begin{equation}
    R_{\mathrm{pair}} = \frac{1}{B}
    \sum_{b=1}^{B}
    \mathrm{Corr}_{(i,j)\in\mathcal{P}_b}
    (y_i-y_j,\ \hat y_i-\hat y_j),
\end{equation}
where $B=3$, $\mathcal{P}_b$ is the set of valid pairs in band $b$, and $y_i$ and $\hat y_i$ are the observation and prediction at station $i$.
For ensembles, we report $R_{\mathrm{pair}}$ of the ensemble mean as well as the average $R_{\mathrm{pair}}$ of the individual members.

\paragraph{De-meaned $R_{\mathrm{pair}}$.}
Pooled pair differences mix two kinds of structure: persistent contrasts between sites from terrain, exposure or siting, and contrasts that change with the weather.
A forecast that only learns each station's typical offset, such as per-station post-processing, reproduces the first and can reach a high $R_{\mathrm{pair}}$ without the second (\cref{tab:mos}).
To separate them, we subtract each pair's mean difference from its observed and predicted differences, taking the mean over the whole evaluation period but separately for each six-hour block of local solar time: $d'_{p,t}=d_{p,t}-\overline d_{p,b(t)}$ and $\widehat d'_{p,t}=\widehat d_{p,t}-\overline{\widehat d}_{p,b(t)}$, where $d_{p,t}=y_{i,t}-y_{j,t}$ for pair $p=(i,j)$ and $b(t)$ is the block that contains hour $t$.
The blocks also remove the regular day--night swing of each contrast, which post-processing fit by time of day can learn, so a forecast scores only by following how the contrast between two stations changes with the weather.
We keep pair-blocks with at least five station-hours and average the correlations over the same bands.

\paragraph{Probabilistic quality.}
For probabilistic models, we report fCRPS~\cite{ferro2014fair} and the spread--skill ratio (SSR) for each variable
\begin{equation}
    \mathrm{SSR} =
    \frac{\sigma_{\mathrm{ens}}\sqrt{1+1/K}}
         {\mathrm{RMSE}(\bar x)},
\end{equation}
where $K$ is the ensemble size, $\sigma_{\mathrm{ens}}$
the ensemble standard deviation, and $\bar x$ the
ensemble mean. Lower fCRPS is better, and SSR
around 1 indicates better agreement between
ensemble spread and forecast error.

\paragraph{Baselines.}
We compare Vernier against eleven baselines, including operational models.
Configuration details are provided in \supp~\ref{supp:baselines}.

\begin{itemize}
    \item \textbf{HRRR references:}
    (1) Bicubically interpolated HRRR,
    (2) an HRRR-only U-Net for station bias correction
    without high-resolution surface inputs, and
    (3) Model Output Statistics (MOS)~\cite{glahn1972mos}, which corrects HRRR at each station with a linear regression fit on that station's own 2017--2025 record, per station or per station and time of day, and its probabilistic form, ensemble MOS (EMOS)~\cite{gneiting2005emos}, a Gaussian around the MOS line.
    All three take HRRR as their only input and differ in how they correct it: not at all, with one network for all stations, or with a separate fit per station.
    MOS and EMOS need each station's history, so we use them only in the operational comparison, where every test station has one.

    \item \textbf{Physical downscaling:}
    (4) Our physics prior alone,
    (5) WOOF~\cite{woof}, a GPU-native regional weather model (WRF-ARW class),
    (6) WindNinja~\cite{forthofer2014comparison}, a diagnostic wind downscaler, and
    (7) UW-WRF~\cite{mass2003regional,uwwrf}, a 1.33\,km operational forecast over the Pacific Northwest domain.

    \item \textbf{ML-based models:}
    (8) U-Net,
    (9) ConvCNP~\cite{sousa2026earth}, and
    (10) CorrDiff~\cite{mardani2025corrdiff}, a residual downscaler that uses the physics prior as its mean.
    Models (8)--(10) receive the same inputs as Vernier: the HRRR forecast, terrain, Earth embeddings and the physics prior.
    (11) WeatherNext~3~\cite{rasp2026weathernext}, a global probabilistic forecast whose temperature and dew point are trained on station observations.
\end{itemize}

\paragraph{Implementation details.}
Vernier-Det (129M parameters) is trained from scratch for 4,000 steps on 8 NVIDIA H100 GPUs with a global batch of 2,240 tiles (77 GPU-hours), using AdamW with a cosine learning-rate schedule.
Vernier-Prob finetunes Vernier-Det for 10,000 steps with two members per sample; its 64 reported members pool 32 each from the weather-modulated and the global-modulated model (* in \cref{tab:ablation}).
In \cref{tab:results}, the Vernier rows are trained without the spectral loss and the \emph{+ Spectral} rows with it; the operational models of \cref{tab:operational} use it.
The \emph{+ Searchlight} rows use Searchlight (\cref{sec:searchlight}) over nine times the tile size, cropped to the central tile (matching evaluation with all other baselines), to assess the effect of using Searchlight in the operational setting.
The HRRR station correction and the U-Net use the same station loss as Vernier-Det (\cref{eq:deterministic_loss}), CorrDiff uses flow matching, and ConvCNP keeps its negative log-likelihood; each learned baseline uses 74--136 GPU-hours.
Full settings are given in \supp~\ref{supp:training} and~\ref{supp:baselines}.

\subsection{Main Results}

\ifdefined\tabresultsbox\else\newsavebox{\tabresultsbox}\fi
\providecommand{\RpairDM}[1]{\,{\scriptsize(#1)}}
\begin{table}[t]
\centering
\footnotesize
\setlength{\tabcolsep}{2pt}
\definecolor{ai2pinkrow}{HTML}{F0529C}%
\sbox{\tabresultsbox}{%
\begin{tabular}{l cccc ccc >{\hspace{0.5pt}}c<{\hspace{0.5pt}}>{\hspace{0.5pt}}c<{\hspace{0.5pt}}>{\hspace{0.5pt}}c<{\hspace{0.5pt}}}
\toprule
\emph{Deterministic} & \multicolumn{4}{c}{MAE $\downarrow$} & \multicolumn{3}{c}{Bias $\to 0$} & \multicolumn{3}{c}{$R_{\mathrm{pair}}$ (de-meaned) $\uparrow$} \\
\cmidrule(lr){2-5}\cmidrule(lr){6-8}\cmidrule(lr){9-11}
Model & ws & dir & $T$ & $T_d$ & ws & $T$ & $T_d$ & ws & $T$ & $T_d$ \\
 & {\scriptsize m\,s$^{-1}$} & {\scriptsize $^\circ$} & {\scriptsize $^\circ$C} & {\scriptsize $^\circ$C} & {\scriptsize m\,s$^{-1}$} & {\scriptsize $^\circ$C} & {\scriptsize $^\circ$C} & {\scriptsize 171k} & {\scriptsize 171k} & {\scriptsize 170k} \\
\midrule
HRRR interp. & 1.97 & 36.98 & 1.71 & 2.53 & +1.70 & +0.07 & $-$1.57 & 0.06\RpairDM{0.05} & 0.19\RpairDM{0.14} & 0.15\RpairDM{0.08}  \\
HRRR stn.\ corr. & 0.99 & 35.44 & 1.42 & 1.70 & +0.12 & $-$0.06 & $-$0.22 & 0.10\RpairDM{0.05} & 0.24\RpairDM{0.18} & 0.20\RpairDM{0.08}  \\
Physics prior & 1.00 & 36.23 & 1.68 & 2.48 & $-$0.29 & +0.07 & $-$1.45 & 0.35\RpairDM{0.16} & 0.36\RpairDM{0.30} & 0.27\RpairDM{0.08}  \\
U-Net & 0.84 & 34.92 & 1.30 & 1.62 & \underline{+0.03} & \textbf{$-$0.03} & \textbf{+0.03} & 0.43\RpairDM{0.20} & 0.52\RpairDM{0.42} & 0.39\RpairDM{0.22}  \\
ConvCNP$^{\dagger}$ & 0.87 & \underline{33.38} & 1.34 & 1.63 & +0.20 & $-$0.09 & $-$0.15 & 0.48\RpairDM{0.21} & 0.48\RpairDM{0.38} & 0.40\RpairDM{0.18}  \\
CorrDiff$^{\dagger}$ & 0.85 & 35.52 & 1.34 & 1.66 & $-$0.25 & +0.15 & \underline{+0.05} & 0.47\RpairDM{0.23} & 0.50\RpairDM{0.40} & 0.35\RpairDM{0.18}  \\
\rowcolor{ai2pinkrow!10}\textbf{Vernier-Det} & \underline{0.83} & 33.45 & \underline{1.28} & 1.57 & +0.08 & \underline{$-$0.04} & +0.07 & \underline{0.50}\RpairDM{0.23} & \underline{0.53}\RpairDM{0.43} & \underline{0.43}\RpairDM{\underline{0.25}}  \\
\rowcolor{ai2pinkrow!10}\quad\emph{+ Spectral} & 0.85 & \textbf{33.32} & \underline{1.28} & \textbf{1.56} & +0.06 & \underline{$-$0.04} & +0.07 & 0.47\RpairDM{\textbf{0.24}} & \textbf{0.54}\RpairDM{\textbf{0.44}} & \underline{0.43}\RpairDM{\underline{0.25}}  \\
\rowcolor{ai2pinkrow!10}\textbf{Vernier-Prob}$^{\dagger}$ & \textbf{0.81} & 33.55 & \underline{1.28} & \textbf{1.56} & \textbf{$-$0.01} & $-$0.05 & +0.10 & \textbf{0.51}\RpairDM{\textbf{0.24}} & \underline{0.53}\RpairDM{\textbf{0.44}} & \underline{0.43}\RpairDM{\underline{0.25}}  \\
\rowcolor{ai2pinkrow!10}\quad\emph{+ Spectral}$^{\dagger}$ & \underline{0.83} & 33.73 & \textbf{1.27} & \textbf{1.56} & $-$0.04 & $-$0.05 & +0.11 & 0.46\RpairDM{\textbf{0.24}} & \underline{0.53}\RpairDM{\textbf{0.44}} & \textbf{0.44}\RpairDM{\textbf{0.26}}  \\
\midrule
\emph{Probabilistic} & \multicolumn{4}{c}{fCRPS $\downarrow$} & \multicolumn{3}{c}{SSR $\to 1$} & \multicolumn{3}{c}{Member $R_{\mathrm{pair}}$ (de-meaned) $\uparrow$} \\
\cmidrule(lr){2-5}\cmidrule(lr){6-8}\cmidrule(lr){9-11}
Model & ws & dir & $T$ & $T_d$ & ws & $T$ & $T_d$ & ws & $T$ & $T_d$ \\
\midrule
ConvCNP & 0.60 & 24.19 & 0.96 & 1.17 & \underline{0.88} & \textbf{0.97} & \textbf{0.98} & 0.24\RpairDM{0.05} & 0.17\RpairDM{0.08} & 0.09\RpairDM{0.02}  \\
CorrDiff & 0.64 & 25.52 & 1.00 & 1.23 & 0.55 & 0.65 & 0.62 & 0.33\RpairDM{0.11} & 0.30\RpairDM{0.18} & 0.15\RpairDM{0.05}  \\
\rowcolor{ai2pinkrow!10}\textbf{Vernier-Prob} & \textbf{0.56} & \textbf{23.97} & \underline{0.91} & \underline{1.12} & \textbf{0.89} & \underline{0.95} & \underline{0.92} & \textbf{0.41}\RpairDM{\underline{0.13}} & \underline{0.45}\RpairDM{\underline{0.33}} & \underline{0.35}\RpairDM{\underline{0.16}}  \\
\rowcolor{ai2pinkrow!10}\quad\emph{+ Spectral} & \underline{0.58} & \underline{24.03} & \textbf{0.90} & \textbf{1.11} & 0.85 & 0.94 & \underline{0.92} & \underline{0.39}\RpairDM{\textbf{0.15}} & \textbf{0.48}\RpairDM{\textbf{0.36}} & \textbf{0.38}\RpairDM{\textbf{0.19}}  \\
\bottomrule
\end{tabular}}%
\ifdim\wd\tabresultsbox>\columnwidth\resizebox{\columnwidth}{!}{\usebox{\tabresultsbox}}\else\usebox{\tabresultsbox}\fi
\caption{\textbf{Spatio-temporal holdout results:} test stations in January--September 2026 (50k samples, 222k station-hours, leads 1--48\,h).
Ensembles have 64 members; member $R_{\mathrm{pair}}$ is averaged over members.
In parentheses: de-meaned $R_{\mathrm{pair}}$, computed after removing each station pair's mean difference, taken over the whole test period separately for each six-hour block of local solar time, so persistent offsets between the two stations, including their average day--night cycle, do not count.
The row below the headings gives units and, for $R_{\mathrm{pair}}$, the number of station pairs 0.1--3\,km apart.
$^{\dagger}$Ensemble mean; $ws$ is the mean of member speeds, and ConvCNP uses its analytic predictive mean for $T$, $T_d$ and dir.
\emph{+ Spectral}: trained with the spectral loss (\cref{sec:imprint}).
Bold: best; underline: second best, none when the best is tied; raw and de-meaned $R_{\mathrm{pair}}$ are ranked separately.}
\label{tab:results}
\end{table}

\subsubsection{Comparison with baselines}
On the strict spatio-temporal holdout (\cref{tab:results}), Vernier achieves the lowest MAE and highest $R_{\mathrm{pair}}$ for $ws$, $T$, and $T_d$.
Removing each station pair's mean difference (de-meaned $R_{\mathrm{pair}}$, in parentheses) lowers every model's score, and Vernier remains highest for all three variables (0.24, 0.44 and 0.25--0.26), three to five times HRRR (0.05, 0.14 and 0.08), while HRRR station correction falls close to HRRR's level (0.05, 0.18 and 0.08), so its raw gain reflects persistent site offsets.
A paired bootstrap on the 2017--2025 spatial holdout, which removes one overall mean per pair, shows that Vernier's gain over the physics prior stays positive for all three variables (95\% intervals above zero), so its gains go beyond persistent site offsets (\cref{tab:pair_structure,fig:pair_structure}, \supp~\ref{supp:structure}).
Vernier-Det raises wind-speed $R_{\mathrm{pair}}$ from HRRR's 0.06 to 0.50, and the Vernier-Prob ensemble mean further improves wind speed, lowering its MAE from 0.83 to 0.81\,m\,s$^{-1}$ and its bias from $+0.08$ to $-0.01$\,m\,s$^{-1}$.
Station correction and the physics prior both reduce HRRR's wind-speed errors, but the prior recovers more spatial structure.
MAE gains over learned baselines are modest: the U-Net has the smallest $T$ and $T_d$ biases, and ConvCNP has a slightly lower direction MAE than Vernier-Det.
The spectral loss (\emph{+ Spectral} rows) suppresses surface artifacts (\cref{fig:spectral_loss}) at a small cost in wind speed: MAE rises by 0.02\,m\,s$^{-1}$ and raw $R_{\mathrm{pair}}$ falls by 0.03--0.05 (de-meaned $R_{\mathrm{pair}}$ does not fall), while the $T$ and $T_d$ scores of the deterministic models and ensemble means change by at most 0.01.

Vernier-Prob's advantage is clearer in ensemble quality: with or without the spectral loss, it has the lowest fCRPS for every variable and the highest member $R_{\mathrm{pair}}$.
Without the spectral loss, its fCRPS is 4--7\% below ConvCNP's for $ws$, $T$, and $T_d$.
It is modestly underdispersed (SSR 0.89--0.95). Among the ensembles, its wind-speed SSR is the closest to one (0.89, against 0.88 for ConvCNP), while ConvCNP's is closer for $T$ and $T_d$ (0.97 and 0.98, against 0.95 and 0.92).
CorrDiff is more underdispersed (SSR 0.55--0.65).
ConvCNP has comparable mean accuracy, but its members recover much less station-pair structure (member $R_{\mathrm{pair}}$ 0.09--0.24 vs.\ 0.35--0.45; de-meaned 0.02--0.08 vs.\ 0.13--0.33).

Separate 2017--2025 ecoregion and spatial-only holdouts also show Vernier's accuracy and structural gains over HRRR and the other models (\supp~\ref{supp:ecoregion_splits_results}).

\providecommand{\RpairDM}[1]{\,{\scriptsize(#1)}}%
\begin{table}[t]
\centering
\footnotesize
\definecolor{ai2pinkrow}{HTML}{F0529C}%
\setlength{\tabcolsep}{3pt}
\resizebox{\tabfitwidth}{!}{%
\begin{tabular}{l ccc ccc >{\hspace{1pt}}c<{\hspace{1pt}}>{\hspace{1pt}}c<{\hspace{1pt}}>{\hspace{1pt}}c<{\hspace{1pt}}}
\toprule
\emph{Deterministic} & \multicolumn{3}{c}{MAE $\downarrow$} & \multicolumn{3}{c}{Bias $\to 0$} & \multicolumn{3}{c}{$R_{\mathrm{pair}}$ (de-meaned) $\uparrow$} \\
\cmidrule(lr){2-4}\cmidrule(lr){5-7}\cmidrule(lr){8-10}
Model & ws & $T$ & $T_d$ & ws & $T$ & $T_d$ & ws & $T$ & $T_d$ \\
 & {\scriptsize m\,s$^{-1}$} & {\scriptsize $^\circ$C} & {\scriptsize $^\circ$C} & {\scriptsize m\,s$^{-1}$} & {\scriptsize $^\circ$C} & {\scriptsize $^\circ$C} & {\scriptsize 278k} & {\scriptsize 276k} & {\scriptsize 275k} \\
\midrule
\multicolumn{10}{l}{\emph{HRRR DOMAIN}} \\
HRRR & 1.91 & 1.67 & 2.57 & +1.64 & \underline{$-$0.06} & $-$1.69 & 0.08\RpairDM{0.08} & 0.25\RpairDM{0.20} & 0.20\RpairDM{0.13}  \\
HRRR + MOS, per station & 0.72 & 1.55 & 2.09 & +0.03 & $-$0.22 & $-$0.63 & 0.68\RpairDM{0.29} & 0.42\RpairDM{0.17} & 0.46\RpairDM{0.14}  \\
HRRR + MOS, station $\times$ ToD & 0.68 & 1.43 & 2.03 & +0.03 & $-$0.20 & $-$0.62 & \textbf{0.71}\RpairDM{0.31} & 0.56\RpairDM{0.22} & \textbf{0.51}\RpairDM{0.15}  \\
WindNinja & 1.87 & -- & -- & +1.61 & -- & -- & 0.31\RpairDM{0.18} & -- & --  \\
WOOF & 2.05 & 1.60 & 2.42 & +1.81 & \textbf{$-$0.01} & $-$1.52 & 0.28\RpairDM{0.19} & 0.39\RpairDM{0.35} & 0.25\RpairDM{0.17}  \\
WeatherNext 3 & 1.32 & 1.26 & \textbf{1.46} & +0.59 & $-$0.08 & $-$0.29 & 0.03\RpairDM{0.03} & 0.24\RpairDM{0.21} & 0.21\RpairDM{0.16}  \\
\rowcolor{ai2pinkrow!10}\textbf{Vernier-Det} & 0.73 & 1.19 & \underline{1.52} & +0.11 & $-$0.13 & \underline{$-$0.03} & 0.62\RpairDM{0.38} & 0.67\RpairDM{0.61} & 0.44\RpairDM{0.33}  \\
\rowcolor{ai2pinkrow!10}\quad\emph{+ Searchlight} & 0.73 & 1.19 & 1.53 & +0.09 & $-$0.10 & \textbf{0.00} & 0.62\RpairDM{0.40} & 0.67\RpairDM{0.61} & 0.44\RpairDM{0.34}  \\
\rowcolor{ai2pinkrow!10}\textbf{Vernier-Prob}$^{\dagger}$ & \textbf{0.67} & \textbf{1.18} & \underline{1.52} & \underline{+0.02} & $-$0.13 & \underline{$-$0.03} & \underline{0.69}\RpairDM{\underline{0.45}} & \underline{0.69}\RpairDM{\textbf{0.63}} & \underline{0.48}\RpairDM{\underline{0.35}}  \\
\rowcolor{ai2pinkrow!10}\quad\emph{+ Searchlight}$^{\dagger}$ & \textbf{0.67} & \textbf{1.18} & 1.53 & \textbf{0.00} & $-$0.09 & \underline{+0.03} & 0.68\RpairDM{\textbf{0.46}} & \textbf{0.70}\RpairDM{\textbf{0.63}} & \underline{0.48}\RpairDM{\textbf{0.36}}  \\
\addlinespace[1pt]
\arrayrulecolor{black!30}\midrule[0.4pt]\arrayrulecolor{black}%
\multicolumn{7}{l}{\emph{PACIFIC NORTHWEST}} & {\scriptsize 97k} & {\scriptsize 96k} & {\scriptsize 96k} \\
UW-WRF & \underline{1.78} & \underline{1.72} & 2.47 & \underline{+1.42} & \underline{+0.28} & $-$1.81 & 0.07\RpairDM{0.04} & 0.23\RpairDM{0.09} & \underline{0.07}\RpairDM{\underline{0.04}}  \\
WOOF & 2.06 & 1.75 & \underline{2.09} & +1.78 & +0.59 & \underline{$-$1.00} & \underline{0.21}\RpairDM{\underline{0.12}} & \underline{0.26}\RpairDM{\underline{0.16}} & 0.05\RpairDM{0.00}  \\
\rowcolor{ai2pinkrow!10}\textbf{Vernier-Det} & \textbf{0.81} & \textbf{1.29} & \textbf{1.40} & \textbf{+0.10} & \textbf{$-$0.01} & \textbf{+0.17} & \textbf{0.50}\RpairDM{\textbf{0.34}} & \textbf{0.53}\RpairDM{\textbf{0.40}} & \textbf{0.41}\RpairDM{\textbf{0.29}}  \\
\addlinespace[6pt]%
\midrule
\emph{Probabilistic} & \multicolumn{3}{c}{fCRPS $\downarrow$} & \multicolumn{3}{c}{SSR $\to 1$} & \multicolumn{3}{c}{Member $R_{\mathrm{pair}}$ (de-meaned) $\uparrow$} \\
\cmidrule(lr){2-4}\cmidrule(lr){5-7}\cmidrule(lr){8-10}
Model & ws & $T$ & $T_d$ & ws & $T$ & $T_d$ & ws & $T$ & $T_d$ \\
\midrule
\multicolumn{10}{l}{\emph{HRRR DOMAIN}} \\%
HRRR + EMOS, per station & 0.54 & 1.11 & 1.51 & 0.88 & \textbf{1.00} & \textbf{0.98} & 0.43\RpairDM{0.09} & 0.14\RpairDM{0.04} & 0.14\RpairDM{0.03}  \\
HRRR + EMOS, station $\times$ ToD & 0.51 & 1.03 & 1.47 & 0.89 & \underline{0.99} & \textbf{0.98} & 0.46\RpairDM{0.09} & 0.24\RpairDM{0.06} & 0.16\RpairDM{0.03}  \\
WeatherNext 3 & 1.10 & 0.89 & \textbf{1.03} & 0.45 & 0.89 & 0.89 & 0.02\RpairDM{0.02} & 0.22\RpairDM{0.18} & 0.19\RpairDM{0.11}  \\
\rowcolor{ai2pinkrow!10}\textbf{Vernier-Prob} & \textbf{0.46} & \textbf{0.83} & \underline{1.08} & \textbf{0.99} & 0.98 & 0.93 & \underline{0.60}\RpairDM{\underline{0.31}} & \textbf{0.63}\RpairDM{\underline{0.53}} & \underline{0.41}\RpairDM{\underline{0.26}}  \\
\rowcolor{ai2pinkrow!10}\quad\emph{+ Searchlight} & \underline{0.47} & \underline{0.84} & 1.09 & \underline{0.95} & 0.91 & 0.87 & \textbf{0.61}\RpairDM{\textbf{0.33}} & \textbf{0.63}\RpairDM{\textbf{0.54}} & \textbf{0.42}\RpairDM{\textbf{0.27}}  \\
\bottomrule
\end{tabular}%
}
\caption{\textbf{Operational comparison with a temporal holdout (2026)}.
\emph{HRRR domain}: January--September 2026, 50k samples, 265k station-hours, leads 3--48\,h.
\emph{Pacific Northwest} (UW-WRF domain; deterministic models only): 50k samples, 168k station-hours, 00 and 12\,UTC initializations, leads 12--24\,h.
The row below the headings gives units and, for $R_{\mathrm{pair}}$, the number of station pairs 0.1--3\,km apart in the HRRR domain; the \emph{Pacific Northwest} row gives that domain's.
In parentheses: de-meaned $R_{\mathrm{pair}}$, computed after removing each station pair's mean difference, taken over the whole test period separately for each six-hour block of local solar time, so persistent offsets between the two stations, including their average day--night cycle, do not count.
MOS: HRRR corrected by a linear fit on each station's 2017--2025 record, per station or per station and six-hour block of local solar time (ToD; \supp~\ref{supp:mos_config}), scored on the 98.5--98.9\% of station-hours where every forecast exists. EMOS: MOS with a Gaussian spread, scored by its exact CRPS, which the fCRPS of an ensemble estimates without bias, and by the SSR of its predictive mean and variance (zero-truncated for wind speed).
Member $R_{\mathrm{pair}}$: $R_{\mathrm{pair}}$ averaged over the individual members (EMOS: 64 draws sampled independently at each station-hour).
$^{\dagger}$Ensemble mean; $ws$ is the mean of member speeds.
\emph{+ Searchlight}: the same model run over $9\times$ the tile area and scored on the central tile.
Bold: best; underline: second best, none when the best is tied, within each domain; raw and de-meaned $R_{\mathrm{pair}}$ are ranked separately.}
\label{tab:operational}
\end{table}

\subsubsection{Comparison with operational models}
When trained on all eligible stations through 2025, Vernier also leads most of the 2026 operational comparison (\cref{tab:operational}), though not every column: WeatherNext~3 (WN3) is better on dew point, and station post-processing fit per station and time of day reaches a higher raw $R_{\mathrm{pair}}$ for wind speed and dew point, which we examine below; WOOF and HRRR also have smaller temperature biases, and EMOS has a spread--skill ratio closer to one for $T$ and $T_d$.
Over the HRRR domain, Vernier-Prob has the lowest wind-speed MAE and the two Vernier models the lowest temperature MAE; among the gridded forecasts, they have the highest $R_{\mathrm{pair}}$ for all three variables, and once persistent offsets are removed, the highest de-meaned $R_{\mathrm{pair}}$ of any forecast, MOS included. WN3's advantage in dew-point MAE is small (1.46 vs.\ 1.52\,$^\circ$C).
Vernier-Prob cuts WN3's wind-speed fCRPS by 58\% and lowers its temperature fCRPS from 0.89 to 0.83, with SSR closer to one for all three variables, while WN3's dew-point fCRPS remains lower (1.03 vs.\ 1.08).
In the Pacific Northwest, Vernier-Det is best in every column, against both UW-WRF and WOOF.

With Searchlight over $9\times$ the tile area (``\textit{+ Searchlight}'' in~\cref{tab:operational}), MAE, fCRPS and raw $R_{\mathrm{pair}}$ change by at most 0.01 (de-meaned $R_{\mathrm{pair}}$ by up to 0.02), so Vernier's ranking against the operational baselines is unchanged.
The main cost is a slightly narrower ensemble (SSR 0.87--0.95, against 0.93--0.99 without Searchlight). For dew point, WN3's SSR (0.89) is then closer to one than Vernier-Prob's (0.87).

\subsubsection{Is Vernier just station de-biasing?}
Operational forecasting removes a station's systematic errors with statistical post-processing, and in the operational test every station has a 2017--2025 record to fit MOS on (\cref{tab:operational}; configuration in \supp~\ref{supp:mos_config}, full comparison in \supp~\ref{supp:mos}).
This removes most of HRRR's wind-speed error, because a station's exposure acts as a fixed multiplier on its wind: MOS per station and time of day lowers the wind-speed MAE from 1.91 to 0.68\,m\,s$^{-1}$, between Vernier-Det (0.73) and Vernier-Prob (0.67).
Temperature and dew-point errors depend more on the weather than on the site, and there Vernier-Det's MAE is a further 17\% and 25\% below the best MOS.
A line per station also builds each site's long-run contrast with its neighbors into the forecast, which gives MOS a high $R_{\mathrm{pair}}$ for wind speed and dew point (0.71 and 0.51, above Vernier-Prob's 0.69 and 0.48).
Once each station pair's mean difference is removed, Vernier leads for every variable: the de-meaned $R_{\mathrm{pair}}$ of Vernier-Prob is 0.45, 0.63 and 0.35, against 0.31, 0.22 and 0.15 for MOS (\cref{tab:operational}; \cref{tab:mos} gives 95\% intervals for Vernier-Det's lead). MOS can only rescale HRRR at each station, while the contrast between neighbors changes with the weather.
De-meaning removes 56--71\% of MOS's $R_{\mathrm{pair}}$ but only 9--35\% of Vernier-Prob's: most of MOS's station-pair skill is the fixed contrast it copies from each station's history, while most of Vernier's is how that contrast changes with the weather.
Its probabilistic form, EMOS, is also behind: Vernier-Prob's fCRPS is 9--27\% below the EMOS CRPS for all three variables, and the EMOS draws, sampled independently at each station, keep almost none of the changing structure (member de-meaned $R_{\mathrm{pair}}$ at most 0.09, against 0.26--0.53 for Vernier-Prob's members; \cref{tab:operational}).
Per-station calibration matches Vernier on wind-speed error, but its de-meaned $R_{\mathrm{pair}}$ is only 35--70\% of Vernier-Prob's: Vernier is not just de-biasing stations, but recovering inter-station structure.

\subsubsection{Does 100 m detail matter?}
\label{sec:smoothing}
\begin{figure*}[t]
  \centering
  \includegraphics[width=\textwidth]{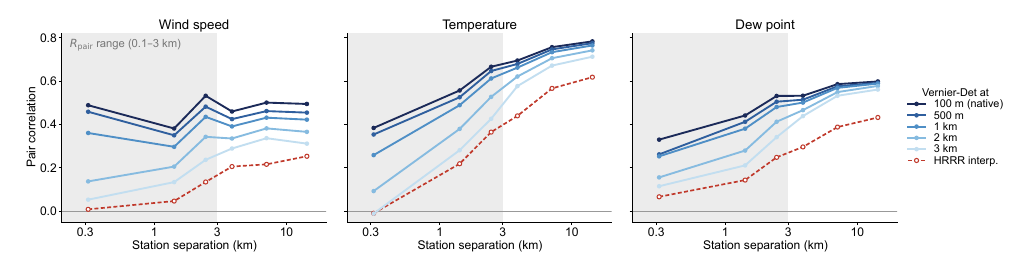}
  \caption{\textbf{Change in station-pair correlation by separation while downsampling} (Vernier-Det with the spectral loss) at 100\,m and averaged to 500\,m--3\,km (bicubic back to 100\,m), and for HRRR (dashed), on the test set of \cref{tab:results}. Points sit at the geometric center of each separation band (0.1--1, 1--2, 2--3, 3--5, 5--10 and 10--20.4\,km); gray marks the 0.1--3\,km separations averaged in $R_{\mathrm{pair}}$.}\label{fig:smoothing}
\end{figure*}

To test what the 100\,m detail adds, we average the prediction of Vernier-Det with the spectral loss (\cref{tab:results}) onto cells of 500\,m, 1\,km, 2\,km and 3\,km, interpolate it back to 100\,m with bicubic interpolation, and score it at the same stations and times. Wind is averaged as $u$ and $v$ components. Averaging changes the MAE little (\cref{tab:smoothing}): from 100\,m to 3\,km, the wind-speed MAE rises from 0.85 to 0.89\,m\,s$^{-1}$, the temperature MAE from 1.28 to 1.34\,$^\circ$C and the dew-point MAE from 1.56 to 1.59\,$^\circ$C. Of Vernier-Det's reduction of HRRR's wind-speed MAE, 96\% remains after averaging to 3\,km. The structure between nearby stations does not remain: $R_{\mathrm{pair}}$ for wind speed falls from 0.47 at 100\,m to 0.43, 0.36, 0.23 and 0.14 at 500\,m, 1\,km, 2\,km and 3\,km, against 0.06 for HRRR, and for temperature and dew point it falls from 0.54 and 0.43 to 0.23 and 0.22. The loss is largest for the closest station pairs (\cref{fig:smoothing}): for wind speed at 0.1--1\,km, the pair correlation falls from 0.49 to 0.05, close to HRRR's 0.01. Every step toward 100\,m lowers the MAE, if only slightly, and raises $R_{\mathrm{pair}}$, and the gain in structure is much larger than the gain in MAE.

\ifdefined\supptabbox\else\newsavebox{\supptabbox}\fi
\begin{table}[t]
\centering
\footnotesize
\setlength{\tabcolsep}{2pt}
\sbox{\supptabbox}{%
\begin{tabular}{l cccc ccc ccc}
\toprule
 & \multicolumn{4}{c}{MAE $\downarrow$} & \multicolumn{3}{c}{Bias $\to 0$} & \multicolumn{3}{c}{$R_{\mathrm{pair}}$ $\uparrow$} \\
\cmidrule(lr){2-5}\cmidrule(lr){6-8}\cmidrule(lr){9-11}
 & ws & dir & $T$ & $T_d$ & ws & $T$ & $T_d$ & ws & $T$ & $T_d$ \\
 & {\scriptsize m\,s$^{-1}$} & {\scriptsize $^\circ$} & {\scriptsize $^\circ$C} & {\scriptsize $^\circ$C} & {\scriptsize m\,s$^{-1}$} & {\scriptsize $^\circ$C} & {\scriptsize $^\circ$C} & {\scriptsize 171k} & {\scriptsize 171k} & {\scriptsize 170k} \\
\midrule
HRRR interp. & 1.97 & 36.98 & 1.71 & 2.53 & +1.70 & +0.07 & $-$1.57 & 0.06 & 0.19 & 0.15 \\
\midrule
100\,m (native) & 0.85 & 33.32 & 1.28 & 1.56 & +0.06 & $-$0.04 & +0.07 & 0.47 & 0.54 & 0.43 \\
500\,m & 0.85 & 33.56 & 1.28 & 1.57 & +0.02 & $-$0.01 & +0.08 & 0.43 & 0.51 & 0.39 \\
1\,km & 0.86 & 33.75 & 1.29 & 1.57 & $-$0.01 & +0.01 & +0.08 & 0.36 & 0.45 & 0.37 \\
2\,km & 0.88 & 34.12 & 1.32 & 1.58 & $-$0.05 & +0.02 & +0.08 & 0.23 & 0.33 & 0.28 \\
3\,km & 0.89 & 34.37 & 1.34 & 1.59 & $-$0.07 & +0.01 & +0.07 & 0.14 & 0.23 & 0.22 \\
\bottomrule
\end{tabular}}%
\ifdim\wd\supptabbox>\columnwidth\resizebox{\columnwidth}{!}{\usebox{\supptabbox}}\else\usebox{\supptabbox}\fi
\caption{\textbf{Importance of 100\,m resolution}. Vernier-Det with the spectral loss on the test set of \cref{tab:results}, scored at 100\,m and after averaging its prediction onto coarser cells and interpolating it back to 100\,m (bicubic). The 100\,m row is the \emph{+ Spectral} row of \cref{tab:results}. The row below the column names gives units and, for $R_{\mathrm{pair}}$, the number of station pairs 0.1--3\,km apart, the same at every scale.}
\label{tab:smoothing}
\end{table}

\definecolor{abBetter}{HTML}{4C9BE8}
\definecolor{abWorse}{HTML}{F0529C}
\begin{table*}[t]
\centering
\footnotesize
\setlength{\tabcolsep}{3.5pt}
\resizebox{\textwidth}{!}{%
\begin{tabular}{l l cccc ccc ccc}
\toprule
\multicolumn{2}{l}{\emph{Deterministic}} & \multicolumn{4}{c}{MAE $\downarrow$} & \multicolumn{3}{c}{Bias $\to 0$} & \multicolumn{3}{c}{$R_{\mathrm{pair}}$ $\uparrow$} \\
\cmidrule(lr){3-6}\cmidrule(lr){7-9}\cmidrule(lr){10-12}
Case & Ablation & ws & dir & $T$ & $T_d$ & ws & $T$ & $T_d$ & ws & $T$ & $T_d$ \\
 &  & {\scriptsize m\,s$^{-1}$} & {\scriptsize $^\circ$} & {\scriptsize $^\circ$C} & {\scriptsize $^\circ$C} & {\scriptsize m\,s$^{-1}$} & {\scriptsize $^\circ$C} & {\scriptsize $^\circ$C} & {\scriptsize 226k} & {\scriptsize 236k} & {\scriptsize 238k} \\
\midrule
\multicolumn{2}{l}{\textbf{Vernier-Det (reference)}} & 0.853\,{\scriptsize$\pm$0.002} & 30.59\,{\scriptsize$\pm$0.056} & 1.242\,{\scriptsize$\pm$0.000} & 1.558\,{\scriptsize$\pm$0.002} & $-$0.032\,{\scriptsize$\pm$0.012} & +0.012\,{\scriptsize$\pm$0.014} & $-$0.001\,{\scriptsize$\pm$0.018} & 0.507\,{\scriptsize$\pm$0.006} & 0.562\,{\scriptsize$\pm$0.004} & 0.320\,{\scriptsize$\pm$0.008} \\
 Finescale streams & No prior & \cellcolor{abWorse!14}+0.009 & \cellcolor{abWorse!50}+0.86 & \cellcolor{abWorse!28}+0.015 & \cellcolor{abWorse!13}+0.003 & \cellcolor{abBetter!19}+0.051 & $-$0.011 & +0.010 & \cellcolor{abWorse!40}$-$0.084 & \cellcolor{abWorse!44}$-$0.044 & +0.007 \\
  & No embedding & \cellcolor{abWorse!50}+0.180 & \cellcolor{abBetter!25}$-$0.19 & \cellcolor{abWorse!50}+0.052 & \cellcolor{abWorse!50}+0.070 & \cellcolor{abWorse!34}+0.206 & \cellcolor{abBetter!43}$-$0.016 & \cellcolor{abWorse!50}$-$0.113 & \cellcolor{abWorse!50}$-$0.136 & \cellcolor{abWorse!39}$-$0.033 & \cellcolor{abWorse!50}$-$0.024 \\
  & No terrain & \cellcolor{abWorse!12}+0.006 & \cellcolor{abWorse!50}+0.87 & \cellcolor{abWorse!37}+0.027 & \cellcolor{abWorse!13}+0.003 & \cellcolor{abWorse!14}$-$0.026 & $-$0.006 & \cellcolor{abWorse!22}+0.019 & \cellcolor{abWorse!14}$-$0.006 & \cellcolor{abWorse!39}$-$0.034 & +0.006 \\
 Decoder & Linear & \cellcolor{abWorse!19}+0.020 & +0.01 & \cellcolor{abWorse!27}+0.014 & \cellcolor{abWorse!16}+0.005 & $-$0.008 & +0.006 & +0.006 & \cellcolor{abWorse!41}$-$0.089 & \cellcolor{abWorse!49}$-$0.053 & $-$0.006 \\
  & U-Net & \cellcolor{abWorse!18}+0.018 & 0.00 & \cellcolor{abWorse!29}+0.016 & +0.001 & \cellcolor{abBetter!13}+0.019 & $-$0.012 & \cellcolor{abWorse!22}+0.019 & \cellcolor{abWorse!37}$-$0.072 & \cellcolor{abWorse!50}$-$0.056 & $-$0.002 \\
 Regularization & No spectral loss & \cellcolor{abBetter!17}$-$0.016 & \cellcolor{abBetter!28}$-$0.26 & \cellcolor{abWorse!18}+0.005 & +0.001 & \cellcolor{abBetter!16}+0.034 & $-$0.012 & \cellcolor{abWorse!22}+0.019 & \cellcolor{abBetter!25}+0.031 & $-$0.003 & +0.005 \\
  & No embedding dropout & \cellcolor{abWorse!12}+0.006 & \cellcolor{abWorse!32}+0.33 & \cellcolor{abWorse!29}+0.016 & +0.001 & \cellcolor{abWorse!14}$-$0.022 & \cellcolor{abWorse!50}+0.022 & +0.007 & $-$0.005 & \cellcolor{abWorse!25}$-$0.012 & $-$0.002 \\
 Loss & Plain L2 & \cellcolor{abWorse!26}+0.045 & \cellcolor{abWorse!22}+0.15 & \cellcolor{abBetter!15}$-$0.003 & \cellcolor{abBetter!18}$-$0.007 & \cellcolor{abWorse!50}$-$0.473 & +0.005 & +0.007 & \cellcolor{abWorse!26}$-$0.033 & 0.000 & $-$0.003 \\
 Embedding product & AlphaEarth & \cellcolor{abWorse!14}+0.010 & $-$0.02 & \cellcolor{abWorse!26}+0.013 & \cellcolor{abWorse!16}+0.005 & \cellcolor{abWorse!11}$-$0.013 & $-$0.013 & $-$0.001 & \cellcolor{abWorse!22}$-$0.022 & \cellcolor{abBetter!18}+0.005 & +0.006 \\
 Multivariate & Separate thermo / wind & $-$0.001 & \cellcolor{abWorse!22}+0.15 & \cellcolor{abWorse!37}+0.028 & \cellcolor{abWorse!15}+0.004 & \cellcolor{abWorse!12}$-$0.015 & \cellcolor{abBetter!45}$-$0.018 & \cellcolor{abWorse!26}+0.028 & \cellcolor{abWorse!16}$-$0.009 & \cellcolor{abWorse!32}$-$0.022 & +0.003 \\
 Temporal context & $T=3$\textsuperscript{$\dagger$} & +0.001 & \cellcolor{abWorse!24}+0.17 & \cellcolor{abBetter!34}$-$0.023 & \cellcolor{abBetter!44}$-$0.054 & \cellcolor{abWorse!30}$-$0.163 & $-$0.010 & 0.000 & \cellcolor{abWorse!25}$-$0.029 & $-$0.004 & \cellcolor{abBetter!41}+0.016 \\
\midrule
\multicolumn{2}{l}{\emph{Probabilistic}} & \multicolumn{4}{c}{fCRPS $\downarrow$} & \multicolumn{3}{c}{SSR $\to 1$} & \multicolumn{3}{c}{Mean $R_{\mathrm{pair}}$ $\uparrow$} \\
\cmidrule(lr){3-6}\cmidrule(lr){7-9}\cmidrule(lr){10-12}
Case & Ablation & ws & dir & $T$ & $T_d$ & ws & $T$ & $T_d$ & ws & $T$ & $T_d$ \\
\midrule
Noise Conditioning & \textbf{Forecast* (reference)} & 0.600\,{\scriptsize$\pm$0.001} & 22.19\,{\scriptsize$\pm$0.03} & 0.881\,{\scriptsize$\pm$0.001} & 1.099\,{\scriptsize$\pm$0.003} & 0.787\,{\scriptsize$\pm$0.005} & 0.947\,{\scriptsize$\pm$0.008} & 0.970\,{\scriptsize$\pm$0.009} & 0.501\,{\scriptsize$\pm$0.005} & 0.566\,{\scriptsize$\pm$0.004} & 0.328\,{\scriptsize$\pm$0.006} \\
 & Global* & \cellcolor{abBetter!14}$-$0.001 & +0.01 & \cellcolor{abWorse!23}+0.004 & \cellcolor{abWorse!31}+0.006 & \cellcolor{abBetter!28}+0.012 & +0.001 & +0.004 & +0.001 & \cellcolor{abWorse!25}$-$0.005 & +0.005 \\
 Stochasticity Source & MC dropout only & 0.000 & \cellcolor{abWorse!50}+0.24 & \cellcolor{abWorse!49}+0.020 & \cellcolor{abWorse!50}+0.016 & \cellcolor{abBetter!50}+0.040 & +0.006 & \cellcolor{abWorse!50}$-$0.018 & \cellcolor{abBetter!21}+0.005 & \cellcolor{abWorse!46}$-$0.018 & \cellcolor{abWorse!50}$-$0.027 \\
 & Noise modulation only & \cellcolor{abWorse!14}+0.001 & +0.01 & 0.000 & +0.002 & $-$0.002 & $-$0.004 & $-$0.006 & $-$0.002 & +0.001 & 0.000 \\
 Loss & No speed CRPS & \cellcolor{abWorse!23}+0.003 & \cellcolor{abWorse!31}+0.09 & \cellcolor{abBetter!17}$-$0.002 & +0.001 & +0.002 & \cellcolor{abWorse!48}$-$0.011 & $-$0.006 & \cellcolor{abWorse!31}$-$0.012 & $-$0.004 & +0.005 \\
 Regularization & No spectral loss & \cellcolor{abBetter!50}$-$0.015 & \cellcolor{abWorse!49}+0.23 & \cellcolor{abWorse!25}+0.005 & \cellcolor{abWorse!40}+0.010 & \cellcolor{abBetter!49}+0.038 & +0.002 & \cellcolor{abWorse!47}$-$0.016 & \cellcolor{abBetter!50}+0.032 & \cellcolor{abWorse!35}$-$0.010 & \cellcolor{abBetter!31}+0.010 \\
 Initialization & Random init (2x steps) & \cellcolor{abWorse!20}+0.002 & \cellcolor{abWorse!49}+0.23 & \cellcolor{abWorse!50}+0.021 & \cellcolor{abWorse!50}+0.016 & \cellcolor{abBetter!28}+0.012 & +0.004 & $-$0.007 & \cellcolor{abWorse!35}$-$0.015 & \cellcolor{abWorse!50}$-$0.021 & \cellcolor{abWorse!25}$-$0.006 \\
 Deep Ensembling & Diff. seeds & \cellcolor{abBetter!27}$-$0.004 & \cellcolor{abBetter!41}$-$0.16 & \cellcolor{abBetter!30}$-$0.007 & $-$0.002 & \cellcolor{abBetter!22}+0.007 & \cellcolor{abBetter!44}+0.009 & +0.002 & \cellcolor{abBetter!33}+0.013 & \cellcolor{abBetter!25}+0.005 & 0.000 \\
 & Diff. models (\textbf{Vernier-Prob}) & \cellcolor{abBetter!30}$-$0.005 & \cellcolor{abBetter!44}$-$0.18 & \cellcolor{abBetter!23}$-$0.004 & $-$0.002 & \cellcolor{abBetter!33}+0.017 & \cellcolor{abBetter!50}+0.012 & \cellcolor{abBetter!41}+0.012 & \cellcolor{abBetter!27}+0.009 & +0.004 & \cellcolor{abBetter!26}+0.007 \\
\bottomrule
\end{tabular}%
}
\caption{\textbf{Ablations on held-out validation stations, 2017-2025} (50k samples, 226k station-hours). The first row of each block is its reference: Vernier-Det, and Forecast*, the weather-modulated probabilistic model. Both references are trained with the spectral loss, as the \emph{+ Spectral} rows of \cref{tab:results}. Other rows show the change from the reference, shaded \colorbox{abBetter!40}{blue} when better and \colorbox{abWorse!40}{pink} when worse if the change exceeds the seed standard deviation (the $\pm$ on the reference rows, over three training seeds). The row below the headings gives units and, for $R_{\mathrm{pair}}$, the number of station pairs 0.1--3\,km apart, the same in both blocks. Each probabilistic row changes one choice of Forecast*, except the deep ensembles. Vernier-Prob combines Forecast* and Global* (32 members each). $^{\dagger}$Scored on leads $\geq 3$\,h.}
\label{tab:ablation}
\end{table*}

\subsection{Stratified Results}
\begin{figure}[t]
  \centering
  \includegraphics[width=\columnwidth]{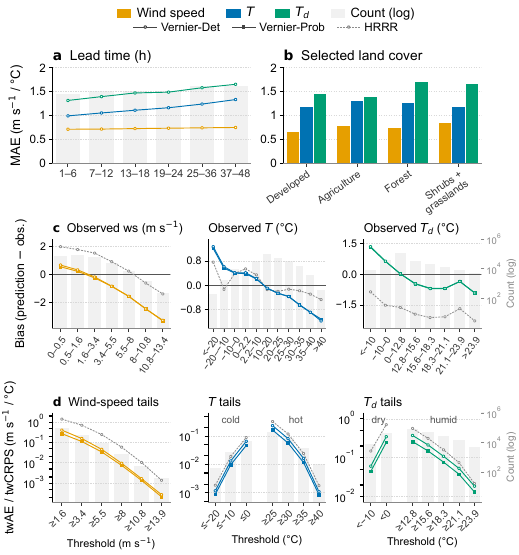}
\caption{\textbf{Stratified results on the 2026 operational test set} (50k samples; \cref{tab:operational}).
\textbf{a,b}: Vernier-Det MAE by lead time and selected land-cover groups.
\textbf{c}: Bias grouped by observed value for HRRR, Vernier-Det, and Vernier-Prob's ensemble mean.
\textbf{d}: Tail scores evaluated over all cases: twAE for deterministic forecasts and fair twCRPS for Vernier-Prob's full ensemble.
Scores shrink for rarer tails; compare models at the same threshold.
Gray bars show station-hour counts per bin, or observed tail-event counts in \textbf{d}. Counts and tail-score axes are logarithmic.}
  \label{fig:stratification}
\end{figure}
On the 2026 operational test set, Vernier-Det's wind-speed MAE remains nearly constant across lead times, while $T$ and $T_d$ errors increase (\cref{fig:stratification}a).
Across the selected land-cover groups (\cref{fig:stratification}b), wind errors are lowest over developed land and highest over shrubs and grasslands; $T$ errors vary little, while $T_d$ errors peak over forest.
Observed-value bins (\cref{fig:stratification}c; \supp~\ref{supp:strat_bias}) reveal similar conditional biases in Vernier-Det and the Vernier-Prob mean: both underestimate strong winds and hot conditions, and overestimate cold conditions.
These bins diagnose conditional errors, but selecting cases by observations can distort forecast rankings---\textit{the forecaster's dilemma}~\cite{lerch2017forecasters}.
We therefore assess tails over all cases at each threshold (\cref{fig:stratification}d), with the threshold-weighted CRPS (twCRPS) for Vernier-Prob's ensemble and the threshold-weighted absolute error (twAE) for the deterministic forecasts, Vernier-Det and HRRR.
The twAE is the twCRPS of a single-valued forecast, so the two scores share units and can be compared directly at each threshold.
Vernier-Prob scores lower than both Vernier-Det and HRRR at every displayed threshold, across strong winds, cold and hot temperatures, and low and high dew points.
The ensemble improves tail prediction even where its mean retains conditional bias.

\subsection{Ablation Studies}
\label{ablations}
\Cref{tab:ablation} evaluates Vernier's design choices on held-out validation stations.
The surface embeddings matter most: without them, wind-speed MAE rises by 0.180\,m\,s$^{-1}$, its $R_{\mathrm{pair}}$ falls by 0.136, and $T$ and $T_d$ also degrade.
The prior and terrain each improve wind direction by about 0.9$^\circ$, and the prior and the surface-anchored decoder each raise wind-speed $R_{\mathrm{pair}}$ by 0.07--0.09.
Our speed--direction loss (\cref{sec:training}) corrects the strong wind-speed underprediction of component-wise L2, whose wind-speed bias is 0.47\,m\,s$^{-1}$ lower.
The spectral loss suppresses surface artifacts (\cref{fig:spectral_loss}) at a cost of 0.016\,m\,s$^{-1}$ in wind-speed MAE, 0.015\,m\,s$^{-1}$ in fCRPS and about 0.03 in $R_{\mathrm{pair}}$. Its tolerance and wavelength bands control this trade-off (\supp~\ref{supp:spectral}).

Probabilistic ablations modify Forecast*, the weather-modulated model (see~\cref{sec:stochastic_prediction}).
Global modulation (Global*) improves wind-speed SSR but worsens $T$ and $T_d$ fCRPS.
Adding latent noise to dropout improves $T$ and $T_d$ fCRPS, whereas adding dropout to noise changes little.
The speed fCRPS term improves wind scores and structure, and deterministic pretraining improves every fCRPS over training from scratch for twice as many steps.
Both deep ensembles improve fCRPS over Forecast*.
Vernier-Prob, the mixture of Forecast* and Global*, also improves SSR and mean $R_{\mathrm{pair}}$ over either component. Pooling seeds of Forecast* gives slightly better temperature fCRPS, but Vernier-Prob's SSR is closer to one for all three variables.

\subsection{Qualitative Analysis}
\label{qualitative}
\begin{figure*}[t]
  \centering
  \includegraphics[width=\textwidth]{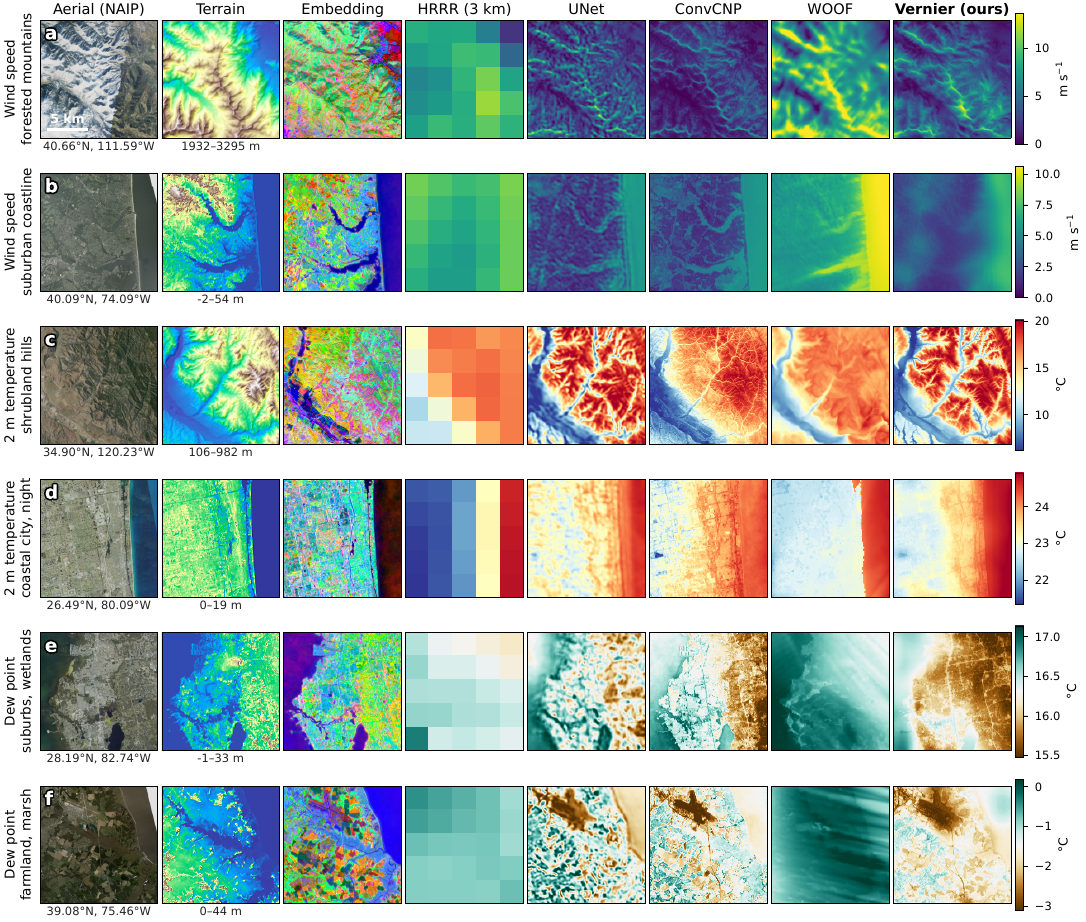}
  \caption{\textbf{Deterministic 100\,m downscaling on test tiles}. 
  Each row shows a 14.4\,km tile for one
    variable. Columns show NAIP imagery, GLO-30 terrain,
    OlmoEarth embeddings (channels 1, 4, 16 as RGB),
    3\,km HRRR forecasts, and 100\,m predictions from
    U-Net, ConvCNP, WOOF, and Vernier-Det trained with the spectral loss. Forecast panels
    share a color scale within each row.
    Wind speed: (\textbf{a}) Wasatch Range, Utah, and
    (\textbf{b}) New Jersey coast. Temperature:
    (\textbf{c}) Santa Barbara County, California, and
    (\textbf{d}) Palm Beach County, Florida. Dew point:
    (\textbf{e}) Tampa Bay, Florida, and
    (\textbf{f}) Delaware Bay coast.}
  \label{fig:qualitative_det}
\end{figure*}

\begin{figure*}[t]
\centering
\includegraphics[width=\textwidth]{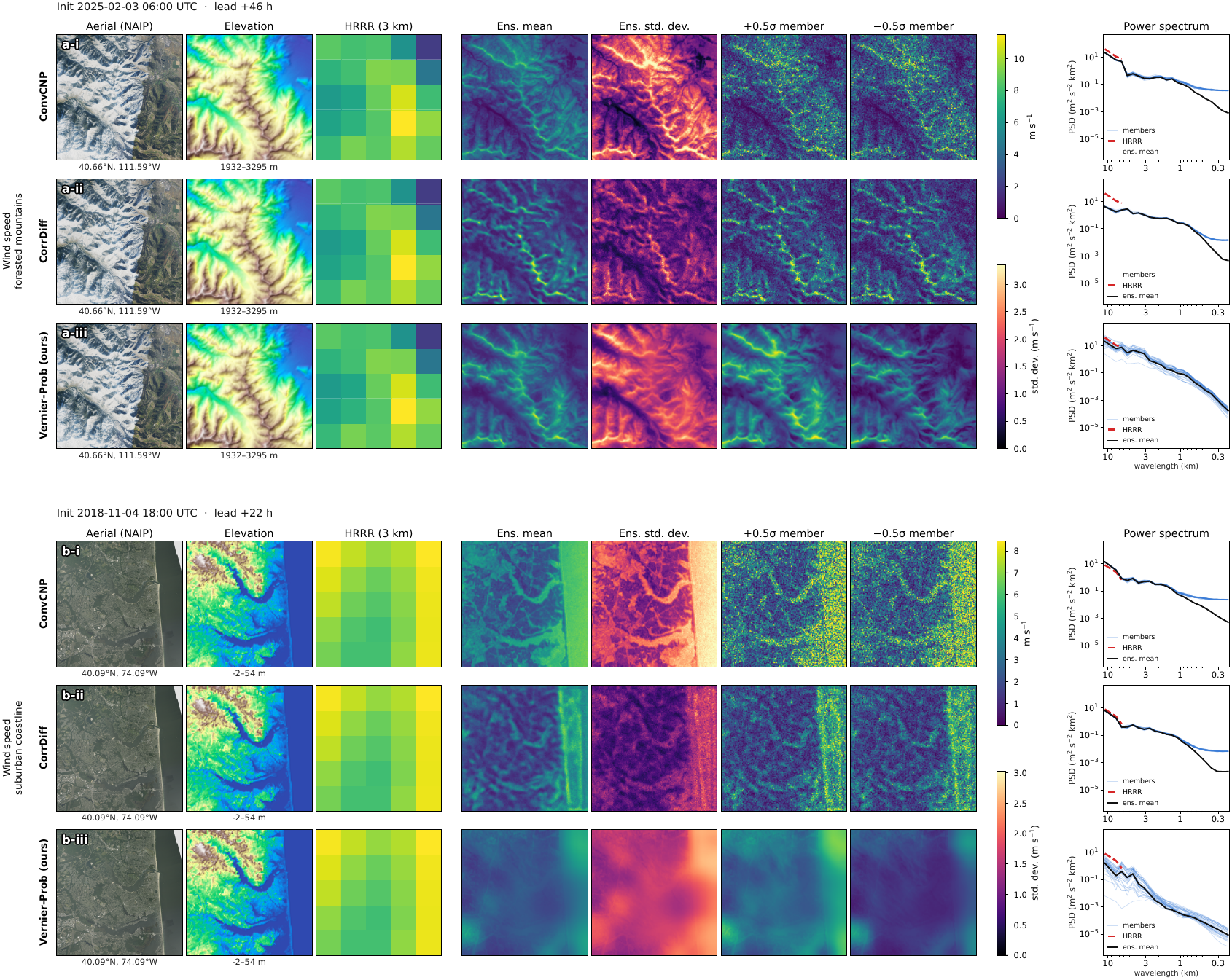}
\caption{\textbf{Probabilistic 100\,m wind-speed downscaling on test tiles}. Each block shows one 14.4\,km tile, (a) Wasatch Range, Utah, and (b) New Jersey coast, with its forecast initialization and lead time above; rows (i)--(iii) show ConvCNP, CorrDiff, and Vernier-Prob (ours), each with 32 ensemble members. Columns show NAIP imagery (tile centroid below), GLO-30 elevation (range below), the 3\,km HRRR forecast, the ensemble mean (mean of the member wind speeds), the ensemble standard deviation, the members whose domain-mean wind speed is nearest the ensemble mean $+0.5\sigma$ and $-0.5\sigma$ (over the members' domain means), and the radially averaged power spectral density (PSD) of each field's Hann-windowed anomaly. PSDs show HRRR (red dashed; bicubically interpolated to 100\,m and shown down to its 6\,km Nyquist wavelength), the ensemble mean (black), and every member (blue), on the same axes in all panels. Within a block, the value panels share one color scale, set by Vernier-Prob's ensemble mean and its two shown members (1st--99th percentiles) and the HRRR range, and the standard-deviation panels share a second scale. Notice the speckling on the generated fields for both CorrDiff and ConvCNP, also visible in the finer scales of the PSD plots.}
\label{fig:prob_qual_ws}
\end{figure*}

\begin{figure*}[t]
\centering
\includegraphics[width=\textwidth]{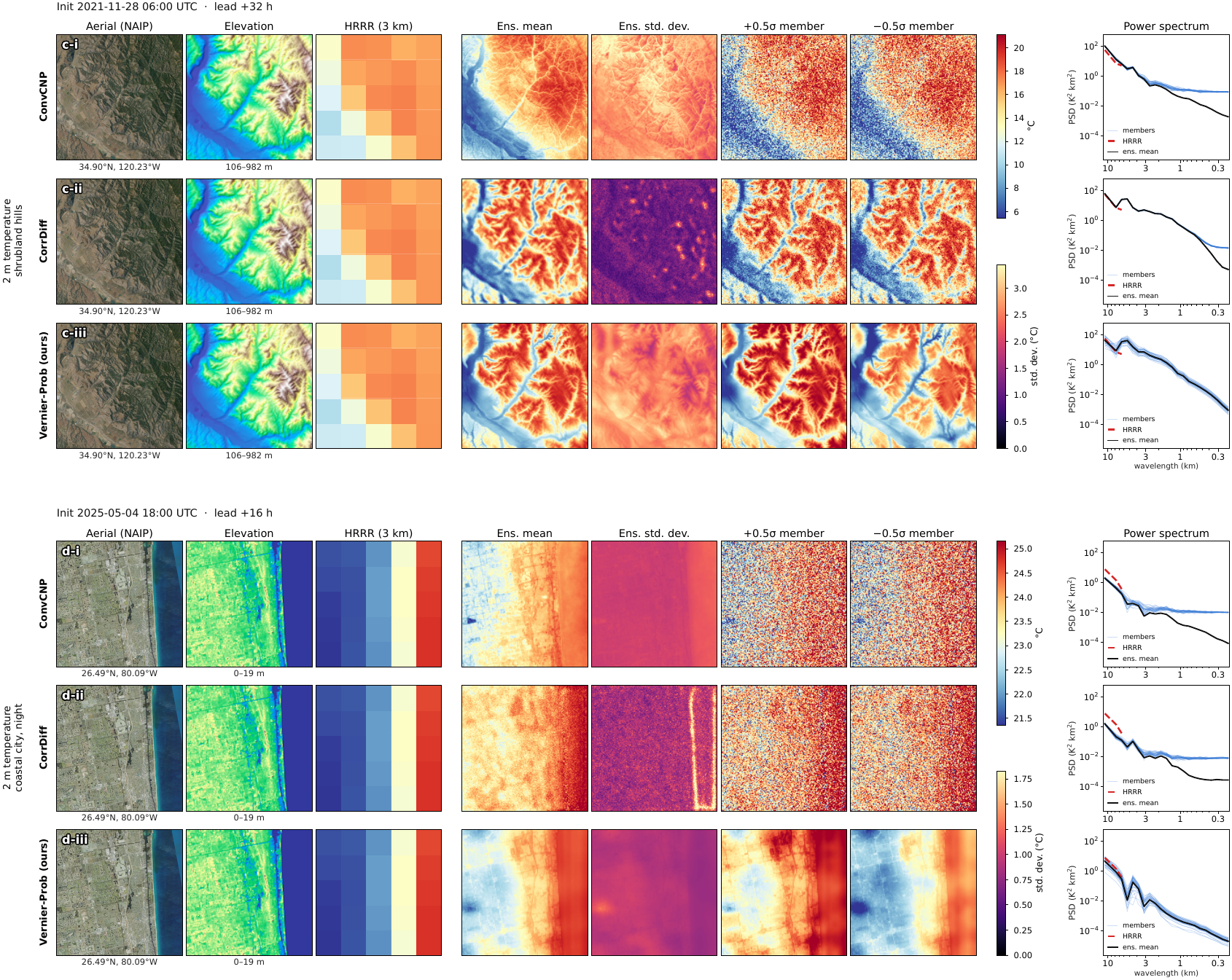}
\caption{\textbf{Probabilistic 100\,m 2\,m-temperature downscaling on test tiles}. Each block shows one 14.4\,km tile, (c) Santa Barbara County, California, and (d) Palm Beach County, Florida, with its forecast initialization and lead time above; rows (i)--(iii) show ConvCNP, CorrDiff, and Vernier-Prob (ours), each with 32 ensemble members. Columns show NAIP imagery (tile centroid below), GLO-30 elevation (range below), the 3\,km HRRR forecast, the ensemble mean, the ensemble standard deviation, the members whose domain-mean temperature is nearest the ensemble mean $+0.5\sigma$ and $-0.5\sigma$ (over the members' domain means), and the radially averaged power spectral density (PSD) of each field's Hann-windowed anomaly. PSDs show HRRR (red dashed; bicubically interpolated to 100\,m and shown down to its 6\,km Nyquist wavelength), the ensemble mean (black), and every member (blue), on the same axes in all panels. Within a block, the value panels share one color scale, set by Vernier-Prob's ensemble mean and its two shown members (1st--99th percentiles) and the HRRR range, and the standard-deviation panels share a second scale. Notice the speckling on the generated fields for both CorrDiff and ConvCNP, also visible in the finer scales of the PSD plots.} 
\label{fig:prob_qual_t2m}
\end{figure*}

\begin{figure*}[t]
\centering
\includegraphics[width=\textwidth]{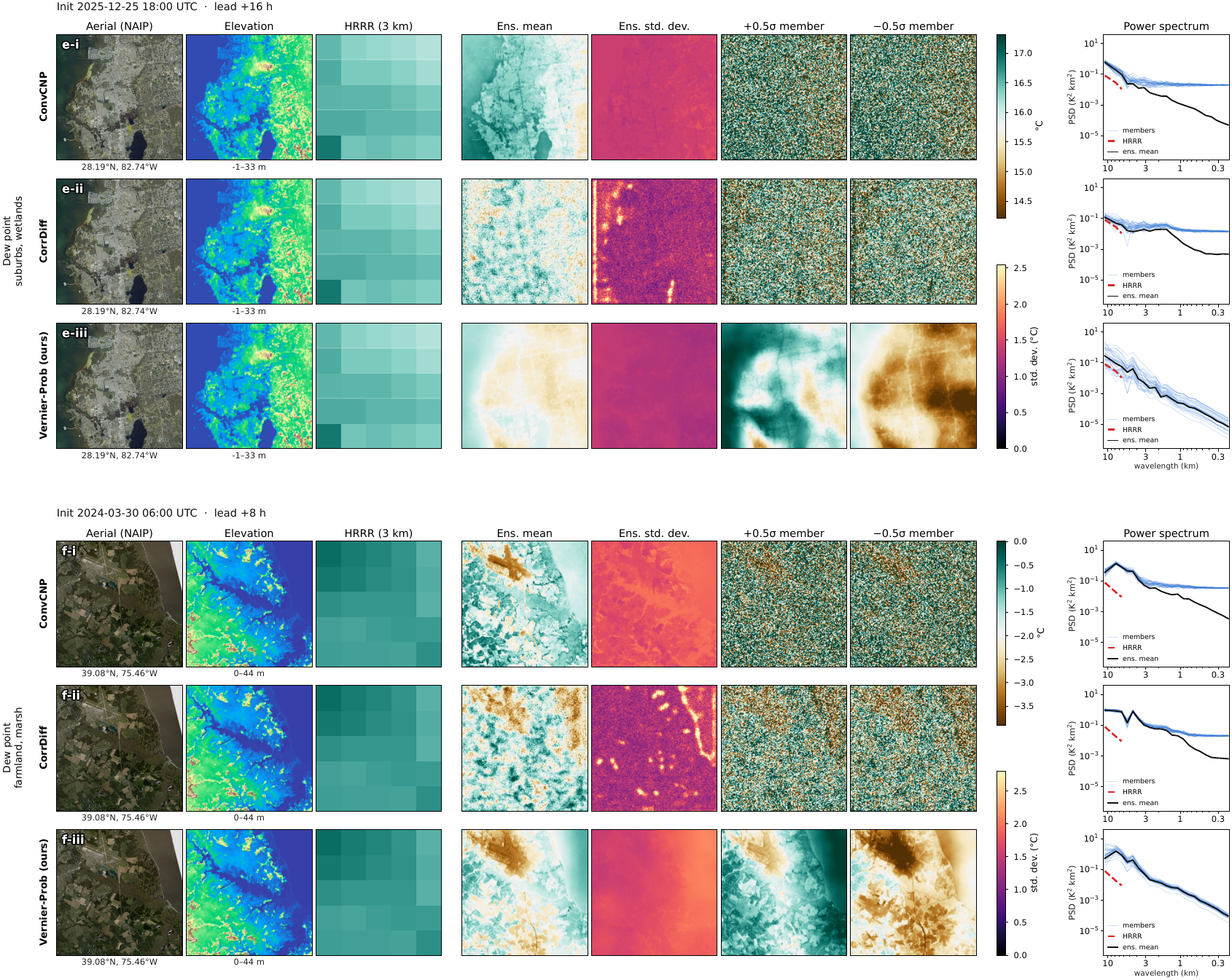}
\caption{\textbf{Probabilistic 100\,m dew-point downscaling on test tiles}. Each block shows one 14.4\,km tile, (e) Tampa Bay, Florida, and (f) Delaware Bay coast, with its forecast initialization and lead time above; rows (i)--(iii) show ConvCNP, CorrDiff, and Vernier-Prob (ours), each with 32 ensemble members. Columns show NAIP imagery (tile centroid below), GLO-30 elevation (range below), the 3\,km HRRR forecast, the ensemble mean, the ensemble standard deviation, the members whose domain-mean dew point is nearest the ensemble mean $+0.5\sigma$ and $-0.5\sigma$ (over the members' domain means), and the radially averaged power spectral density (PSD) of each field's Hann-windowed anomaly. PSDs show HRRR (red dashed; bicubically interpolated to 100\,m and shown down to its 6\,km Nyquist wavelength), the ensemble mean (black), and every member (blue), on the same axes in all panels. Within a block, the value panels share one color scale, set by Vernier-Prob's ensemble mean and its two shown members (1st--99th percentiles) and the HRRR range, and the standard-deviation panels share a second scale. Notice the speckling on the generated fields for both CorrDiff and ConvCNP, also visible in the finer scales of the PSD plots.}
\label{fig:prob_qual_td}
\end{figure*}

\Cref{fig:qualitative_det} compares six held-out test tiles. WOOF captures terrain-driven flow but overestimates wind speed (a, b) and produces flow-aligned dew-point streaks (e, f).
The ML baselines add 100\,m detail, but the U-Net adds blotchy fine-scale noise (a, e, f), and ConvCNP imprints land-cover boundaries (e, f) and thin lines along drainages (c).
Vernier captures terrain-driven winds (a), cold-air pooling in valleys (c), and urban and coastal dew-point patterns (e, f).
Its spectral loss suppresses wind artifacts but does not regularize $T$ or $T_d$.

In the probabilistic examples (\cref{fig:prob_qual_ws,fig:prob_qual_t2m,fig:prob_qual_td}), Vernier-Prob members differ both in overall level and in local pattern while keeping fine-scale power, and their power spectra vary appreciably.
ConvCNP's mean is sharp and coherent, but its members are noisy, with a flat, white-noise-like plateau at fine scales in their spectra, consistent with their low member $R_{\mathrm{pair}}$ (\cref{tab:results,tab:extreme}).
CorrDiff's members are speckled as well, most clearly for $T$ and $T_d$ (d--f). The speckle partly averages out in the mean but leaves a similar plateau in the member spectra.
The members of both baselines have nearly identical spectra.

\subsection{Searchlight}
\label{sec:searchlight_results}
\begin{figure*}[t]
  \centering
  \includegraphics[width=0.84\textwidth]{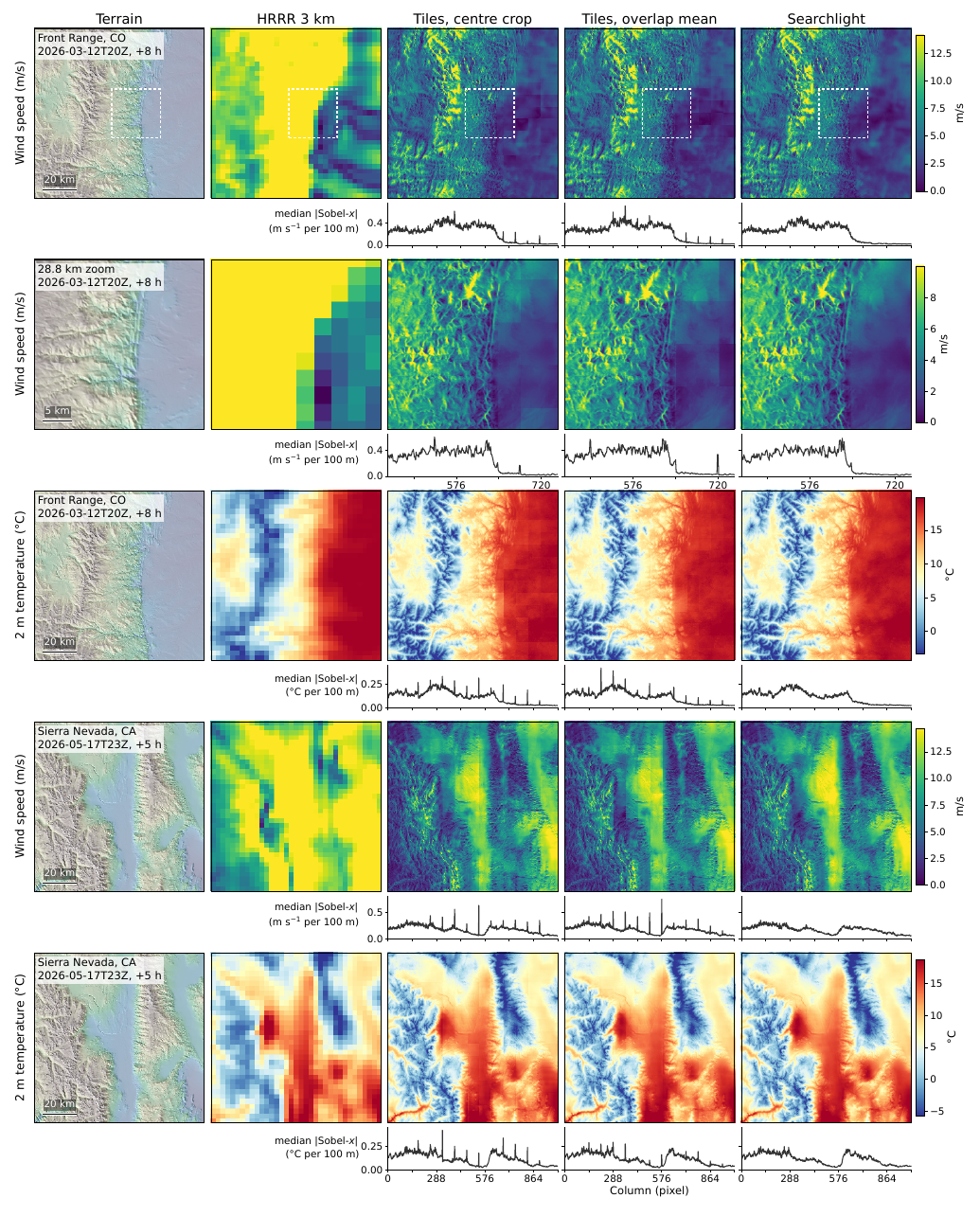}
  \caption{\textbf{Searchlight versus tile composites (Vernier-Det).} Two $100.8\,\mathrm{km}$ domains ($1{,}008\times1{,}008$ pixels at $100\,\mathrm{m}$), one hour each: the Colorado Front Range (2026-03-12 20\,UTC, lead $8$\,h, a downslope windstorm) and the Sierra Nevada with Owens Valley (2026-05-17 23\,UTC, lead $5$\,h). Rows: 10\,m wind speed and 2\,m temperature; the second row zooms into the $28.8\,\mathrm{km}$ window outlined in the first (the foothills east of the Continental Divide), with HRRR's 3\,km cells drawn as blocks. Columns: terrain; the HRRR forecast on its 3\,km cells; $144$-pixel tiles at a $72$-pixel stride, each predicted on its own and composited by center crop; the same tiles with overlaps averaged; Searchlight, one forward pass over the domain. The physics prior is the same tile-composited prior in all three. Color scales are shared along a row and set by the $100\,\mathrm{m}$ fields, so HRRR's higher wind speeds saturate; the zoom has its own. Under each composite: the median over its rows of the absolute Sobel-$x$ gradient, by column; the center crop jumps at every tile cut (columns $108, 180, \dots, 900$), the overlap mean at every tile edge (multiples of $72$ pixels), Searchlight at neither. In the zoom, the center crop changes value at each cut, $7.2\,\mathrm{km}$ apart, and averaging the overlaps turns the cuts into bands.}
  \label{fig:supp_searchlight_det}
\end{figure*}

\begin{figure*}[t]
  \centering
  \includegraphics[width=\textwidth]{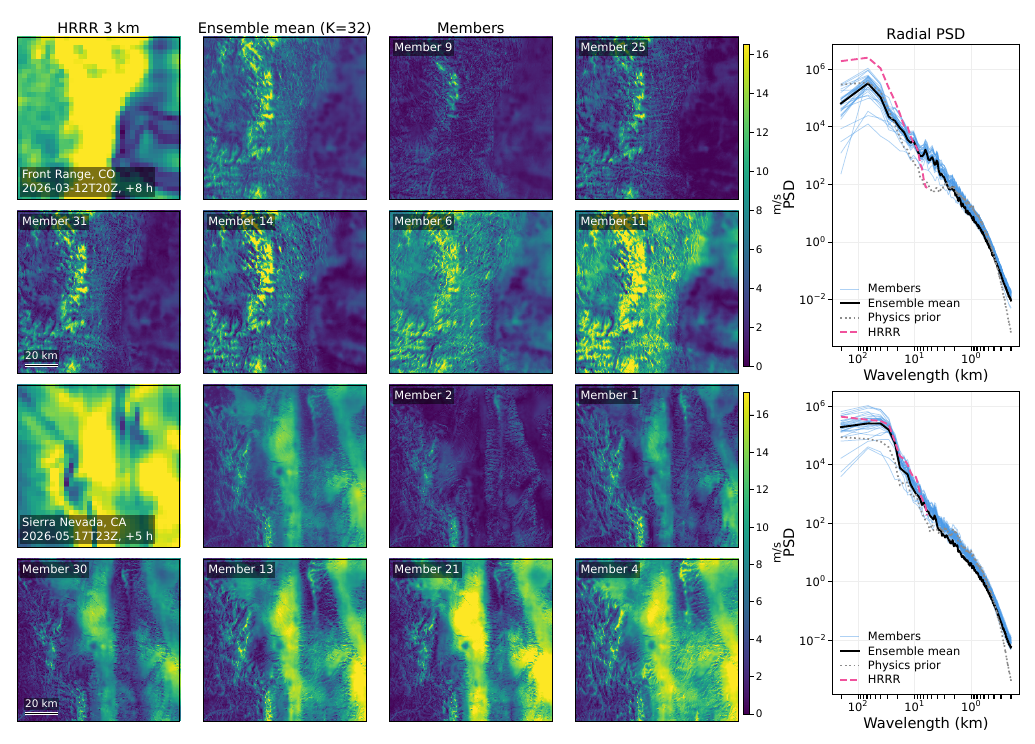}
  \caption{\textbf{Searchlight ensemble members (Vernier-Prob).} The same two domains and hours as \cref{fig:supp_searchlight_det}, wind speed, $K=32$ members ($16$ from each of the two parts of the operational ensemble) inferred with Searchlight. Panels: the HRRR forecast, the ensemble mean, six members spanning the ensemble from the calmest to the windiest domain mean, and the radially averaged power spectral density (PSD) of the Hann-windowed anomaly of every member (blue), the ensemble mean (black), the physics prior (dotted gray) and HRRR interpolated to $100\,\mathrm{m}$ (pink dashed, shown down to its $6\,\mathrm{km}$ Nyquist wavelength).}
  \label{fig:supp_searchlight_prob}
\end{figure*}

Searchlight (\cref{sec:searchlight}) runs Vernier over a domain much larger than its training tile in one forward pass, with no tile boundaries.
\Cref{fig:supp_searchlight_det} compares it with tiling on two $100.8\,\mathrm{km}$ domains on windy hours over mountains in 2026, the Colorado Front Range during a downslope windstorm and the Sierra Nevada with Owens Valley, with the same checkpoint and the same physics prior.
Tiles predicted on their own and composited by center crop carry a lattice of seams in wind speed and temperature where neighboring tiles disagree; averaging their overlaps blurs the seams into bands but does not remove them. Searchlight has none.
With Vernier-Prob (\cref{fig:supp_searchlight_prob}), each member is one coherent wind field across the domain. The members differ mostly at scales of several kilometers and more; below about 1\,km their spectra converge, because much of the finest structure follows the terrain and is shared by all members.
On the test tiles, Searchlight changes MAE, fCRPS and raw $R_{\mathrm{pair}}$ by at most 0.01 (de-meaned $R_{\mathrm{pair}}$ by up to 0.02) and narrows the ensemble slightly (\cref{tab:operational}); the domain-size scaling, the prior compositing and the cost are in \supp~\ref{supp:searchlight}.

\section{Discussion and Limitations}\label{sec:discussion}

\paragraph{Station networks and exposure.}
Station networks site their sensors differently: airports follow the 10\,m open-exposure standard, while CWOP amateur-run stations and many MesoWest stations are lower or sheltered and read weaker winds.
These two networks carry most of the wind station-hours (89\% in the 2026 operational test set), and HRRR's wind error differs strongly between networks, so a single 100\,m field must serve stations with very different exposures.
Training on every network still works best: restricting the wind loss to standard-exposure networks hardly improves the airports, makes the wind too fast at the networks left out, and weakens the structure between nearby stations.
Exposure also explains why station calibration (MOS) is a strong baseline for wind, since a station's siting acts as a fixed multiplier that years of its own record pin down.
Vernier has no per-station parameters, yet it is close to per-station MOS for wind speed, improves further on temperature and dew point, and, unlike MOS, follows how the contrast between neighboring stations changes with the weather.
It therefore learns both bias correction and downscaling.
Its gains still vary by network, smallest at airports and largest at sheltered CWOP stations.
Modeling exposure explicitly, for example with a per-network observation operator, could let Vernier predict a standard-exposure field while still matching each network.

\vspace{-2mm}
\paragraph{Evaluating 100\,m structure.}
Station pairs let us measure local structure at unseen stations without dense ground truth, and they test exactly what a coarse forecast cannot provide, since nearby stations usually share an HRRR cell.
The pair metric, however, mixes persistent contrasts between sites, including siting and exposure, with contrasts that change with the weather, so a method that only copies fixed site differences, such as per-station MOS, can also score well.
We therefore also report de-meaned correlations, which remove each pair's mean contrast.
These are much lower, especially for wind speed and dew point, so weather-dependent local contrasts remain the hardest part of the problem, although Vernier keeps a clear lead.
Dense references do not settle the question either: WRF carries its own biases, and ECOSTRESS observes land-surface skin temperature rather than 2\,m air temperature.
Still, the pair metrics are informative.
Averaging Vernier's output to a 3\,km grid barely changes its MAE but lowers its wind-speed $R_{\mathrm{pair}}$ from 0.47 to 0.14, so the 100\,m detail carries information that point errors alone would miss.

\vspace{-2mm}
\paragraph{Higher-fidelity physical priors.}
Our priors, a mass-consistent wind solve and lapse-rate adjustments for $T$ and $T_d$, are simple enough to compute on the GPU during training.
Stations alone cannot teach all the physics needed: they are sparse, unevenly spread across terrain and weather, and biased towards certain human infrastructural needs. Physical models therefore remain a reliable route for generalization across regions and weather regimes.
A higher-fidelity prior would give a better starting point.
It could also serve as a structural target that constrains the full 2-D field rather than only the station locations, leaving the model less room to create structure that no observation supports.
It would also strengthen the spectral loss.
Our prior omits surface roughness, so the reference for the finest wind scales is extrapolated from coarser ones, and the loss does not constrain $T$ or $T_d$.
A prior that represents the actual near-surface flow, including its response to stability and wind direction, would give a better reference and help the model learn how surface effects change with the weather.

\vspace{-2mm}
\paragraph{Ensembles and extremes.}
Vernier's ensembles improve threshold-weighted tail scores over HRRR (\cref{fig:stratification}d), and Vernier-Prob's fCRPS stays below HRRR's MAE in every bin of observed wind speed.
Both Vernier models nevertheless underestimate the strongest observed winds and keep conditional biases at temperature extremes.
One reason is the training distribution: most station-hours have light winds and moderate temperatures and dew points, so rare conditions contribute little to the loss.
Weighting the loss by wind-speed stratum did not help in our experiments; sampling training cases to balance weather regimes, or losses aimed at the tails, may be more effective.
Better tail scores also do not establish reliable exceedance probabilities, and Searchlight slightly narrows ensemble spread, so the calibration of extreme-event probabilities needs further study.

\vspace{-2mm}
\paragraph{Scale and variables.}
Vernier currently covers the HRRR domain.
Its Earth-observation embeddings are available globally, so a global version mainly needs global forecasts, such as IFS or WeatherNext~3, as inputs, and must cope with station networks whose density and quality vary substantially between regions.
Other surface variables, precipitation in particular, will need dedicated targets, priors and probabilistic modeling, since precipitation is intermittent, heavy-tailed and sparsely observed.

\section{Conclusion}\label{sec:conclusion}
We presented Vernier, which learns 100\,m forecasts of surface wind, temperature and dew point from 3\,km HRRR forecasts, supervised only by sparse weather stations.
Vernier learns a residual correction to a physics-based prior, which gives plausible structure where no stations constrain the model.
Its multi-stream encoder and surface-anchored decoder combine the forecast with Earth-observation embeddings of the land surface, and the same architecture serves both deterministic and probabilistic forecasting.
Searchlight extends each ensemble member to regional domains at inference, without retraining, producing seamless fields over large domains.
Vernier generalizes to unseen stations in an unseen year, better reproduces differences between nearby stations than the baselines, and outperforms the compared operational models in wind-speed and temperature error. Vernier also generates coherent ensemble members which competing probabilistic downscaling methods are unable to recover from sparse station targets alone.

\section*{Acknowledgments}
We thank Professor Cliff Mass and his group at the University of Washington for sharing the 2026 UW-WRF forecasts~\cite{mass2003regional,uwwrf} (\url{https://a.atmos.washington.edu/wrfrt/gfsinit.html}; domain d04, 1.33\,km) used in our Pacific Northwest comparison. We thank Professor Niklas Boers for hosting a research visit at the Potsdam Institute for Climate Impact Research in Potsdam, Germany, in support of this collaboration. Finally, we extend our gratitude to the European Space Agency and the European Union for the Copernicus DEM GLO-30 (DOI 10.5270/ESA-c5d3d65), from which our terrain fields were derived.

\section*{Code}
The Vernier code is hosted on \href{https://github.com/allenai/Vernier}{\mbox{\github GitHub}}.
Model weights will be available on
\mbox{\raisebox{-1.5pt}{\includegraphics[height=1.05em]{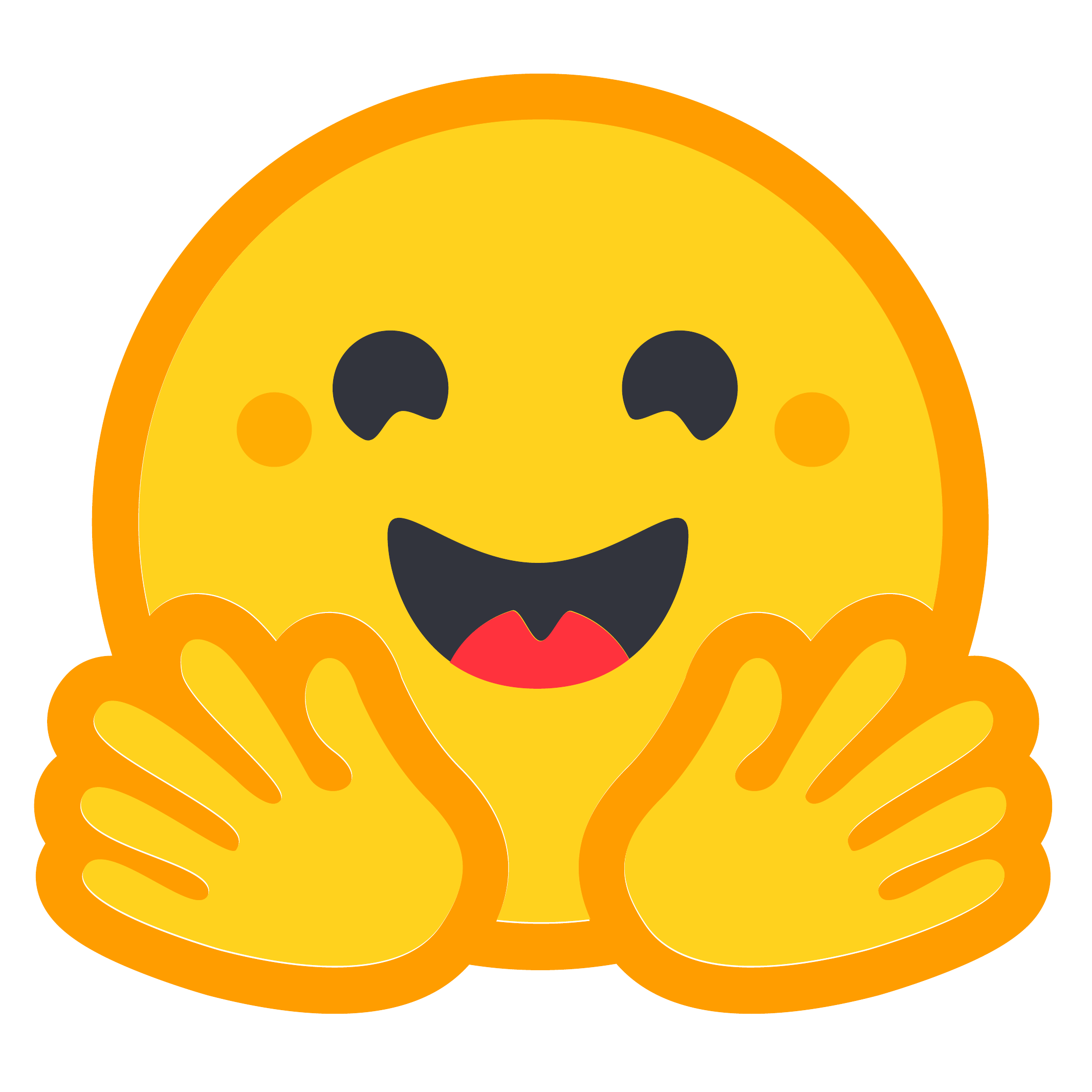}}\,Hugging Face}.

\clearpage
\bibliographystyle{ieeenat_fullname}
\bibliography{main}

\renewcommand{\onecolumn}{}
\renewcommand{\twocolumn}[1][]{\FloatBarrier\ifblank{#1}{}{\begin{figure}[tbp]#1\end{figure}}}
\clearpage
\phantomsection
\addcontentsline{toc}{section}{Appendix}
\begin{center}
  {\sffamily\bfseries\LARGE\color{colorone} Appendix\par}
\end{center}
\vspace{0.5em}
\ifdefined\supptabbox\else\newsavebox{\supptabbox}\fi
\providecommand{\supptabfit}[2]{\sbox{\supptabbox}{#2}%
  \ifdim\wd\supptabbox>#1\relax\resizebox{#1}{!}{\usebox{\supptabbox}}\else\usebox{\supptabbox}\fi}
\newcounter{suppref}
\makeatletter
\providecommand{\suppreflist}{1}  %
\providecommand{\citesupp}[1]{[\def\@csep{}\@for\@sk:=#1\do{\@csep S\ref{suppref:\@sk}\def\@csep{, }}]}
\makeatother
\providecommand{\etal}{\emph{et al.}\xspace}  %

\appendix

Here we present additional methodological and experimental details, along with important additional results that would not fit in the main text. The important sections include \cref{supp:structure}, which supports the structural metrics we introduce; \cref{supp:mos}, which uses Model Output Statistics (MOS), an operational station de-biasing method, as a very strong set of baselines to show that Vernier does not just de-bias stations but also learns inter-station structure; and \cref{supp:searchlight}, which details Searchlight, including detailed qualitative figures on generated fields with quantitative support for seamless outputs.

\section{Datasets and Splits}
\label{supp:data}

\subsection{HRRR domain and inputs}
\label{supp:hrrr}

\paragraph{The HRRR model.}
The High-Resolution Rapid Refresh (HRRR) is NOAA's operational weather model for the contiguous United States and adjacent parts of Canada and Mexico~\cite{dowell2022hrrr}.
It is rerun every hour: each cycle assimilates the latest observations, including radar reflectivity, and starts a new forecast on a 3\,km grid, with lateral boundaries from the coarser Rapid Refresh model.
Every hourly cycle forecasts 18 hours ahead; the 00, 06, 12 and 18 UTC cycles extend to 48 hours (36 hours from July 2018 to December 2020, 18 hours before that).
At 3\,km, HRRR resolves thunderstorms explicitly, but its terrain and land surface are smoothed to that grid, so valleys, ridges, coastlines and land-cover contrasts smaller than a few kilometers are missing from its near-surface fields.
These are the details that Vernier adds at 100\,m.

\paragraph{Forecast inputs.}
We pair HRRR forecasts with surface observations from January 2017 through September 2026.
HRRR uses a $1799\times1059$ Lambert conformal grid at 3\,km spacing~\cite{dowell2022hrrr}.
Its projection has central longitude $97.5^\circ$W, standard latitude $38.5^\circ$N, and spherical radius 6,371,229\,m.
We use forecasts initialized at 00, 06, 12, and 18 UTC.
Hourly leads at these cycles extend to 18, 36, or 48 hours under HRRR versions 2, 3, and 4 (until July 2018, until December 2020, and since), respectively.
The sampler enforces these limits and excludes missing forecast fields using an archive coverage mask.
Observations are matched to forecast valid time, defined as initialization plus lead.
Inputs comprise 2\,m temperature and dew point, 10\,m wind components, cloud cover, skin temperature, surface pressure, and soil moisture.
Grid-relative winds are rotated to geographic east and north before conditioning or scoring.
Hourly analyses support observation quality control.
\Cref{fig:supp_split_map} shows the domain together with the spatial split cells (\supp~\ref{supp:splits}).

\subsection{Surface covariates and registration}
\label{supp:surface}
Copernicus GLO-30 elevation~\cite{copernicus2021glo30} is resampled to 100\,m and accompanied by eastward, northward, and vertical surface-normal components.
OlmoEarth v1.3 supplies annual surface embeddings~\cite{tseng2026olmoearth}.
We retain the first 64 of 128 dimensions, average from 10\,m to 100\,m, and renormalize the resulting vectors to unit length.
AlphaEarth's 64-dimensional embeddings provide an ablation~\cite{brown2025alphaearth}.
For January--August, the loader selects the previous year's embeddings; for September--December, it selects the current year's.
An annual embedding summarizes imagery from its whole calendar year, so it also describes the surface after the valid time.
For a valid time in March, for example, the current year's embedding would already include that year's spring green-up, summer growth and harvests, and any land-cover change up to December, which a forecast issued in March cannot know.
Switching to the current year only from September 1, when most of its imagery has been acquired, limits this look-ahead while keeping the embedding recent.
The selected year is clamped to the available 2017--2025 range, so all 2026 samples use 2025 embeddings.

Fine-resolution inputs share a fixed 100\,m lattice in the HRRR projection.
For each crop, coarse conditioning fields are registered to a $5\times5$ grid with 2.88\,km spacing.
The fine-grid HRRR reference is interpolated directly from native forecast coordinates, avoiding a second interpolation through this conditioning grid.

\subsection{Station networks and quality control}
\label{supp:stations}

\paragraph{Station observations.}
We merge NOAA MADIS METAR, mesonet, and Climate Reference Network observations~\cite{madis} with HadISD v3.4.3.2025f~\cite{dunn2016hadisd}.
Maritime records remain in the archive but are excluded from our splits.
Reports passing source-level quality checks are aggregated over the half-open interval $[t-30\,\mathrm{min},t+30\,\mathrm{min})$ around each UTC hour.
Wind speed and direction are converted to eastward and northward components before averaging.
We use these unweighted hourly means; the archive also stores a temporally weighted alternative.
Missing quantities remain missing, including wind components at stations that report speed without direction.
Station identifiers incorporate source and location information to distinguish reused identifiers.
Available metadata do not support a uniform correction for sensor height.

\paragraph{Quality control.}
Additional filters remove implausible values, stuck sensors, duplicate sites, and large discrepancies from HRRR analyses.
Plausibility bounds are 200--330\,K for temperature, 180--310\,K for dew point, and 75\,m\,s$^{-1}$ for wind speed.
Dew point exceeding a valid temperature by more than 0.5\,K is rejected.
Runs of at least 24 identical nonzero hourly values are masked, and constant nonzero records are excluded for the affected variable.
Wind screening, adapted from WIND-Bench~\citesupp{bazlen2026windbench}, identifies sheltered or inactive sensors with a 95th-percentile speed below 1\,m\,s$^{-1}$.
It also rejects persistently elevated winds when a station's 10th-percentile speed exceeds its climate-region median by six median absolute deviations, with a threshold floor of 5\,m\,s$^{-1}$.
These wind tests operate by station and calendar year, using whole-record statistics when fewer than 720 valid hours are available.
A HadISD record whose airport code matches a METAR station within 3\,km is dropped in favor of the METAR record.

A separate gross-error filter compares observations with bicubically interpolated HRRR analyses.
It rejects discrepancies exceeding six global residual standard deviations, treating wind discrepancy as a vector magnitude and rejecting both components together.
The reference spread is computed from observation-minus-analysis residuals, rather than absolute weather fields.
These decisions are stored as per-variable hourly masks.
The same masks govern sampling eligibility and station targets across training, validation, and test.

\paragraph{Impact and coverage.}
Quality control removes little data except for wind.
Duplicate airport records, range checks, stuck sensors and the analysis cut each remove at most 0.15\% of the hourly values of any variable.
The sheltered-sensor test removes 6.7\% of wind hours (6,534 stations in at least one year), and the elevated-wind test removes less than 0.01\%.
After quality control, 59,535 land stations in the HRRR domain contribute 2.0 billion temperature, 1.7 billion dew-point and 1.7 billion wind station-hours between January 2017 and September 2026 (\cref{tab:supp_station_stats}); the 77,460 stations of the main text count every land station identifier in the merged catalog, before quality control.
Volunteer and regional mesonets supply most of them; airports (METAR) supply under 5\% of the stations but 10--12\% of the station-hours.
HadISD adds few stations once airport duplicates are removed, and its v3.4.3.2025f release ends in August 2025.
\Cref{fig:supp_station_snapshot} shows the stations reporting on one day of the 2026 test period.

\subsection{Splits}
\label{supp:splits}

\paragraph{Spatial and temporal partitions.}
The spatial benchmark assigns 3,554 eligible $50\times50$\,km cells to training, validation, and test, with 2,844, 355, and 355 cells, respectively (\cref{fig:supp_split_map}a).
Assignment uses seed 9 and stratifies by station availability, terrain relief, and surface-embedding diversity.
Stations are disjoint across partitions, while input footprints may cross cell boundaries; only stations belonging to the selected split supply targets.
These partition cells are distinct from the smaller training crops.

The same cells define two holdouts (\cref{fig:supp_split_map}b).
In the \emph{spatial holdout}, all three partitions share the 2017--2025 period, so validation and test stations are unseen in training but their years are not.
In the \emph{spatio-temporal holdout}, models train on the training cells over 2017--2025, and the validation and test cells are additionally held out in time, covering January--September 2026.
\Cref{tab:results} reports the spatio-temporal test set, and \supp~\ref{supp:ecoregion_splits_results} reports the spatial holdout.

For the operational comparison, training uses 2017--2025 over the full footprint and testing uses January--September 2026.
The operational runs' 2024 validation stream overlaps training and serves as a monitoring set, not an independent temporal holdout.
Operational comparisons use fixed sampled tiles so forecasts are evaluated at matching locations, valid times, initializations, and leads.

\subsection{PNW domain}
\label{supp:pnw}
\begin{figure}[h]
  \centering
  \includegraphics[width=0.5\columnwidth]{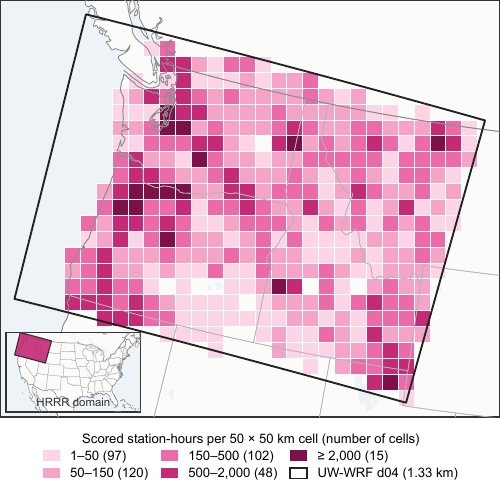}
  \caption{\textbf{PNW domain.} The black outline is UW-WRF's innermost domain (d04, 1.33\,km), drawn in HRRR's Lambert conformal projection as in \cref{fig:supp_split_map}; the inset locates it in the HRRR domain. Shaded are the 382 $50\times50$\,km cells that hold stations scored in the Pacific Northwest evaluation of \cref{tab:operational}, colored by their station-hours (number of cells per class in the legend). Cells along the edge can extend past the outline, since only their stations' tiles must lie inside d04.}
  \label{fig:supp_wrf_domain}
\end{figure}

UW-WRF covers only the Pacific Northwest (\supp~\ref{supp:uwwrf}), so the Pacific Northwest block of \cref{tab:operational} uses its own test set inside UW-WRF's innermost domain (d04, 1.33\,km), which spans Washington, Oregon, Idaho and western Montana (\cref{fig:supp_wrf_domain}).
The set keeps every station in the contiguous United States whose 14.4\,km tile lies inside d04, away from its 10-cell lateral boundary band.
As in the operational test, stations are not limited to the test partition of the spatial split.
We draw 50k samples from January to September 2026 with the sampler of the HRRR-domain test set, restricted to UW-WRF's 00 and 12\,UTC initializations and its 12--24\,h leads, so every model is scored at the same stations, valid times, initializations and leads.
The samples cover 3,062 stations in 382 of our $50\times50$\,km cells, and station-hours are unevenly spread, from fewer than 50 in 97 cells to more than 2,000 in 15.
We exclude about 135 co-located RAWS units at a fire-equipment depot in Boise, Idaho, which would otherwise contribute nearly a third of the station-hours.

\subsection{Ecoregion Holdouts}
\label{supp:ecoregion_splits}
\begin{figure}[t]
  \centering
  \includegraphics[width=\columnwidth]{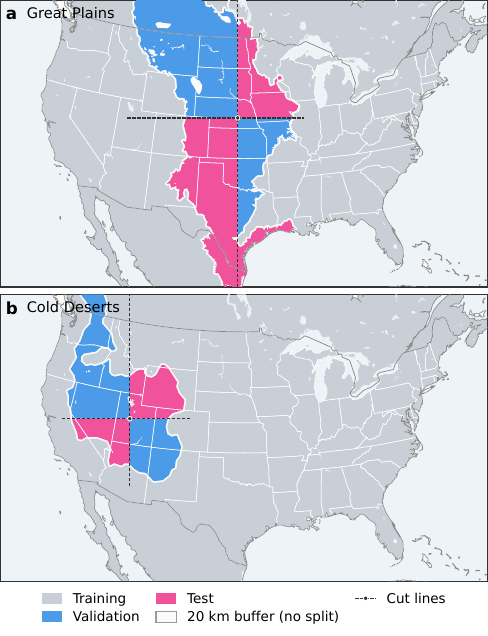}
  \caption{\textbf{Ecoregion holdouts.} \textbf{a}, Great Plains; \textbf{b}, Cold Deserts, in HRRR's Lambert conformal projection. Dashed lines cut each region into quarters: the northwest and southeast quarters form the validation part (blue), and the northeast and southwest quarters the test part (pink). Training (gray) uses all other land; a 20\,km buffer around the region (white outline) belongs to no split.}
  \label{fig:supp_ecoregion_splits}
\end{figure}

The spatial split holds out stations, but every held-out station still has training stations nearby.
To test transfer to regions with no training stations, we also hold out two ecoregions in turn (\cref{fig:supp_ecoregion_splits}), using the ecoregions of North America defined by the Commission for Environmental Cooperation (CEC).
The \emph{Great Plains} is CEC Level I region 9.0, a large and mostly flat region spanning semi-arid and humid climates.
The \emph{Cold Deserts} is CEC Level II region 10.1 (the Great Basin, Colorado Plateau, Wyoming Basin, and the Snake and Columbia plateaus); we join its lobes into one region, which also holds out the mountain ranges between them, such as the Wasatch and Uinta Mountains.

A vertical and a horizontal line cut each region into four quarters: the northwest and southeast quarters form the validation part, and the northeast and southwest quarters the test part.
The cut point balances station-hours between the two parts.
Training uses all other land in the HRRR domain except a 20\,km buffer around the held-out region, which belongs to no split, so no training station lies within 20\,km of the region.
Because the buffer is wider than a 14.4\,km crop, training crops hardly reach into the held-out region.
Only stations are held out: HRRR, terrain, and surface embeddings remain inputs everywhere, as in deployment.
All partitions cover 2017--2025, so these holdouts are spatial only, and the physics-prior constants (\cref{tab:prior_params}) are not refitted.
The Great Plains split has 35,442 training, 2,860 validation, and 2,983 test stations; the Cold Deserts split has 36,332, 2,866, and 2,228.
Each test part is scored on 50k fixed samples (\supp~\ref{supp:ecoregion_splits_results}).

\subsection{Station and Tile Sampling}
\label{supp:sampling}

\paragraph{Anchor and forecast sampling.}
Eligibility requires all four weather variables to pass quality control for $k=3$ consecutive hours at the anchor station.
Candidate windows must lie within the split's allowed dates, avoid excluded intervals, and have compatible HRRR coverage.
Anchors whose full input footprint falls outside a required data source are excluded.
Let $n_s$ count the remaining eligible windows at station $s$.
We sample
\begin{equation}
    p(s)=\frac{n_s^\alpha}
    {\sum_{j:n_j>0}n_j^\alpha},
    \qquad \alpha=0.5.
    \label{eq:appendix_station_sampling}
\end{equation}
This lies between uniform station sampling ($\alpha=0$) and sampling proportional to available windows ($\alpha=1$).
It reduces the dominance of long records without making geographic sampling uniform.

We then select an eligible window uniformly and use its final hour as the prediction time.
The model consumes and predicts this single hour; $k=3$ controls eligibility, not input sequence length.
At the selected hour, we sample uniformly among admissible HRRR initialization--lead pairs.
The initialization is therefore determined by the sampled lead and valid time.
This single-hour configuration admits leads starting at one hour, subject to archive availability and version limits.

\begin{figure}[t]
  \centering
  \includegraphics[width=\columnwidth]{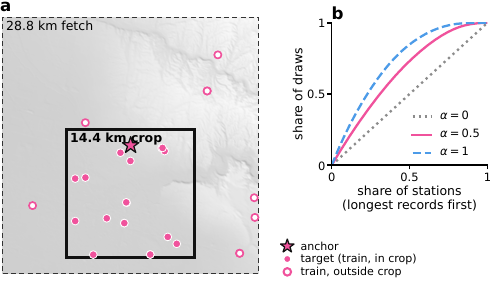}
  \caption{\textbf{Sampling example.} \textbf{a}, One draw near Bakersfield, California: the inputs are fetched over a 28.8\,km region around the anchor station (star), from which the 14.4\,km crop that the model sees and predicts is cut. Training stations inside the crop are targets (filled); those outside it are not used. \textbf{b}, Share of draws against share of training stations, ranked by record length $n_s$, for three values of $\alpha$; $\alpha=0.5$ lies between uniform station sampling ($\alpha=0$) and sampling every eligible window equally ($\alpha=1$).}
  \label{fig:supp_sampling_example}
\end{figure}

\paragraph{Crop construction and supervision.}
We fetch a $28.8\times28.8$\,km region around the anchor and select a $14.4\times14.4$\,km crop, yielding $144\times144$ output pixels.
Crop offsets are random during training and centered during evaluation, providing jitter augmentation.
Tiles are thus selected through anchors and crop offsets, rather than uniformly over land.
All valid stations from the same split inside the final crop contribute targets.
Observations sharing a pixel are averaged separately for each variable, and missing or rejected variables are masked individually.
Unlike anchor eligibility, supervision does not require every station to observe all four variables.
This configuration uses neither spatial flips nor station dropout.
Training draws are reproducible from a seed and sample index; fixed evaluation draws use seed 1234.
\Cref{fig:supp_sampling_algo} summarizes one draw, and \cref{fig:supp_sampling_example} shows an example.
With $\alpha=0.5$, the 10\% of training stations with the longest records receive 18\% of the draws, compared with 25\% when sampling proportionally to available windows ($\alpha=1$).

\begin{figure}[t]
  \centering
  \small
  \fbox{\begin{minipage}{0.94\columnwidth}
  \textbf{One training draw} (split $\mathcal{S}$, dates $\mathcal{D}$, $k=3$, $\alpha=0.5$)
  \begin{enumerate}\setlength{\itemsep}{1pt}\setlength{\parskip}{0pt}
    \item For each station $s$ in $\mathcal{S}$, count its eligible windows $n_s$: $k$ consecutive hours in $\mathcal{D}$ in which all four variables pass quality control and HRRR forecasts exist.
    \item Draw an anchor station with probability $p(s)\propto n_s^{\alpha}$ (\cref{eq:appendix_station_sampling}).
    \item Draw one of its eligible windows uniformly; its last hour is the valid time $t$.
    \item Draw an HRRR initialization and lead uniformly among the pairs that are available for $t$.
    \item Fetch all inputs on a $28.8\times28.8$\,km region centered on the anchor, and cut a $14.4\times14.4$\,km crop at a random offset (centered during evaluation).
    \item Use every station of $\mathcal{S}$ inside the crop as a target at $t$: average stations that share a pixel, and mask variables that are missing or fail quality control.
  \end{enumerate}
  \end{minipage}}
  \caption{Station-anchored sampling of a training example.}
  \label{fig:supp_sampling_algo}
\end{figure}

\subsection{Normalization}
\label{supp:normalization}
Inputs and target are scaled per variable (\cref{tab:supp_normalization}).
The HRRR statistics include the validation and test areas, but not the 2026 test year.
The correction $\Delta_\theta$ (\cref{eq:residual}) and the prior anomaly use fixed scales: the within-lead spread of the observation minus the HRRR forecast in training samples (4.2\,$^\circ$C for $T$, 4.8\,$^\circ$C for $T_d$, 2.2\,m\,s$^{-1}$ for wind), rounded up, with $T$ lowered to 3\,$^\circ$C to balance the four targets.
These fields are scaled but not centered: the prior anomaly is a departure from the forecast and the correction a residual to the prior, so their means are small compared with these scales.
Both Earth embeddings are unit vectors per 100\,m cell (\cref{supp:surface}), stored as 8-bit integers $q$ and decoded as $\operatorname{sign}(q)\,(q/127.5)^2$.

\begin{table}[t]
  \centering
  \footnotesize
  \setlength{\tabcolsep}{4pt}
  \begin{tabular}{@{}l r r@{}}
    \toprule
    & Mean & Scale \\
    \midrule
    \multicolumn{3}{@{}l}{\emph{HRRR forecast, 3\,km: $z$-score with HRRR analysis statistics}} \\
    $T$ & 11.3\,$^\circ$C & 12.5\,$^\circ$C \\
    $T_d$ & 3.2\,$^\circ$C & 11.4\,$^\circ$C \\
    $u$, $v$ & 0 & 3.0\,m\,s$^{-1}$ \\
    Skin temperature & 12.0\,$^\circ$C & 14.0\,$^\circ$C \\
    Surface pressure & 934.1\,hPa & 73.9\,hPa \\
    Cloud cover & 34\,\% & 41\,\% \\
    Soil moisture availability & 41\,\% & 36\,\% \\
    \midrule
    \multicolumn{3}{@{}l}{\emph{Correction $\Delta_\theta$ and prior anomaly: fixed scale}} \\
    $T$ & 0 & 3\,$^\circ$C \\
    $T_d$ & 0 & 5\,$^\circ$C \\
    $u$, $v$ & 0 & 2.5\,m\,s$^{-1}$ \\
    \midrule
    \multicolumn{3}{@{}l}{\emph{Surface, 100\,m}} \\
    Elevation & 735\,m & 693\,m \\
    Surface normals & \multicolumn{2}{r}{unscaled} \\
    Earth embeddings (OlmoEarth, AlphaEarth) & \multicolumn{2}{r}{unscaled, unit length} \\
    \bottomrule
  \end{tabular}
  \caption{\textbf{Normalization.} A scaled input is $(x-\text{mean})/\text{scale}$. HRRR statistics are from the HRRR analysis: every hour of 2017--2025 and every land cell in the United States, Canada and Mexico. The two wind components share one scale.}
  \label{tab:supp_normalization}
\end{table}

\section{Physics Prior}
\label{supp:prior}
The prior $P=\mathcal{P}(X,S_t,c)$ (\cref{sec:physical_prior}) adjusts the 3\,km HRRR forecast to the 100\,m terrain. It is differentiable and runs on the GPU for every training sample.

\paragraph{Wind.} The HRRR 10\,m wind is interpolated bicubically to 100\,m and projected onto a divergence-free flow over a terrain-following mesh of depth $H_0$, following WindNinja's mass-consistent formulation~\cite{windninja2025}: the solve finds the divergence-free field closest to the interpolated wind, and the weight ratio $\alpha$ sets how much of the adjustment goes over rather than around terrain. Only the divergence induced by the 100\,m terrain is removed; the forecast's own large-scale convergence is kept. Instead of WindNinja's stability classes, a small network (two hidden layers of 16 units, about 400 parameters) predicts $\alpha$ per pixel, bounded to $[0.2,5]$, from the forecast skin--air temperature difference $\Delta T$, cloud fraction and wind speed, the terrain slope and the solar zenith angle. The speed is then scaled by $\exp\!\big(k_0+k_1\tanh(\Delta T/3\,\mathrm{K})\big)$, which removes most of HRRR's fast bias at stations (wind-speed bias $+1.70$ against $-0.29$\,m\,s$^{-1}$, \cref{tab:prior_skill}). The solve uses conjugate gradients with a fast-Poisson preconditioner (tolerance $10^{-4}$).

\paragraph{Temperature and dew point.} Both are corrected for the elevation anomaly $\Delta h$, the difference between the 100\,m terrain and its local mean over 3\,km. Temperature uses the lapse rate $\Gamma_T+\Gamma_{\mathrm{stab}}\tanh(\Delta T/3\,\mathrm{K})$, which steepens on sunny afternoons and flattens or inverts on stable nights, capped at $\Gamma_{\max}$. Dew point uses separate lapse rates below and above the local mean, plus a valley-moisture term $A_v\tanh(\max(-\Delta h,0)/d_v)$ that moistens valley floors and levels off in deep valleys. Finally, $T_d\le T$ is enforced.

\begin{figure*}[t]
  \centering
  \includegraphics[width=\textwidth]{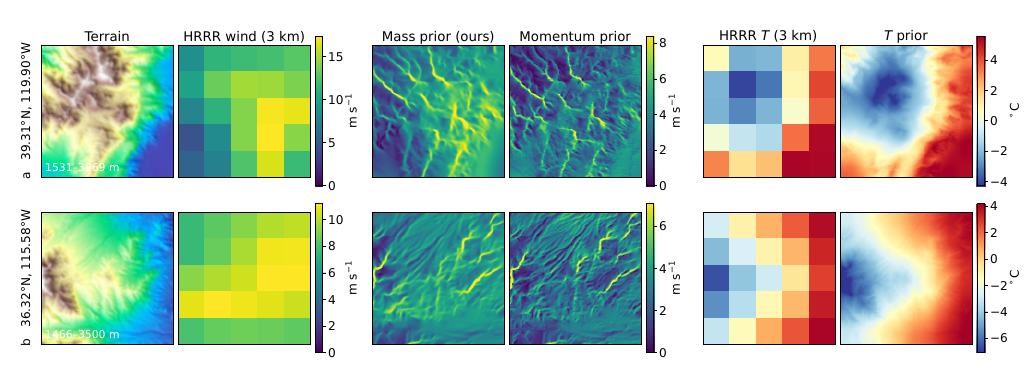}
  \caption{\textbf{The physics prior compared with HRRR} on two held-out test tiles with strong relief: (\textbf{a}) Carson Range, Nevada; (\textbf{b}) Spring Mountains, Nevada. From left: terrain; HRRR 10\,m wind speed at 3\,km (own color scale, about three times the priors' speeds); the mass prior and the momentum prior (shared scale); HRRR 2\,m temperature and the temperature prior (shared scale). The mass prior concentrates the wind on exposed ridges, the momentum prior adds finer streaks, and the temperature prior cools with elevation.}
  \label{fig:supp_prior}
\end{figure*}

\paragraph{Calibration.} The constants in \cref{tab:prior_params} are fitted to observations at the training stations by gradient descent through the prior (1{,}000 Adam steps), and checked on held-out validation stations. The wind constants and the $\alpha$ network are fitted to station error plus a station-pair structure term; the temperature and dew-point constants to station error only. We keep the constants at the end of the fit and freeze them for training. We also tested a momentum variant of the wind prior, which follows the mass solve with two rounds of a layered momentum solve and has its own calibrated constants. Alone, it matches the mass prior in wind-speed MAE and is slightly better in direction (35.94$^\circ$ against 36.23$^\circ$), but it underestimates wind speed more (bias $-$0.37 against $-$0.29\,m\,s$^{-1}$) and has slightly weaker structure ($R_{\mathrm{pair}}$ 0.33 against 0.35; \cref{tab:prior_skill}), so we keep the simpler mass prior. \Cref{fig:supp_prior} shows both priors and the temperature prior next to HRRR.

\begin{table*}[t]
  \centering
  \footnotesize
  \setlength{\tabcolsep}{3pt}
  \begin{tabular}{@{}l l r l@{}}
    \toprule
    & Parameter & Value & Unit \\
    \midrule
    Wind & Mesh depth $H_0$ & 1284 & m \\
     & Over/around weight $\alpha$ & per pixel, $[0.2,5]$ & -- \\
     & Speed factor $k_0$ / $k_1$ & $-0.914$ / $0.040$ & -- \\
     & Bulk speed scaling $e^{k_0}$ & 0.40 & -- \\
    \midrule
    Temperature & Base lapse rate $\Gamma_T$ & 3.3 & K\,km$^{-1}$ \\
     & Stability term $\Gamma_{\mathrm{stab}}$ & 11.2 & K\,km$^{-1}$ \\
     & Cap $\Gamma_{\max}$ & 7 & K\,km$^{-1}$ \\
    \midrule
    Dew point & Lapse rate below / above local mean & 3.0 / 0.1 & K\,km$^{-1}$ \\
     & Valley moisture $A_v$ & 1.42 & K \\
     & Valley depth scale $d_v$ & 300 & m \\
    \bottomrule
  \end{tabular}
  \caption{Calibrated parameters of the physics prior, fitted on the training stations.}
  \label{tab:prior_params}
\end{table*}

\ifdefined\supptabbox\else\newsavebox{\supptabbox}\fi
\begin{table}[t]
\centering
\footnotesize
\setlength{\tabcolsep}{2pt}
\sbox{\supptabbox}{%
\begin{tabular}{l cccc ccc ccc}
\toprule
 & \multicolumn{4}{c}{MAE $\downarrow$} & \multicolumn{3}{c}{Bias $\to 0$} & \multicolumn{3}{c}{$R_{\mathrm{pair}}$ $\uparrow$} \\
\cmidrule(lr){2-5}\cmidrule(lr){6-8}\cmidrule(lr){9-11}
 & ws & dir & $T$ & $T_d$ & ws & $T$ & $T_d$ & ws & $T$ & $T_d$ \\
\midrule
HRRR interp. & 1.97 & 36.98 & 1.71 & 2.53 & +1.70 & +0.07 & $-$1.57 & 0.06 & 0.19 & 0.15 \\
Prior (mass, ours) & 1.00 & 36.23 & 1.68 & 2.48 & $-$0.29 & +0.07 & $-$1.45 & 0.35 & 0.36 & 0.27 \\
Prior (momentum) & 1.00 & 35.94 & 1.68 & 2.48 & $-$0.37 & +0.07 & $-$1.45 & 0.33 & 0.36 & 0.27 \\
\bottomrule
\end{tabular}}%
\ifdim\wd\supptabbox>\columnwidth\resizebox{\columnwidth}{!}{\usebox{\supptabbox}}\else\usebox{\supptabbox}\fi
\caption{The physics prior alone on the test set of \cref{tab:results} (2026, held-out stations). Units as in \cref{tab:results}. The two wind priors share the temperature and dew-point prior.}
\label{tab:prior_skill}
\end{table}

\section{Vernier Architecture}
\label{supp:architecture}
\Cref{tab:model_io} lists Vernier's inputs, outputs and station targets.

\begin{table}[t]
  \centering
  \scriptsize
  \setlength{\tabcolsep}{3pt}
  \renewcommand{\arraystretch}{1.08}
  \supptabfit{\columnwidth}{%
  \begin{tabular}{@{}l>{\raggedright\arraybackslash}p{0.60\columnwidth}l@{}}
    \toprule
    Notation & Content / role & Size \\
    \midrule
    $X_{\mathrm{fc}}$ & HRRR forecasts of $(T, T_d, u, v)$ & $4\times5\times5$ \\
    $X_{\mathrm{aux}}$ & Cloud cover, skin temperature, surface pressure, soil moisture & $4\times5\times5$ \\
    \multirow{2}{*}{$S$} & Terrain: elevation and surface normal & $4\times144\times144$ \\
     & Earth-observation embeddings & $64\times144\times144$ \\
    $c$ & Time of day, solar insolation from latitude and day of year, forecast lead & Metadata \\
    $z^{(k)}$ & One Gaussian draw per member, entering through AdaLN & 64 \\
    $m^{(k)}$ & Dropout masks for fine-grid feed-forward activations & Per layer \\
    \midrule
    $P$ & Computed physics prior; prediction baseline & $4\times144\times144$ \\
    $A$ & Normalized prior anomaly; fine-grid neural input & $4\times144\times144$ \\
    \midrule
    $\Delta_\theta$ & Learned correction to the physics prior & $4\times144\times144$ \\
    $\widehat{Y}^{(k)}$ & Final member: $(T, T_d, u, v)$ & $4\times144\times144$ \\
    \midrule
    $y_i$ & Station observations of $(T, T_d, u, v)$; supervision only & Sparse \\
    \bottomrule
  \end{tabular}
  }%
  \caption{\textbf{Vernier inputs, outputs and station targets.} Sizes are channels $\times$ height $\times$ width for a 14.4\,km crop. Fine grids have 100\,m spacing; the 3\,km HRRR fields are registered to a $5\times5$ grid at 2.88\,km. $z^{(k)}$ and $m^{(k)}$ together form the random input $\epsilon^{(k)}$ of \cref{eq:prob}. Outputs are in $^\circ$C for $T$ and $T_d$ and m\,s$^{-1}$ for the east and north wind. Temperature and dew-point targets are residuals to the prior in normalized units, the wind is scored as the full vector (\cref{supp:deterministic,supp:probabilistic}), and missing variables are masked.}
  \label{tab:model_io}
\end{table}

\subsection{Multi-stream encoder}
\label{supp:encoder}
The encoder is a multi-stream transformer with joint attention across streams~\cite{esser2024mmdit}: eight blocks of width 512 with 16 heads. Each input has its own patch embedding. The prior anomaly $A$ (4 channels), the terrain (4 channels) and the Earth embedding (64 channels) enter at 100\,m in $8\times8$-pixel patches, an $18\times18$ token grid per $144\times144$ tile (embed dim: 512); the HRRR forecast of the four target variables and the auxiliary HRRR fields enter at 3\,km with one token per grid cell (embed dim: 128). All tokens attend to each other in every block, with rotary position embeddings~\cite{heo2024ropevit} in physical coordinates so that 100\,m and 3\,km tokens share one frame. The context (time of day, solar insolation from latitude and day of year, and forecast lead time) modulates every block through AdaLN. The output tokens of the prior stream go to the decoder (\cref{supp:decoder}).

\paragraph{Rotary positions.}
Each token's position $p=(p_y,p_x)$ is its center in meters, in one frame shared by all streams: the 100\,m streams have a token every 800\,m and the HRRR streams one per grid cell.
Each 32-dimensional attention head rotates 12 of its dimensions by $p_y$ and 12 by $p_x$, in pairs $i=0,\dots,5$ at the angles
\begin{equation}
  \phi_{a,i} = \frac{p_a}{s}\,\theta^{-i/6}, \qquad a\in\{y,x\},
  \label{eq:supp_rope}
\end{equation}
with $s=800$\,m and $\theta=10^{4}$.
This is the 2D RoPE of Heo~\etal~\cite{heo2024ropevit} with token indices replaced by ground coordinates in units of the fine token stride $s$, so the attention score of two tokens depends only on their displacement on the ground, whatever their resolution.
The remaining 8 dimensions carry a temporal rotation that is the same for every token, since all inputs refer to one valid time.

\paragraph{Solar insolation.}
The insolation input is the daily-mean top-of-atmosphere insolation at the tile's central latitude $\varphi$ on day of year $d$,
\begin{equation}
  Q=\frac{S_0}{\pi}\,E\,\bigl(\omega_s\sin\varphi\sin\delta+\cos\varphi\cos\delta\sin\omega_s\bigr),
  \label{eq:supp_insolation}
\end{equation}
with the solar constant $S_0=1361$\,W\,m$^{-2}$, the Earth--Sun distance correction $E=\bigl(1+0.0167\cos(2\pi(d-3)/365)\bigr)^{2}$, the declination $\delta=-23.44^\circ\cos\bigl(2\pi(d+10)/365\bigr)$ and the sunset hour angle $\omega_s=\arccos(-\tan\varphi\tan\delta)$.
It enters as three scalars: $\hat Q=Q/(550\,\mathrm{W\,m^{-2}})$; its trend $\bigl(\hat Q(d+1)-\hat Q(d-1)\bigr)\sin(\varphi-\delta)/0.0352$, scaled to lie in $[-1,1]$; and the seasonal phase $2\pi(d-172)/365$.
Each scalar is expanded into sinusoidal features and mapped by a two-layer MLP to 64 dimensions.
Together with the hour, this tells the model how much sunlight the tile receives, which sets the strength of daytime heating and nighttime cooling and hence how the near-surface fields depart from the coarse forecast.

\subsection{Surface-anchored decoder}
\label{supp:decoder}
The decoder maps the $18\times18$ grid of 512-dimensional output tokens (one per $8\times8$-pixel patch of a $144\times144$ tile) to the 100\,m correction of \cref{eq:surface_decoder} for each output variable $v\in\{T,T_d,u,v\}$:
\begin{equation}
    \Delta_v(x)=b_v(x)+\sum_{g}\sum_{k=1}^{8}W_{v,g,k}(x)\,\phi_{g,k}(x),
\end{equation}
with $g\in\{\text{terrain},\text{embedding},\text{prior}\}$. Each feature map $\phi_g$ is a small pointwise network at 100\,m (two $1\times1$ convolutions with 64 hidden channels) applied to the terrain channels, the 64 embedding channels or the four prior fields, giving eight features per input. The coefficients $b$ and $W$ are linear projections of the tokens (4 and 96 channels) on the token grid, upsampled bicubically by a factor of 8. There is no learned spatial convolution at 100\,m, so detail finer than a token can only follow the terrain, the embedding and the prior. The coefficient heads start at zero, so training starts from the prior. The decoder has 58K parameters; the eight transformer blocks have 126M.

\subsection{Generative Ensembles}
\label{supp:gen-ens}
Vernier-Prob keeps the encoder and decoder of \cref{supp:encoder,supp:decoder} and adds two sources of randomness per member (\cref{sec:stochastic_prediction}): a noise vector that enters the transformer blocks through their adaptive layer normalization, as in FGN~\cite{alet2025fgn}, and dropout kept active at inference (Monte Carlo dropout), as in U-Cast~\cite{cachay2026ucast}; WeatherNext~3 combines the two~\cite{rasp2026weathernext}. Each member is one forward pass. The prior, the patch embeddings and the decoder features $\phi_g$ of \cref{eq:surface_decoder} are computed once per tile and shared by all members, so members differ only in their encoder tokens and hence in the decoder coefficients $b$ and $W_g$.

\begin{figure}[t]
  \centering
  \includegraphics[width=\columnwidth]{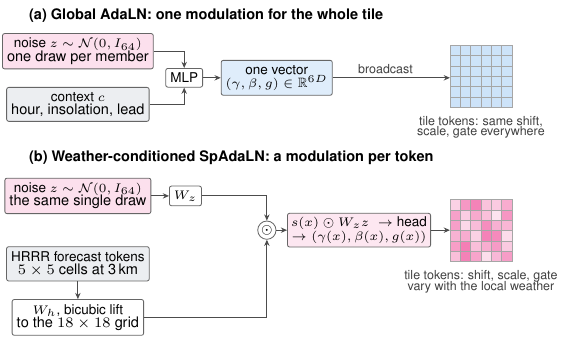}
  \caption{\textbf{Global AdaLN versus weather-conditioned SpAdaLN.} (a) Global modulation: one noise draw per member, together with the context, is mapped to one shift, scale and gate vector that every token of the tile receives. (b) SpAdaLN: the same single draw $z$ is multiplied by the lifted HRRR-forecast embedding $s(x)$, so each token gets its own shift, scale and gate, and the perturbation follows the local weather.}
  \label{fig:supp_adaln_vs_spadaln}
\end{figure}

\paragraph{AdaLN.}
Every block modulates its prior-stream tokens $h\in\mathbb{R}^{N\times D}$ ($N=18\times18$ tokens, $D=512$) with six vectors derived from the context $c$: a scale $\gamma$, a shift $\beta$ and a gate $g$ for the attention sub-layer, and three more for the MLP sub-layer,
\begin{equation}
    h \leftarrow h + g\odot \mathrm{Attn}\big((1+\gamma)\odot\mathrm{LN}(h)+\beta\big),
    \label{eq:supp_adaln}
\end{equation}
and likewise for the MLP. In the deterministic model, $(\gamma,\beta,g)=M_\ell(c)$ is one vector per block $\ell$, shared by all $N$ tokens.

\paragraph{Global modulation.}
Each member draws $z\sim\mathcal{N}(0,I_{64})$ once per tile. An MLP ($64\to512\to512$, SiLU) maps it to a vector $q$, and every block adds a learned correction to its six AdaLN parameters,
$(\Delta\gamma,\Delta\beta,\Delta g)_\ell = M^{z}_\ell\,\mathrm{SiLU}(q)$, with $M^z_\ell$ a linear layer $512\to6\cdot512$. The correction is the same at every token: one draw perturbs the whole tile coherently, as in FGN~\cite{alet2025fgn} (\cref{fig:supp_adaln_vs_spadaln}a).

\paragraph{Weather modulation (SpAdaLN).}
Spatial AdaLN gives every token its own correction, computed from the same single draw and the local forecast (\cref{fig:supp_adaln_vs_spadaln}b). Let $e$ be the patch embeddings of the HRRR forecast tokens (the four target variables on the $5\times5$ grid of 2.88\,km cells, width 128, taken before the first block). A linear layer $W_h$ ($128\to512$) maps them to the token width and a bicubic lift brings the $5\times5$ field to the $18\times18$ token grid, giving a weather field $s(x)\in\mathbb{R}^{512}$ at every token $x$. The draw is projected by a bias-free linear layer $W_z$ ($64\to512$) and multiplied elementwise,
\begin{equation}
    \begin{aligned}
        u(x) &= s(x)\odot W_z z,\\
        (\Delta\gamma,\Delta\beta,\Delta g)_\ell(x) &= M^{z}_\ell\,\mathrm{SiLU}\big(u(x)\big),
    \end{aligned}
    \label{eq:supp_spadaln}
\end{equation}
and \cref{eq:supp_adaln} is applied with token-dependent $\gamma(x)$, $\beta(x)$ and $g(x)$. Because $u(x)$ is bilinear in the draw and the forecast, the direction and size of the perturbation at a token depend on the weather there: the same draw can perturb a windy ridge and a calm valley differently, so the ensemble spread can vary with the regime inside a tile. The forecast embedding is shared by all members, so the only randomness is still the single 64-dimensional draw. Only the prior stream is modulated; the conditioning streams keep their deterministic AdaLN, but joint attention lets them see the perturbed prior tokens.

Both forms add the eight per-block heads $M^z_\ell$ (12.6M parameters) plus $W_h$ and $W_z$ (0.1M) or the MLP (0.3M) to the 129.4M parameters of Vernier-Det, for 142.1M and 142.3M parameters. The heads are initialized with small random weights (standard deviation $10^{-3}$) rather than zero: probabilistic training starts from the deterministic checkpoint, and zero-initialized heads would make all members identical at the start.

\paragraph{Surface modulation.}
An alternative version used SpAdaLN that took the field $s(x)$ from the surface descriptors instead of the forecast: a linear layer over each token's terrain and Earth-embedding patch embeddings, so the perturbation varied with the surface but not with the weather.
It gave no overall advantage in our ablations, even when added to the forecast field (Surface SpAdaLN in \cref{tab:supp_prob_ablation}: better direction but worse $T$, $T_d$ and spread--skill), and was dropped.

\paragraph{Dropout.}
Following WeatherNext~3~\cite{rasp2026weathernext}, dropout ($p=0.1$) is applied to the hidden activations of the MLP sub-layer of every block in the prior, terrain and embedding streams, during training and inference. In training, the $K=2$ members of a sample share one dropout mask and differ only in their draws; at inference, every member draws its own mask and noise.

\paragraph{Design evidence.}
The probabilistic block of \cref{tab:ablation} supports these choices. Dropout alone (MC dropout, no noise) is clearly worse in temperature, dew point and direction ($+0.020$ and $+0.016$ in fCRPS, $+0.24^\circ$). Noise without dropout is within seed variation of the full model; we keep dropout following WeatherNext~3. As a single model, global modulation (Global*) is slightly worse than weather modulation for $T$ and $T_d$ ($+0.004$ and $+0.006$ in fCRPS) and slightly better in wind-speed spread--skill, so the two make different errors. Pooling 32 members of each (``Diff.\ models'') improves every fCRPS and every spread--skill ratio over the weather-modulated model; compared with pooling two seeds of the weather-modulated model, it brings the spread--skill ratio closer to one for all three variables, at slightly worse $T$ fCRPS. This 32+32 deep ensemble is Vernier-Prob. Starting from the deterministic model also matters: a randomly initialized probabilistic model trained for twice as many steps is worse on every variable.

\paragraph{Inference.}
The members are generated in one batched forward pass, with the prior solved once per tile. With Searchlight (\cref{sec:searchlight}), each member draws one $z$ for the whole domain, so the weather modulation produces one coherent perturbation field over it. \Cref{tab:supp_prob_impl} lists the training configuration of the two parts.

\section{Training}
\label{supp:training}

\subsection{Deterministic training}
\label{supp:deterministic}
Vernier-Det has 129M parameters, almost all in the eight transformer blocks (width 512, 16 heads). Temperature and dew point are trained on their residual to the prior, scaled by 3\,$^\circ$C and 5\,$^\circ$C, with an L2 loss and $\lambda_T=\lambda_{T_d}=1$. The wind loss of \cref{eq:deterministic_loss} is computed on the physical wind in m\,s$^{-1}$ with $\lambda_{\mathrm{wind}}=0.04$; in its direction term the predicted speed is softened to $\sqrt{|\widehat{\mathbf{v}}|^2+\epsilon^2}$ with $\epsilon=0.5$\,m\,s$^{-1}$, so near-calm predictions get bounded gradients. Embedding dropout (rate 0.3) is applied in the encoder and the decoder during training. We train for 4{,}000 steps on 8 GPUs with 280 tiles per GPU (2{,}240 per step, 8.96M samples in total) using AdamW ($\beta_1=0.9$, $\beta_2=0.999$, weight decay 0.01), a peak learning rate of $5.9\times10^{-4}$ (a base rate of $2\times10^{-4}$ scaled by $\sqrt{2240/256}$), 500 warm-up steps followed by cosine decay, gradient clipping at 1.0 and bf16 mixed precision. Evaluation uses an exponential moving average of the weights (decay 0.995). \Cref{tab:supp_det_impl} summarizes the configuration.

\begin{table}[t]
  \centering
  \footnotesize
  \begin{tabular}{@{}lp{0.56\columnwidth}@{}}
    \toprule
    Parameters & 129.4M \\
    Optimizer & AdamW ($\beta_1=0.9$, $\beta_2=0.999$, weight decay 0.01), gradient clip 1, bf16 \\
    Peak learning rate & $5.9\times10^{-4}$ ($2\times10^{-4}$ scaled by $\sqrt{2240/256}$) \\
    Learning-rate schedule & 500 warm-up steps, cosine decay to zero \\
    Training steps & 4k, from scratch \\
    Global batch size & 2,240 tiles (280 per GPU) \\
    Samples seen & 8.96M \\
    Hardware & 8 $\times$ NVIDIA H100 \\
    Wall-clock time & 9.7\,h (with in-loop validation) \\
    GPU-hours (H100) & 77 \\
    Checkpoint used & Last (step 4k), EMA weights (decay 0.995) \\
    \bottomrule
  \end{tabular}
  \caption{Vernier-Det implementation details.}
  \label{tab:supp_det_impl}
\end{table}

\subsection{Spectral loss}
\label{supp:spectral}
The spectral loss is added to the station loss of the deterministic model and, for the probabilistic model, applied to the ensemble mean:
\begin{equation}
    \mathcal{L}_{\mathrm{spec}}
    =\frac{1}{|\mathcal{B}|}\sum_{b\in\mathcal{B}}
    \left[\log\frac{\max(E_b^{\mathrm{pred}},\varepsilon)}
    {\tau\,\widetilde{E}_b^{\mathrm{prior}}}\right]_+^2,
    \label{eq:spectral_regularization}
\end{equation}
where $[a]_+=\max(a,0)$, $\varepsilon=10^{-20}$ guards the logarithm, and $E_b^{\mathrm{pred}}$ and $E_b^{\mathrm{prior}}$ are the mean Fourier powers of the predicted and prior wind in band $b$, summed over $u$ and $v$ and computed per tile. Only power above $\tau$ times the prior's is penalized, so detail below this limit is kept. The reference $\widetilde{E}_b^{\mathrm{prior}}=\max(E_b^{\mathrm{prior}},\rho\langle E_b^{\mathrm{prior}}\rangle_{\mathrm{batch}},\varepsilon)$ has a floor at $\rho=0.01$ of the batch-mean prior power, so nearly flat tiles keep a positive reference, and the prior is detached. We use five bands covering 200\,m--3.2\,km (200--400, 400--600, 600--800\,m, 0.8--1.6 and 1.6--3.2\,km), $\tau=2.5$ and $\lambda_{\mathrm{spec}}=0.01$. The prior has almost no power below about 250\,m, so in the 200--400\,m band its power is replaced by an extrapolation of its own 0.8--3.2\,km power-law slope.

\begin{figure*}[t]
  \centering
  \includegraphics[width=\textwidth]{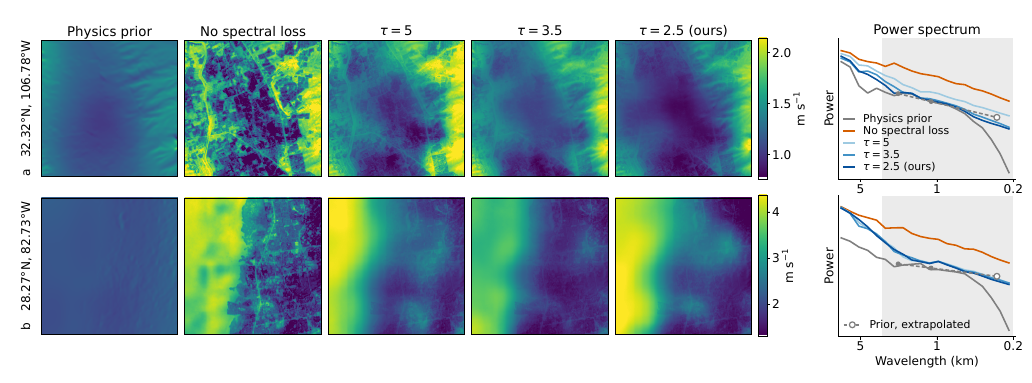}
  \caption{\textbf{The spectral loss at different tolerances} on two held-out test tiles: (\textbf{a}) Mesilla Valley, New Mexico; (\textbf{b}) Tampa Bay area, Florida. Wind speed of the physics prior and of Vernier-Det without the spectral loss and with it at $\tau=5$, 3.5 and 2.5 (ours), on one color scale per row, and their power spectra (loss band shaded). Filled points: the prior's mean power in the 1.6--3.2 and 0.8--1.6\,km bands; the dashed power law through them gives the reference of the 200--400\,m band (open circle). Without the loss, the wind copies the field and urban pattern of the surface; with it, the power in the loss band drops toward the prior's, and a smaller $\tau$ removes more of it (\textbf{a}).}
  \label{fig:supp_spectral}
\end{figure*}

\Cref{tab:spectral_tradeoff} shows the trade-off on the validation stations. Without the loss, the model has the best wind-speed MAE and $R_{\mathrm{pair}}$ but copies much of the embedding into the wind: on flat tiles, the embedding explains 71\% of its 200--800\,m departure from the prior. Any of the tested losses cuts this to 30--42\%, at a cost of 0.01--0.02\,m\,s$^{-1}$ in wind-speed MAE and 0.01--0.04 in $R_{\mathrm{pair}}$. A smaller $\tau$ removes more of the imprint and generally costs more structure; the band range matters less than $\tau$, and the widest range with $\tau=2.5$ gives the lowest imprint. Temperature and dew point change by at most 0.006\,$^\circ$C in MAE. \Cref{fig:supp_spectral} shows the fields.

\ifdefined\supptabbox\else\newsavebox{\supptabbox}\fi
\begin{table}[t]
\centering
\footnotesize
\setlength{\tabcolsep}{2pt}
\sbox{\supptabbox}{%
\begin{tabular}{ll cccc}
\toprule
Bands & $\tau$ & ws MAE $\downarrow$ & dir MAE $\downarrow$ & ws $R_{\mathrm{pair}}$ $\uparrow$ & Imprint $\downarrow$ \\
 & & {\scriptsize m\,s$^{-1}$} & {\scriptsize $^\circ$} & & \\
\midrule
\multicolumn{2}{l}{No spectral loss} & 0.836 & 30.33 & 0.538 & 0.71 \\
\midrule
200--800\,m & 2.5 & 0.846 & 30.60 & 0.502 & 0.33 \\
 & 3.5 & 0.849 & 30.63 & 0.512 & 0.37 \\
 & 5 & 0.842 & 30.59 & 0.524 & 0.41 \\
\addlinespace
200\,m--1.6\,km & 2.5 & 0.849 & 30.54 & 0.500 & 0.33 \\
 & 3.5 & 0.849 & 30.63 & 0.512 & 0.38 \\
 & 5 & 0.843 & 30.48 & 0.515 & 0.42 \\
\addlinespace
200\,m--3.2\,km & \textbf{2.5} & 0.853 & 30.59 & 0.507 & 0.30 \\
 & 3.5 & 0.847 & 30.33 & 0.518 & 0.34 \\
 & 5 & 0.849 & 30.57 & 0.510 & 0.39 \\
\bottomrule
\end{tabular}}%
\ifdim\wd\supptabbox>\columnwidth\resizebox{\columnwidth}{!}{\usebox{\supptabbox}}\else\usebox{\supptabbox}\fi
\caption{Spectral-loss trade-off on the validation stations (50k samples, 2017--2025). Rows change only the bands and the tolerance $\tau$; bold: our setting. \emph{Imprint}: the share of the 200--800\,m variance of the predicted wind's departure from the prior that the embedding explains linearly, on flat tiles.}
\label{tab:spectral_tradeoff}
\end{table}

\begin{table}[t]
  \centering
  \footnotesize
  \supptabfit{\columnwidth}{%
  \begin{tabular}{@{}lp{0.56\columnwidth}@{}}
    \toprule
    Parameters & 142.1M (weather mod.) / 142.3M (global mod.), 129.4M of them shared with Vernier-Det \\
    Optimizer & AdamW ($\beta_1=0.9$, $\beta_2=0.999$, weight decay 0.01), gradient clip 1, bf16 \\
    Peak learning rate & $4.2\times10^{-4}$ ($2\times10^{-4}$ scaled by $\sqrt{1120/256}$) \\
    Learning-rate schedule & 500 warm-up steps, cosine decay to zero \\
    Training steps & 10k per part, from the Vernier-Det checkpoint, $K=2$ members per sample \\
    Global batch size & 1,120 tiles (140 per GPU) \\
    Samples seen & 11.2M per part \\
    Hardware & 8 $\times$ NVIDIA H100 \\
    Wall-clock time & $\approx$14\,h per part (with in-loop validation where $K=4$) \\
    GPU-hours (H100) & $\approx$115 per part, $\approx$230 for both \\
    Checkpoint used & Last (step 10k), EMA weights (decay 0.99) \\
    Ensemble size at evaluation & 64 = 32 (weather mod.) + 32 (global mod.) \\
    \bottomrule
  \end{tabular}
  }%
  \caption{Vernier-Prob implementation details. The two parts (weather-modulated and global-modulated) are trained separately with the same settings and pooled at evaluation.}
  \label{tab:supp_prob_impl}
\end{table}

\subsection{Probabilistic training}
\label{supp:probabilistic}
\paragraph{Objective.}
Vernier-Prob is trained with the fair scores of \cref{eq:fair_score,eq:probabilistic_loss} on $K=2$ members per sample. All scores are computed in normalized units: the residuals to the prior divided by 3\,$^\circ$C for $T$ and 5\,$^\circ$C for $T_d$ (the scales of the deterministic loss, \cref{supp:deterministic}), and the full wind (prior plus correction) divided by 2.5\,m\,s$^{-1}$ (its normalization scale, \cref{tab:supp_normalization}). $C_s$ is the fair CRPS of the member wind speeds $|\widehat{\mathbf v}^{(k)}|$. $E_{uv}$ is the fair energy score of the member wind vectors with the smoothed distance $d_\epsilon(\mathbf a,\mathbf b)=\sqrt{\|\mathbf a-\mathbf b\|^2+\epsilon^2}-\epsilon$, $\epsilon=10^{-4}$, which has a finite gradient when two members coincide. The terms are summed over all observed station-variables in a batch and divided by the number of scored coordinates (one each for $T$, $T_d$ and the speed, two for the wind vector), so a station reporting all four variables counts five. We set $\lambda_T=\lambda_{T_d}=\lambda_s=1$ and $\lambda_{uv}=4/\pi$.

\paragraph{Why a joint wind score.}
Scoring $u$ and $v$ separately with the CRPS checks the two marginal distributions but not their pairing: an ensemble whose $u$ and $v$ members are each well calibrated but wrongly combined scores well while its speeds and directions are wrong. The energy score is a proper score for the joint distribution of the vector~\cite{gneiting2007strictly}. Its weight $4/\pi$ makes it comparable to the two component scores it replaces: for an isotropic Gaussian difference $\boldsymbol\Delta$, $\mathbb{E}|\Delta_u|+\mathbb{E}|\Delta_v|=\tfrac{4}{\pi}\,\mathbb{E}\|\boldsymbol\Delta\|$, so the wind keeps the same weight relative to $T$ and $T_d$ as under component-wise training.

\paragraph{Why an additional speed score.}
Wind speed is what users and our evaluation look at, and the energy score constrains it only through the vector. The fair CRPS of the member speeds is a proper score of the speed distribution and is added with $\lambda_s=1$; without it, speed fCRPS, direction and the speed $R_{\mathrm{pair}}$ get worse (\cref{tab:ablation}, ``No speed CRPS''). No term uses a calm-wind threshold: calm observations are scored by both wind terms, which are defined at zero speed. Only the evaluation's direction metric, not the loss, excludes observed speeds below 1.6\,m\,s$^{-1}$.

\paragraph{Relation to the deterministic loss.}
The deterministic wind loss (\cref{eq:deterministic_loss}) scores speed and direction, with the direction term weighted by the squared observed speed. This does not carry over to a proper scoring rule: a direction score weighted by the observation is no longer proper, and a separate direction CRPS needs a threshold below which the direction is undefined, so calm cases drop out of it. The variants we tried (speed plus circular-direction fair CRPS with such a threshold, a direction score weighted by the prior's speed, and the energy score on a polar lift of the vector) were within seed-to-seed variation of, or behind, the form above on the validation stations, so we kept the simplest proper one.

\paragraph{Schedule.}
Training starts from the Vernier-Det checkpoint and runs for 10k steps with the same optimizer, warm-up, embedding dropout and crop settings as \cref{supp:deterministic}; \cref{tab:supp_prob_impl} gives the full configuration. The spectral loss, when used, is applied to the mean of the $K$ members (\cref{supp:spectral}). Evaluation uses an exponential moving average of the weights (decay 0.99) and 64 members. In \cref{tab:results}, Vernier-Det and Vernier-Prob are trained without the spectral loss and the ``+ Spectral'' rows with it; the operational models of \cref{tab:operational} are trained with it on all stations.

\section{Evaluation}
\label{supp:evaluation}

\subsection{Spatial Structure}
\label{supp:structure}

\begin{figure*}[t]
\centering
\includegraphics[width=\textwidth]{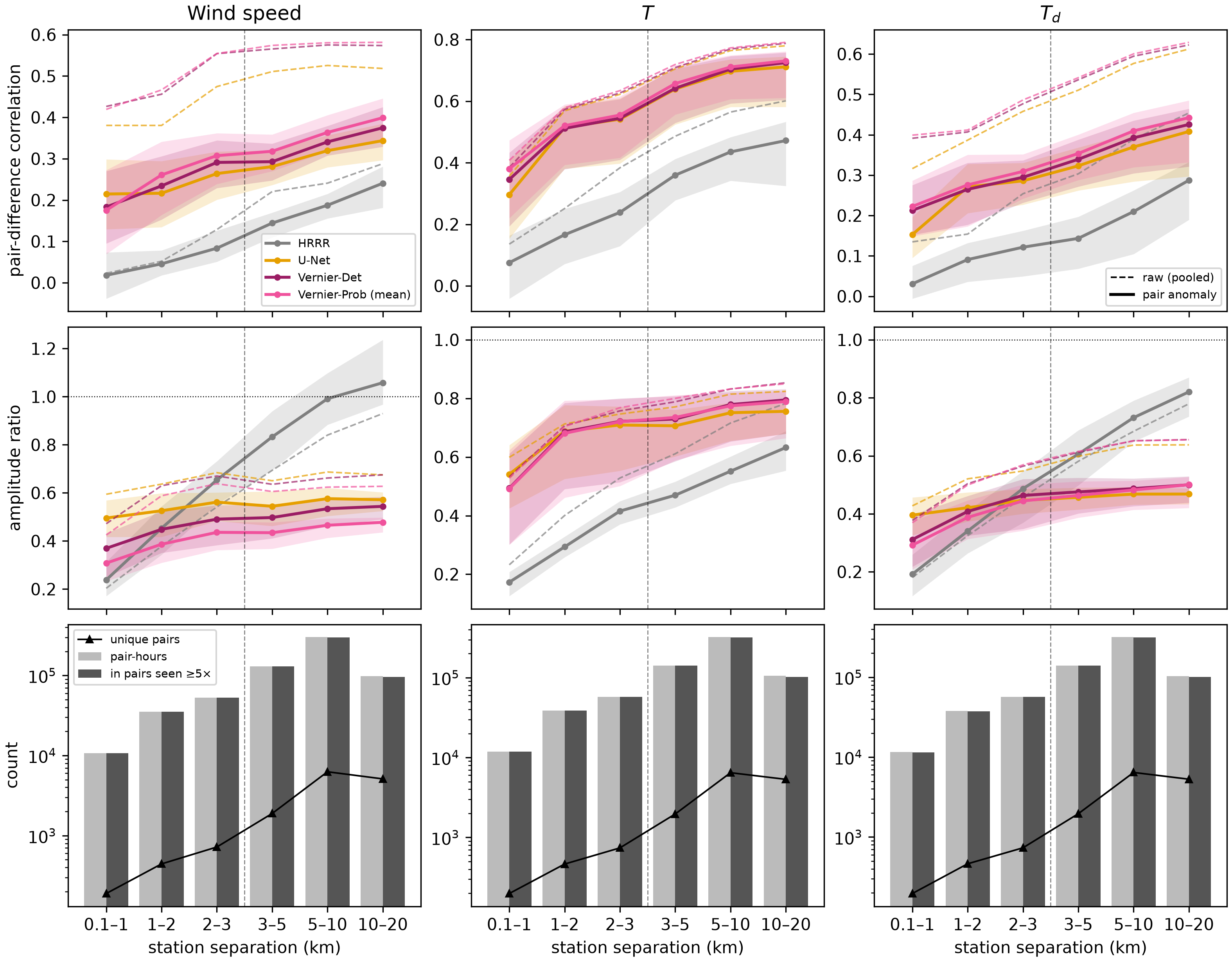}
\caption{Station-pair structure at the held-out test stations over 2017--2025 (the spatial holdout of \cref{tab:extreme}; 184k station-hours), for wind speed, $T$ and $T_d$, for HRRR (gray), U-Net (orange), Vernier-Det (dark pink) and the Vernier-Prob 64-member ensemble mean (pink); \cref{tab:pair_structure} adds the physics prior. \emph{Top}: pooled correlation between model and observed station-pair differences per separation band, raw (dashed) and after removing each pair's own mean difference over its repeated observations (pair anomaly, solid); shading gives 95\% intervals from a two-way cluster bootstrap over 50\,km cells and days. \emph{Middle}: amplitude ratio (RMS model over RMS observed pair difference), raw (dashed) and pair-de-meaned (solid); 1 means matching spatial variability. \emph{Bottom}: pair-hours, pair-hours in pairs observed at least five times, and unique station pairs per band. The dashed vertical line marks the 3\,km upper limit of $R_{\mathrm{pair}}$.}
\label{fig:pair_structure}
\end{figure*}

\paragraph{Separating persistent and time-varying structure.}
Pooled station-pair correlations can reflect persistent differences between sites, including terrain and exposure effects.
To assess whether the models also recover changes in these differences, we remove each pair's temporal mean.
For pair $p=(i,j)$, let $d_{p,t}=y_{i,t}-y_{j,t}$ and $\widehat d_{p,t}=\widehat y_{i,t}-\widehat y_{j,t}$ denote observed and predicted differences.
We form
\begin{equation}
    d'_{p,t}=d_{p,t}-\overline d_p,
    \qquad
    \widehat d'_{p,t}=\widehat d_{p,t}-\overline{\widehat d}_p,
\end{equation}
where means are computed separately for observations and each model over the pair's evaluation cases.
This analysis removes one overall mean per pair; the main-text tables use a stricter variant that removes it separately in each six-hour block of local solar time, which also cancels offsets that vary with the time of day (\cref{sec:evaluations}; \cref{supp:mos}).
These means are used only for the diagnostic; forecasts are unchanged.
We retain pairs observed at least five times, pool their anomalies within each distance band, and average the correlations over 0.1--1, 1--2, and 2--3\,km, as for $R_{\mathrm{pair}}$.
Longer-distance bands in \cref{fig:pair_structure} provide context but do not enter the reported score.
We estimate 95\% intervals using 500 two-way cluster-bootstrap replicates over 50\,km cells and valid days, with model differences computed within the same replicates.

\paragraph{Evidence beyond persistent site differences.}
De-meaning reduces correlations, particularly for wind speed and dew point, but preserves Vernier's advantage over HRRR and the physics prior (\cref{tab:pair_structure}, \cref{fig:pair_structure}).
The Vernier-Prob ensemble mean achieves anomaly correlations of $(0.25,0.49,0.27)$ for $(ws,T,T_d)$, compared with $(0.05,0.16,0.08)$ for HRRR and $(0.16,0.32,0.08)$ for the physics prior.
Its gains over the prior are $0.09$, $0.17$, and $0.18$, with paired 95\% intervals of $[0.03,0.14]$, $[0.13,0.22]$, and $[0.14,0.23]$, respectively.
Vernier-Det shows the same pattern.
The U-Net, trained on the same inputs and station loss, also follows it: de-meaning removes a similar share of its correlation (44\%, 13\% and 39\% for $(ws,T,T_d)$, against 48\%, 10\% and 38\% for the Vernier-Prob mean), and its anomaly correlations $(0.23,0.45,0.24)$ also exceed the prior's.
The drop is therefore not specific to Vernier.
After de-meaning, Vernier keeps a clear lead over the U-Net for the wind components ($+0.06$ $[0.03,0.09]$ for $u_{10}$, $+0.09$ $[0.03,0.12]$ for $v_{10}$), while for $(ws,T,T_d)$ the differences ($+0.02$, $+0.03$, $+0.03$) have 95\% intervals that reach zero.
These results indicate that the structural gains are not explained solely by persistent pair offsets.

\paragraph{Amplitude and observational support.}
Correlation does not measure the magnitude of spatial variation.
We therefore also report the ratio of predicted to observed RMS pair differences (``amplitude ratio''), both before and after de-meaning (\cref{fig:pair_structure}).
Across the three bands below 3\,km, the Vernier-Prob ensemble mean's anomaly amplitude ratios are 0.31--0.44 for wind speed, 0.49--0.72 for temperature, and 0.29--0.44 for dew point.
The mean fields therefore recover changing spatial patterns with attenuated amplitude; this does not directly measure ensemble dispersion.
The figure also reports unique station pairs and repeated pair observations, distinguishing spatial coverage from temporal sampling.
The closest band contains 198 temperature station pairs with a median separation of 0.67\,km.
Thus, these diagnostics support improved structure at observed separations below 3\,km, but do not establish effective 100\,m resolution (even the closest station pairs are mostly several hundred meters apart; median 0.67\,km).

\begin{table*}[t]
\centering
\definecolor{diffcol}{HTML}{D55E00}%
\definecolor{inccol}{HTML}{0072B2}%
\setlength{\tabcolsep}{3pt}
\resizebox{\textwidth}{!}{%
\begin{tabular}{l cc cc cc cc cc}
\toprule
 & \multicolumn{2}{c}{ws} & \multicolumn{2}{c}{$u_{10}$} & \multicolumn{2}{c}{$v_{10}$} & \multicolumn{2}{c}{$T$} & \multicolumn{2}{c}{$T_d$} \\
\cmidrule(lr){2-3}\cmidrule(lr){4-5}\cmidrule(lr){6-7}\cmidrule(lr){8-9}\cmidrule(lr){10-11}
Model & raw & $\to$ de-meaned & raw & $\to$ de-meaned & raw & $\to$ de-meaned & raw & $\to$ de-meaned & raw & $\to$ de-meaned \\
\midrule
HRRR & 0.07 & $\to$\,0.05\,{\scriptsize[0.02, 0.08]} & 0.06 & $\to$\,0.04\,{\scriptsize[0.02, 0.07]} & 0.07 & $\to$\,0.08\,{\scriptsize[0.03, 0.11]} & 0.26 & $\to$\,0.16\,{\scriptsize[0.08, 0.21]} & 0.18 & $\to$\,0.08\,{\scriptsize[0.04, 0.11]} \\
 & \multicolumn{2}{c}{\cellcolor{diffcol!25}\scriptsize\textcolor{diffcol!75!black}{$\Delta$\,$-$0.02 ($-$28\%)}} & \multicolumn{2}{c}{\cellcolor{diffcol!46}\scriptsize\textcolor{diffcol!75!black}{$\Delta$\,$-$0.02 ($-$31\%)}} & \multicolumn{2}{c}{\cellcolor{inccol!46}\scriptsize\textcolor{inccol!75!black}{$\Delta$\,+0.01 (+8\%)}} & \multicolumn{2}{c}{\cellcolor{diffcol!46}\scriptsize\textcolor{diffcol!75!black}{$\Delta$\,$-$0.10 ($-$38\%)}} & \multicolumn{2}{c}{\cellcolor{diffcol!36}\scriptsize\textcolor{diffcol!75!black}{$\Delta$\,$-$0.10 ($-$55\%)}} \\[2pt]
Physics prior & 0.32 & $\to$\,0.16\,{\scriptsize[0.11, 0.19]} & 0.27 & $\to$\,0.22\,{\scriptsize[0.15, 0.28]} & 0.25 & $\to$\,0.24\,{\scriptsize[0.08, 0.33]} & 0.35 & $\to$\,0.32\,{\scriptsize[0.20, 0.38]} & 0.28 & $\to$\,0.08\,{\scriptsize[0.05, 0.11]} \\
 & \multicolumn{2}{c}{\cellcolor{diffcol!46}\scriptsize\textcolor{diffcol!75!black}{$\Delta$\,$-$0.16 ($-$51\%)}} & \multicolumn{2}{c}{\cellcolor{diffcol!28}\scriptsize\textcolor{diffcol!75!black}{$\Delta$\,$-$0.05 ($-$19\%)}} & \multicolumn{2}{c}{\cellcolor{diffcol!37}\scriptsize\textcolor{diffcol!75!black}{$\Delta$\,$-$0.01 ($-$6\%)}} & \multicolumn{2}{c}{\cellcolor{diffcol!13}\scriptsize\textcolor{diffcol!75!black}{$\Delta$\,$-$0.03 ($-$10\%)}} & \multicolumn{2}{c}{\cellcolor{diffcol!46}\scriptsize\textcolor{diffcol!75!black}{$\Delta$\,$-$0.20 ($-$70\%)}} \\[2pt]
U-Net & 0.41 & $\to$\,0.23\,{\scriptsize[0.16, 0.28]} & 0.34 & $\to$\,0.28\,{\scriptsize[0.23, 0.34]} & 0.28 & $\to$\,0.27\,{\scriptsize[0.17, 0.34]} & 0.52 & $\to$\,0.45\,{\scriptsize[0.33, 0.51]} & 0.39 & $\to$\,0.24\,{\scriptsize[0.19, 0.27]} \\
 & \multicolumn{2}{c}{\cellcolor{diffcol!40}\scriptsize\textcolor{diffcol!75!black}{$\Delta$\,$-$0.18 ($-$44\%)}} & \multicolumn{2}{c}{\cellcolor{diffcol!25}\scriptsize\textcolor{diffcol!75!black}{$\Delta$\,$-$0.06 ($-$16\%)}} & \multicolumn{2}{c}{\cellcolor{diffcol!20}\scriptsize\textcolor{diffcol!75!black}{$\Delta$\,$-$0.01 ($-$3\%)}} & \multicolumn{2}{c}{\cellcolor{diffcol!15}\scriptsize\textcolor{diffcol!75!black}{$\Delta$\,$-$0.07 ($-$13\%)}} & \multicolumn{2}{c}{\cellcolor{diffcol!25}\scriptsize\textcolor{diffcol!75!black}{$\Delta$\,$-$0.15 ($-$39\%)}} \\[2pt]
Vernier-Det & 0.48 & $\to$\,0.24\,{\scriptsize[0.18, 0.28]} & 0.38 & $\to$\,0.33\,{\scriptsize[0.28, 0.39]} & 0.36 & $\to$\,0.35\,{\scriptsize[0.22, 0.42]} & 0.53 & $\to$\,0.47\,{\scriptsize[0.34, 0.52]} & 0.43 & $\to$\,0.26\,{\scriptsize[0.20, 0.30]} \\
 & \multicolumn{2}{c}{\cellcolor{diffcol!46}\scriptsize\textcolor{diffcol!75!black}{$\Delta$\,$-$0.24 ($-$51\%)}} & \multicolumn{2}{c}{\cellcolor{diffcol!22}\scriptsize\textcolor{diffcol!75!black}{$\Delta$\,$-$0.05 ($-$15\%)}} & \multicolumn{2}{c}{\cellcolor{diffcol!24}\scriptsize\textcolor{diffcol!75!black}{$\Delta$\,$-$0.01 ($-$4\%)}} & \multicolumn{2}{c}{\cellcolor{diffcol!14}\scriptsize\textcolor{diffcol!75!black}{$\Delta$\,$-$0.06 ($-$12\%)}} & \multicolumn{2}{c}{\cellcolor{diffcol!26}\scriptsize\textcolor{diffcol!75!black}{$\Delta$\,$-$0.17 ($-$39\%)}} \\[2pt]
Vernier-Prob (mean) & 0.48 & $\to$\,0.25\,{\scriptsize[0.19, 0.29]} & 0.39 & $\to$\,0.34\,{\scriptsize[0.29, 0.39]} & 0.38 & $\to$\,0.36\,{\scriptsize[0.25, 0.43]} & 0.54 & $\to$\,0.49\,{\scriptsize[0.35, 0.55]} & 0.43 & $\to$\,0.27\,{\scriptsize[0.20, 0.31]} \\
 & \multicolumn{2}{c}{\cellcolor{diffcol!43}\scriptsize\textcolor{diffcol!75!black}{$\Delta$\,$-$0.23 ($-$48\%)}} & \multicolumn{2}{c}{\cellcolor{diffcol!20}\scriptsize\textcolor{diffcol!75!black}{$\Delta$\,$-$0.05 ($-$13\%)}} & \multicolumn{2}{c}{\cellcolor{diffcol!27}\scriptsize\textcolor{diffcol!75!black}{$\Delta$\,$-$0.02 ($-$4\%)}} & \multicolumn{2}{c}{\cellcolor{diffcol!12}\scriptsize\textcolor{diffcol!75!black}{$\Delta$\,$-$0.05 ($-$10\%)}} & \multicolumn{2}{c}{\cellcolor{diffcol!25}\scriptsize\textcolor{diffcol!75!black}{$\Delta$\,$-$0.16 ($-$38\%)}} \\
\midrule
Vernier-Prob $-$ prior & +0.16\,{\scriptsize[0.10, 0.27]} & +0.09\,{\scriptsize[0.03, 0.14]} & +0.13\,{\scriptsize[0.08, 0.18]} & +0.12\,{\scriptsize[0.09, 0.18]} & +0.12\,{\scriptsize[0.08, 0.18]} & +0.12\,{\scriptsize[0.09, 0.18]} & +0.19\,{\scriptsize[0.14, 0.25]} & +0.17\,{\scriptsize[0.13, 0.22]} & +0.15\,{\scriptsize[0.12, 0.20]} & +0.18\,{\scriptsize[0.14, 0.23]} \\
Vernier-Prob $-$ HRRR & +0.41\,{\scriptsize[0.27, 0.49]} & +0.20\,{\scriptsize[0.13, 0.25]} & +0.33\,{\scriptsize[0.27, 0.37]} & +0.30\,{\scriptsize[0.24, 0.36]} & +0.30\,{\scriptsize[0.16, 0.39]} & +0.28\,{\scriptsize[0.17, 0.37]} & +0.28\,{\scriptsize[0.22, 0.38]} & +0.33\,{\scriptsize[0.25, 0.40]} & +0.25\,{\scriptsize[0.19, 0.30]} & +0.19\,{\scriptsize[0.14, 0.23]} \\
Vernier-Det $-$ prior & +0.16\,{\scriptsize[0.11, 0.26]} & +0.08\,{\scriptsize[0.03, 0.13]} & +0.12\,{\scriptsize[0.07, 0.18]} & +0.11\,{\scriptsize[0.08, 0.17]} & +0.11\,{\scriptsize[0.06, 0.18]} & +0.11\,{\scriptsize[0.07, 0.18]} & +0.18\,{\scriptsize[0.13, 0.25]} & +0.15\,{\scriptsize[0.11, 0.21]} & +0.14\,{\scriptsize[0.12, 0.19]} & +0.17\,{\scriptsize[0.14, 0.21]} \\
Vernier-Prob $-$ U-Net & +0.07\,{\scriptsize[0.02, 0.09]} & +0.02\,{\scriptsize[$-$0.02, 0.05]} & +0.06\,{\scriptsize[0.03, 0.08]} & +0.06\,{\scriptsize[0.03, 0.09]} & +0.10\,{\scriptsize[0.03, 0.13]} & +0.09\,{\scriptsize[0.03, 0.12]} & +0.02\,{\scriptsize[$-$0.01, 0.05]} & +0.03\,{\scriptsize[0.00, 0.07]} & +0.05\,{\scriptsize[$-$0.00, 0.09]} & +0.03\,{\scriptsize[$-$0.01, 0.07]} \\
Vernier-Det $-$ U-Net & +0.07\,{\scriptsize[0.03, 0.09]} & +0.00\,{\scriptsize[$-$0.03, 0.03]} & +0.05\,{\scriptsize[0.00, 0.08]} & +0.05\,{\scriptsize[0.01, 0.08]} & +0.08\,{\scriptsize[0.01, 0.12]} & +0.08\,{\scriptsize[0.01, 0.11]} & +0.01\,{\scriptsize[$-$0.02, 0.04]} & +0.02\,{\scriptsize[$-$0.02, 0.05]} & +0.04\,{\scriptsize[$-$0.00, 0.08]} & +0.02\,{\scriptsize[$-$0.01, 0.05]} \\
\bottomrule
\end{tabular}}
\caption{$R_{\mathrm{pair}}$ (mean pair-difference correlation over the 0.1--1, 1--2 and 2--3\,km bands) at the held-out test stations over 2017--2025 (the spatial holdout of \cref{tab:extreme}), raw and after removing each station pair's own mean difference (de-meaned), with 95\% two-way cluster-bootstrap intervals. Under each model, $\Delta$ is the change from raw to de-meaned (and as a share of the raw value); its shading is proportional to that relative change, scaled to the largest in each column (vermillion: decrease, blue: increase). Bottom: paired differences, with intervals from the same bootstrap replicates. Wind components: ensemble mean of each component (wind speed: mean of the member speeds).}
\label{tab:pair_structure}
\end{table*}

\section{Baseline Configurations}
\label{supp:baselines}

\subsection{WOOF}
\label{supp:woof}
\begin{figure}[t]
  \centering
  \includegraphics[width=\columnwidth]{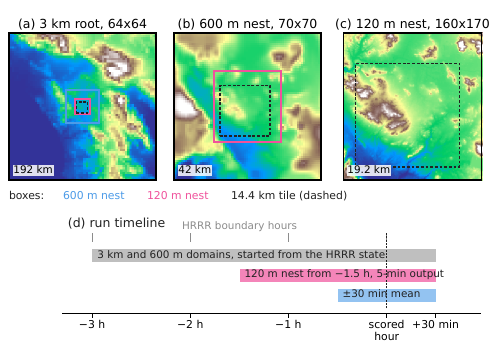}
  \caption{\textbf{WOOF nesting and timing.} (a--c) The three one-way nests of one production run over their terrain: the 3\,km root, the 600\,m nest and the 120\,m nest, each centered on the 14.4\,km tile (dashed). (d) The run starts from the HRRR forecast state 3\,h before the scored hour, the 120\,m nest is born from the 600\,m state 1.5\,h later, and the 120\,m output of the last hour is averaged over $\pm$30\,min around the scored hour.}
  \label{fig:supp_woof_nesting}
\end{figure}

WOOF (v1.1.3)~\cite{woof} is a regional numerical weather model written for GPUs that follows the equations, physics options and file formats of WRF~\cite{skamarock2008wrf}. We use it as a physics-based downscaler of the same HRRR forecasts that the learned models see: each test sample is one independent run, nested from 3\,km down to 120\,m around its tile (\cref{fig:supp_woof_nesting}).

\paragraph{Physics.}
We run WOOF's default suite for sub-kilometer grids, which is HRRR's physics family~\cite{dowell2022hrrr}: Thompson microphysics, the MYNN surface layer and boundary-layer scheme, the RUC land-surface model and RTE+RRTMGP radiation (called every 12\,min). In the 120\,m domain the boundary-layer scheme is switched off and turbulence is resolved with a 3-D Smagorinsky closure, WOOF's large-eddy setting for nests at or below 250\,m. WOOF therefore differs from HRRR mostly in resolution and surface detail, not in physics.

\paragraph{Nesting.}
The three domains sit on HRRR's Lambert projection with a 5:1 ratio between levels: a 3\,km root of $64\times64$ cells (192\,km across, time step 15\,s), a 600\,m nest of $70\times70$ cells (42\,km, 3\,s) and a 120\,m nest of about $160\times170$ cells ($19\times20$\,km, 0.6\,s). The root, the 600\,m nest and the 120\,m nest extend 83, 11 and 1.2\,km past the tile, so that air entering from a coarser domain adjusts before it reaches the tile. Terrain comes from the 30\,m Copernicus DEM~\cite{copernicus2021glo30}, land cover from the 100\,m CGLC-MODIS-LCZ map~\citesupp{demuzere2022lcz,demuzere2023cglcmodislcz} and soil texture from WRF's default 30$''$ (about 1\,km) data. The hourly HRRR forecasts of the same run supply the lateral boundaries of the root.

\paragraph{Initialization and timing.}
For a sample with HRRR initialization $t_0$ and lead $\ell$, the run starts from the three-dimensional HRRR forecast state (native levels and soil) at lead $\ell-3$\,h of the same forecast. The first 3\,h are spin-up, in which the fine-scale flow and the near-surface temperature and humidity develop from HRRR's smooth 3\,km state. The 120\,m nest is created from the 600\,m state 1.5\,h into the run, and the run continues 30\,min past the scored hour. Because a run cannot start before the HRRR analysis, WOOF needs $\ell\ge3$\,h, which excludes about 5\% of the samples.

\begin{figure}[t]
  \centering
  \includegraphics[width=\columnwidth]{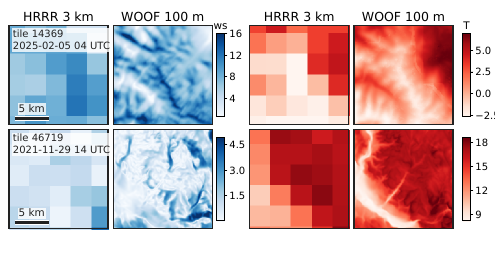}
  \caption{\textbf{WOOF output examples.} HRRR at 3\,km and the scored WOOF field at 100\,m ($\pm$30\,min mean) for 10\,m wind speed (m\,s$^{-1}$) and 2\,m temperature ($^\circ$C) on two tiles of \cref{fig:qualitative_det}: the Wasatch Range, Utah, on a winter night (top: 2025-02-05 04\,UTC, lead 46\,h, terrain 1.9--3.3\,km) and the coastal hills near Santa Maria, California, before dawn (bottom: 2021-11-29 14\,UTC, lead 32\,h). Color scales are shared within each pair.}
  \label{fig:supp_woof_examples}
\end{figure}

\paragraph{Output and scoring.}
The 120\,m domain writes 2\,m temperature, 2\,m dew point (from the 2\,m mixing ratio and surface pressure) and 10\,m wind every 5\,min. Winds are rotated from the model grid to east and north components, and every field is resampled to the tile's $144\times144$ pixel centers with the same cubic convolution we use to lift HRRR to 100\,m. We then average the 13 outputs within $\pm$30\,min of the scored hour, which matches the aggregation of the station observations: a single 120\,m snapshot carries minute-scale eddies that the stations average out, and on a 50-tile pilot the averaged field lowered wind-speed MAE by 0.12\,m\,s$^{-1}$ relative to a snapshot and left temperature and dew point unchanged. The 100\,m fields then enter our evaluation pipeline in place of a model prediction, so WOOF is scored at the same stations, with the same metrics, as every other model (\cref{fig:supp_woof_examples}).

\paragraph{Configuration choices.}
We fixed the setup on a 50-tile pilot before running the test sets (\cref{tab:woof_ablation}). \emph{Static data}: the 100\,m land-cover map removes the $\sim$1\,km blocks that WRF's default land-use data leaves in the temperature and humidity fields, at station errors within the tile-to-tile scatter of the default; higher-resolution soil texture (SoilGrids) raised the dew-point error by 0.13--0.17\,$^\circ$C, likely because HRRR's soil moisture is copied unchanged onto the new texture classes, and is not used. \emph{Spin-up}: temperature and dew point are flat from 2\,h on, and 3\,h reduced the wind-speed overestimate compared with 2\,h. Longer spin-ups gain little for their cost: 4--6\,h lower the wind-speed MAE by at most 0.05\,m\,s$^{-1}$, within the tile-to-tile scatter, and do not improve temperature or dew point, while each run costs 1.4--2.1 times as much. Since a run needs a lead at least as long as its spin-up, each extra hour would also exclude a further 2--3\% of the test samples, those at the shortest leads. \emph{Late start of the 120\,m nest and its 1.2\,km buffer}: together they gave about 1.8 times more tiles per GPU-hour, with dew point slightly better and $u_{10}$ 0.04\,m\,s$^{-1}$ worse.

\begin{table}[t]
  \centering
  \small
  \setlength{\tabcolsep}{3pt}
  \resizebox{\tabfitwidth}{!}{%
  \begin{tabular}{l cccc cccc}
  \toprule
   & \multicolumn{4}{c}{MAE $\downarrow$} & \multicolumn{4}{c}{$R_{\mathrm{all}}$ $\uparrow$} \\
  \cmidrule(lr){2-5}\cmidrule(lr){6-9}
   & ws & $u_{10}/v_{10}$ & $T$ & $T_d$ & ws & $u_{10}/v_{10}$ & $T$ & $T_d$ \\
  \midrule
  \multicolumn{9}{l}{\emph{Static surface data} (50 tiles, 3\,h spin-up)} \\
  \quad HRRR & 2.171 & 1.777 & 1.938 & 2.773 & 0.312 & 0.214 & 0.439 & 0.034 \\
  \quad Default & 2.407 & 1.856 & 1.845 & \textbf{2.598} & \textbf{0.519} & \textbf{0.454} & \textbf{0.541} & \textbf{0.180} \\
  \quad CGLC$^{*}$ & \textbf{2.343} & \textbf{1.820} & 1.867 & 2.638 & 0.510 & 0.448 & 0.529 & 0.172 \\
  \quad CGLC + SoilGrids & 2.362 & 1.828 & \textbf{1.833} & 2.772 & 0.507 & 0.437 & 0.506 & 0.149 \\
  \addlinespace
  \multicolumn{9}{l}{\emph{Spin-up} (7 tiles, CGLC + SoilGrids)} \\
  \quad HRRR & 2.461 & 2.064 & 1.575 & 5.015 & 0.283 & 0.180 & 0.511 & 0.030 \\
  \quad 3\,h$^{*}$ (1.00$\times$) & 2.826 & 2.109 & \textbf{1.641} & \textbf{4.742} & 0.471 & \textbf{0.449} & 0.611 & \textbf{0.113} \\
  \quad 4\,h (1.38$\times$) & 2.784 & \textbf{2.093} & 1.674 & 4.895 & 0.488 & 0.446 & \textbf{0.613} & 0.028 \\
  \quad 5\,h (1.73$\times$) & 2.794 & 2.111 & 1.698 & 4.890 & \textbf{0.490} & 0.444 & 0.581 & 0.068 \\
  \quad 6\,h (2.06$\times$) & \textbf{2.777} & 2.111 & 1.648 & 4.914 & 0.480 & 0.441 & 0.587 & 0.097 \\
  \bottomrule
  \end{tabular}}
  \caption{WOOF ablations. Station MAE (m\,s$^{-1}$ for ws and $u_{10}/v_{10}$, $^\circ$C for $T$ and $T_d$) and $R_{\mathrm{all}}$ (station-pair correlation averaged over six bands from 0.1 to 20.4\,km), computed on the
  $\pm$30\,min mean. \emph{Top}: static surface data on 50 test tiles (359 station reports) with a 3\,h spin-up. Default: WRF's $\sim$1\,km (30$''$) land use and soil; CGLC: 100\,m
  CGLC-MODIS-LCZ land cover with the default soil; CGLC + SoilGrids: CGLC with 250\,m SoilGrids soil. \emph{Bottom}: spin-up length on 7 tiles (107 reports), every run verifying at the
  same hour. The run time of each spin-up relative to the 3\,h spin-up is in parentheses next to its name. HRRR rows are the reference at the same stations.
  Bold: best WOOF value per column within each block. $^{*}$: the setting used for the WOOF baseline. With this few tiles, most differences between rows are within the tile-to-tile
  scatter.}
  \label{tab:woof_ablation}
\end{table}

\paragraph{Compute.}
On the 2017--2025 test set, runs were packed three to four per GPU on NVIDIA H100, A100 and L40S cards and a sample cost about 0.16 GPU-hours, failed attempts included, or about 7{,}100 GPU-hours for the set. Running Vernier on the same 50k samples, with data loading, the prior and scoring included, takes about 10 GPU-hours (H100) for the 64-member Vernier-Prob and 2.5 for Vernier-Det, so WOOF uses about 700 and 2{,}800 times as many GPU-hours. The 2026 sets ran eight to a GPU under CUDA MPS, at a lower cost per sample.

\paragraph{Failed runs and coverage.}
About 3\% of runs stopped with a numerical failure, almost all of them in the land-surface model when a run started over snow: 8.6\% of January samples failed against 0.2\% of August samples. Leaving them out would score WOOF on easier weather than the other models. We therefore re-ran every such sample with WOOF's alternative snow treatment (the RUC snow option of WRF 4.6.1), changing nothing else, and more than 90\% of the re-runs completed. The remaining gaps are samples whose HRRR input is unusable (the root domain extends past the HRRR grid, or HRRR hours are missing from the archive), samples whose terrain is too steep for a stable run, and re-runs that failed again. WOOF has a forecast for 96.6\% of the eligible samples of the 2017--2025 test set and 99.2\% of CONUS 2026, and is scored on those samples. On the 2017--2025 set, HRRR scored on this subset differs from HRRR on the full set by at most 0.04\,$^\circ$C in temperature and dew point and 0.006 in $R_{\mathrm{all}}$, and the subset is, if anything, slightly harder.

\paragraph{Comparison with the developer's scoreboard.}
Recast Systems publish a rolling scoreboard of WOOF against the HRRR forecast it starts from, over the previous 30 days, at ASOS airport stations, as RMSE pooled over all forecast hours, for 2.25\,km and 750\,m nests over four cities (and a 250\,m nest for San Francisco Bay fog). Their numbers are not directly comparable with ours: different stations, metric, period, grid spacing and spin-up. The error magnitudes agree with what we see (about 2\,$^\circ$C in temperature and dew point and 1.5--2\,m\,s$^{-1}$ in wind speed, with WOOF and HRRR within a few tenths of each other), but the sign of the difference does not consistently match. On their 750\,m nests WOOF's wind-speed RMSE is 3--9\% above HRRR's in Los Angeles, New York and Toronto, as in our results, but below it in the San Francisco case (1.54 against 1.64\,m\,s$^{-1}$); temperature is better than HRRR in New York and Toronto and worse in San Francisco and Los Angeles; and dew point is significantly worse than HRRR in three of the four cities, where ours is better. We therefore do not claim agreement beyond the magnitude of the errors.

\subsection{ConvCNP}
\label{supp:convcnp}
\paragraph{Architecture.}
We adapt the ConvCNP station downscaler of Sousa~\etal~\cite{sousa2026earth}, which builds on the architecture of Vaughan~\etal~\cite{vaughan2022convcnp}. The context is the full-domain HRRR forecast at 3\,km for one (initialization, lead) pair. It holds eight z-scored fields: $T$, $T_d$, $u_{10}$, $v_{10}$, cloud cover, skin temperature, surface pressure and soil moisture, corresponding to the four fields to be downscaled, and four auxiliary fields used in the main model. Constant channels add latitude, longitude, $\cos$/$\sin$ of day-of-year and hour, and lead\,/\,72\,h, following~\cite{sousa2026earth}. A residual CNN encodes this grid at full resolution, with no striding: a $3{\times}3$ convolution to 128 channels, three residual blocks of two $3{\times}3$ convolutions each, and a $1{\times}1$ projection. Its receptive field is 7 cells ($\sim$21\,km) on each side. The feature map is read bilinearly at each target location. The readout is concatenated with a \emph{station descriptor}~$\mathbf{d}$ and decoded by a 3-layer ReLU MLP of width 128. $T$ and $T_d$ each get a Gaussian head $(\mu, \log\sigma^2)$. The wind $(u_{10}, v_{10})$ gets a single bivariate Gaussian with a learned correlation $\rho = 0.99\tanh(\cdot)$, and wind speed is computed from samples of this joint distribution. Heads are initialized to the training climatology. The descriptor is the only information below the 3\,km grid scale that the decoder receives. Because it is a function of location alone, the trained model can be evaluated at any 100\,m pixel.

\paragraph{Descriptors.}
Every descriptor is built from a 64-dimensional geospatial foundation-model embedding on a common 100\,m grid. We test two embeddings, AlphaEarth Foundations~\cite{brown2025alphaearth} and OlmoEarth v1.3~\cite{tseng2026olmoearth}, using the same annual layer as Vernier (\cref{supp:surface}). Terrain features come from GLO-30 resampled to 100\,m: elevation and the three components of the Earth-frame surface normal (as used in Vernier). \Cref{tab:convcnp_descriptors} lists the three descriptors, and \cref{tab:convcnp_ablation} compares them. Each descriptor column is z-scored with statistics from the training stations.

\begin{table}[h]
  \centering
  \small
  \begin{tabular}{@{}lp{0.56\linewidth}r@{}}
    \toprule
    Descriptor & Contents & Dim. \\
    \midrule
    Pixel embedding & Embedding vector of the target's 100\,m cell & 64 \\
    Pixel stack     & Pixel embedding + elevation + surface normals & 68 \\
     VAE             & Frozen patch-VAE posterior mean of the $7{\times}7$ (700\,m) embedding patch & 16 \\
    \bottomrule
  \end{tabular}
  \caption{ConvCNP station descriptors. Each is built on either embedding (AlphaEarth or OlmoEarth).}
  \label{tab:convcnp_descriptors}
\end{table}

\paragraph{Patch VAE.}
The VAE descriptor follows~\cite{sousa2026earth}, adapted to our 100\,m resolution. The encoder reduces a $7{\times}7$ patch to $1{\times}1$ with three valid $3{\times}3$ convolution stages (128/256/512 channels, each with BatchNorm, ReLU and dropout 0.1). Linear layers then give the mean and log-variance of a 16-d latent. The decoder mirrors the encoder. The objective is reconstruction MSE, plus a spatial-gradient reconstruction term (weight 0.5), plus a KL term whose weight ramps linearly to $\beta = 5{\times}10^{-4}$ over 5k steps. Auxiliary regressions of elevation (weight 1) and latitude/longitude (weight 0.5 each) from the latent are added following~\cite{sousa2026earth}.
One VAE per embedding is trained on station-centered patches from the training split, for 25k steps at batch size 1024, and then frozen. Because the station loss never reaches the encoder, the decoder cannot learn to fingerprint individual training stations through their descriptor. This was a failure mode we saw when the descriptor encoder was trained end-to-end.

\paragraph{Physics-prior descriptor.}
In the prior variant, four time-varying columns are appended to the static descriptor: the physics prior's
$(u_{10}, v_{10}, T, T_d)$ at the target (\supp~\ref{supp:prior}), computed from the same forecast as the context and z-scored with the context grid's statistics. However, this is prohibitively expensive. The prior is defined on 100\,m crops, and its wind component needs a 3D mass-consistent solve on each one. Every ConvCNP training episode spans the whole domain, so each target station would need its own crop and solve at every step. Pre-computing the prior for every (initialization, lead) pair is also impractical. We avoid both because, with every wind term that is nonlinear in the wind switched off, the station wind prior is exactly linear in the native HRRR winds for fixed terrain. It therefore factorizes as $s(\Delta T)\,K\mathbf{w}$. Here $\mathbf{w}$ is the $10{\times}10$ block of native HRRR wind cells around the station, and $s$ is the prior's pointwise bulk speed factor. We compute each station's kernel $K \in \mathbb{R}^{2\times2\times10\times10}$ once, by automatic differentiation of the full solver, including the bicubic lift. The thermodynamic prior is pointwise, so it is evaluated exactly from the station's interpolation weights and its elevation anomaly. During training the prior then costs one small tensor contraction per target. On identical inputs the kernel reproduces the full solve to about $10^{-4}$\,m\,s$^{-1}$; end to end through the data pipeline, the median difference is 0.003\,m\,s$^{-1}$ (99th percentile 0.03\,m\,s$^{-1}$). At validation the prior is evaluated densely at every pixel of each crop.

\begin{table}[t]
  \centering
  \footnotesize
  \setlength{\tabcolsep}{2.2pt}
  \begin{tabular}{@{}ll ccccc@{}}
    \toprule
    Embedding & Descriptor & $u_{10}$ & $v_{10}$ & WS & $T$ & $T_d$ \\
    \midrule
    \multicolumn{2}{@{}l}{HRRR (MAE)} & 1.680 & 1.691 & 2.015 & 1.719 & 2.388 \\
    \midrule
    \multirow{3}{*}{AlphaEarth}
      & Pixel embedding & 0.576 & 0.598 & 0.622 & 0.976 & 1.147 \\
      & Pixel stack     & 0.582 & 0.603 & 0.637 & 0.971 & 1.133 \\
      & VAE             & 0.598 & 0.630 & 0.679 & 1.023 & 1.178 \\
    \midrule
    \multirow{3}{*}{OlmoEarth}
      & Pixel embedding & 0.582 & 0.598 & 0.626 & 0.996 & 1.136 \\
      & Pixel stack     & 0.578 & 0.594 & \textbf{0.617} & 0.967 & \textbf{1.128} \\
      & VAE             & 0.614 & 0.635 & 0.706 & 1.027 & 1.184 \\
    \midrule
    OlmoEarth & Pixel stack + prior & \textbf{0.571} & \textbf{0.591} & 0.622 & \textbf{0.966} & 1.145 \\
    \bottomrule
  \end{tabular}
  \caption{ConvCNP descriptor ablation: CRPS on the 50k fixed samples of the 2017--2025 spatial holdout (held-out test stations, \cref{tab:extreme}; wind in m\,s$^{-1}$, temperatures in $^\circ$C; lower is
  better). WS is wind speed, derived from the joint $(u_{10}, v_{10})$ samples. HRRR
  is a point forecast, so its CRPS equals its MAE. Best per column in bold. The last row is the model used in
  \cref{tab:results,tab:extreme}.}
  \label{tab:convcnp_ablation}
\end{table}

\paragraph{Training.}
Each training example is an \emph{episode}: one (initialization, lead) pair, its full-domain context grid, and every training station with a valid observation at that valid hour as a target. Held-out stations are removed from the targets only; the context always covers the whole domain. For each variable, the Gaussian NLL is averaged over the valid targets within an episode and then over episodes; the per-variable losses are summed. We train with Adam at learning rate $2{\times}10^{-4}$ for the pixel descriptors and $1{\times}10^{-4}$ for the VAE,
with weight decay $10^{-4}$, gradient clipping at 1 and a constant schedule. Each step uses 4 episodes per GPU on 4 GPUs, with the CNN in bf16.
Training runs for up to 60k steps. The checkpoint is selected on mean ensemble CRPS over $T$, $T_d$, $u_{10}$, $v_{10}$ and wind speed ($K{=}4$ members), with early stopping and patience of 8 validations.
ConvCNP trains for far more steps than the other models (\cref{tab:supp_convcnp_impl}) because each step sees only 16 valid hours, and its small network keeps improving until the 60k cap.

\begin{table}[t]
  \centering
  \footnotesize
  \supptabfit{\columnwidth}{%
  \begin{tabular}{@{}ll@{}}
    \toprule
    Parameters & 0.98M \\
    Optimizer & Adam, weight decay $10^{-4}$, gradient clip 1 \\
    Peak learning rate & $2\times10^{-4}$ \\
    Learning-rate schedule & Constant, no warm-up \\
    Training steps & 60k \\
    Global batch size & 16 episodes (4 per GPU) \\
    Samples seen & 960k episodes \\
    Hardware & 4 $\times$ NVIDIA H100 \\
    Wall-clock time & 19.7\,h \\
    GPU-hours (H100) & 79 \\
    Checkpoint used & Best validation CRPS (step 60k) \\
    Ensemble size at evaluation & 64 \\
    \bottomrule
  \end{tabular}
  }%
  \caption{ConvCNP implementation details. An episode is one valid hour over the full domain, with every valid training station as a target (hundreds to about 1,500).}
  \label{tab:supp_convcnp_impl}
\end{table}

\paragraph{Choice of the prior variant.}
On CRPS, the prior variant matches the best descriptor without the prior (\cref{tab:convcnp_ablation}): it is better for $u_{10}$, $v_{10}$ and $T$, and worse for wind speed (+0.8\%) and $T_d$ (+1.5\%).
Its single members also match the observed station-pair differences slightly better for wind speed and $T$ (for wind speed, a correlation of 0.26 against 0.24 at 2--3\,km) and equally for $T_d$.
It also gives ConvCNP the same physical information that Vernier receives.
We therefore use the prior variant in the main comparison (\cref{tab:results}).

\paragraph{Evaluation.}
We evaluate the ConvCNP with the same validator and fixed sample set as our model. Each validation crop is decoded at every 100\,m pixel, using the HRRR context around the crop plus an 8-cell halo, which covers the CNN encoder's receptive field. The VAE descriptor additionally receives a 3-pixel margin so that edge pixels see their true neighborhood. Station metrics and spatial-structure metrics are therefore computed on identical (station, hour, lead) sets.

\subsection{U-Net}
\label{supp:unet}
The U-Net baseline is a convolutional encoder--decoder in the style of the CorrDiff regression network~\cite{mardani2025corrdiff}: four resolution levels with channel multipliers $(1,2,2,4)$ on 96 base channels, two residual blocks per level, self-attention blocks (8 heads), and reflect padding, which avoids the edge artifacts of zero padding. It receives the same inputs as Vernier at 100\,m (the HRRR forecast lifted to 100\,m, terrain, OlmoEarth embeddings and the physics prior) and the same context embedding, and is trained with the same station loss as Vernier-Det (\cref{supp:deterministic}). It has 51.9M parameters and is trained for 8{,}000 steps with a batch of 140 tiles per GPU on 8 GPUs, i.e., the same 8.96M training samples as Vernier-Det, with the same optimizer settings. The HRRR station correction baseline is the same U-Net with the HRRR forecast as its only input (51.8M parameters). \Cref{tab:supp_unet_impl} lists both configurations.

\begin{table}[t]
  \centering
  \footnotesize
  \supptabfit{\columnwidth}{%
  \begin{tabular}{@{}lp{0.27\columnwidth}p{0.27\columnwidth}@{}}
    \toprule
     & U-Net & HRRR station corr. \\
    \midrule
    Parameters & 51.9M & 51.8M \\
    Optimizer & \multicolumn{2}{p{0.56\columnwidth}@{}}{AdamW ($\beta_1=0.9$, $\beta_2=0.999$, weight decay 0.01), gradient clip 1, bf16} \\
    Peak learning rate & \multicolumn{2}{p{0.56\columnwidth}@{}}{$4.2\times10^{-4}$ ($2\times10^{-4}$ scaled by $\sqrt{1120/256}$)} \\
    Learning-rate schedule & \multicolumn{2}{l@{}}{500 warm-up steps, cosine decay to zero} \\
    Training steps & \multicolumn{2}{l@{}}{8k, from scratch} \\
    Global batch size & \multicolumn{2}{l@{}}{1,120 tiles (140 per GPU)} \\
    Samples seen & \multicolumn{2}{l@{}}{8.96M} \\
    Hardware & \multicolumn{2}{l@{}}{8 $\times$ NVIDIA H100} \\
    Wall-clock time & 9.5\,h & 9.3\,h \\
    GPU-hours (H100) & 76 & 74 \\
    Checkpoint used & \multicolumn{2}{l@{}}{Last (step 8k), EMA weights (decay 0.995)} \\
    \bottomrule
  \end{tabular}
  }%
  \caption{U-Net implementation details. Wall-clock time includes in-loop validation.}
  \label{tab:supp_unet_impl}
\end{table}

\subsection{CorrDiff}
\label{supp:corrdiff}
CorrDiff~\cite{mardani2025corrdiff} downscales a coarse forecast in two steps: a regression U-Net predicts the conditional mean of the fine field, and a diffusion model generates the residual around that mean. Both are trained on dense high-resolution analyses. We keep this mean-plus-generated-residual structure and adapt it to station supervision.

\paragraph{Mean.}
The mean is the physics prior $P$ of \cref{sec:physical_prior} rather than a trained regression network. There is no dense 100\,m target to train the regression on, and using the same prior as Vernier makes the generative part the only difference between the two models: both learn the same correction $Y-P$ from the same inputs.

\paragraph{Generative residual with flow matching.}
Instead of EDM diffusion~\citesupp{karras2022edm}, the residual is generated with flow matching (FM)~\citesupp{lipman2023flowmatching}. The network $v_\theta(x_t,t,c)$ predicts the velocity of a straight path from Gaussian noise $x_0\sim\mathcal{N}(0,I)$ to the target $x_1$: with $x_t=(1-t)\,x_0+t\,x_1$ and $t\sim\mathcal{U}(0,1)$, the loss is $\|v_\theta(x_t,t,c)-(x_1-x_0)\|^2$, and sampling integrates $\dot x=v_\theta$ from $t=0$ to $1$ with Euler steps. Here $x_1$ has four channels, the residuals $(Y-P)/s$ of $T$, $T_d$, $u$ and $v$ with $s=(3\,{}^\circ\mathrm{C},5\,{}^\circ\mathrm{C},2.5,2.5\,\mathrm{m\,s^{-1}})$, and the conditioning $c$ holds the same inputs as Vernier: the HRRR forecast and its auxiliary fields lifted bicubically to 100\,m, the terrain, the OlmoEarth embedding and the prior itself, channel-concatenated with $x_t$. A member is $P+s\odot x_1$.

\paragraph{Sparse targets.}
The loss is evaluated only at station pixels and only for the variables each station reports, as in \cref{eq:station_loss}. The path $x_t$, however, must be formed at every pixel, so the 99.98\% of pixels without a station need a value in $x_1$. We set the residual there to zero, i.e., to the prior itself. This is the fill that is consistent at inference: away from stations the model has nothing to correct, a trained model generates a residual near zero, and the sampled trajectory then stays close to the training paths; it also makes the target mean zero over the tile.
This residual formulation is the same decomposition CorrDiff uses with a learned mean.

\paragraph{Network and training.}
The generator is the same U-Net as the regression baseline of \cref{supp:unet} (51.9M parameters, zero padding), conditioned on the flow time $t$, the hour, the solar insolation and the lead time through adaptive group normalization, with the same context embedders as Vernier. It is trained from scratch for 8k steps at a global batch of 1,440 tiles (11.5M samples) with AdamW and a peak learning rate of $2.4\times10^{-4}$ (\cref{tab:supp_corrdiff_impl}); a first run at twice this rate diverged after 1,500 steps (in-loop wind-speed fCRPS 1.38 against 0.64, spread--skill ratio 1.35). No spectral or physics loss is used. The in-loop validation fCRPS (1,000 samples, 4 members) reaches its minimum between 3k and 5k steps and rises slowly afterward: the mean fCRPS over $T$, $T_d$, $u$, $v$ and speed is 0.831 at 5k and 0.838 at 8k steps, and a run continued to 10k steps reaches 0.847 and is worse on the test stations (mean fCRPS 0.832 against 0.819 with 64 members). The baseline uses its final EMA weights.

\paragraph{Sampling.}
Each tile gets 64 members from independent noise, each integrated with 50 Euler steps (25 on the 2026 test set of \cref{tab:results}), with one prior solve per tile shared by all members.

\begin{table}[t]
  \centering
  \footnotesize
  \supptabfit{\columnwidth}{%
  \begin{tabular}{@{}lp{0.56\columnwidth}@{}}
    \toprule
    Parameters & 51.9M \\
    Optimizer & AdamW ($\beta_1=0.9$, $\beta_2=0.999$, weight decay 0.01), gradient clip 1, bf16 \\
    Peak learning rate & $2.4\times10^{-4}$ ($10^{-4}$ scaled by $\sqrt{1440/256}$) \\
    Learning-rate schedule & 500 warm-up steps, cosine decay to zero \\
    Training steps & 8k, from scratch \\
    Global batch size & 1,440 tiles (180 per GPU) \\
    Samples seen & 11.5M \\
    Hardware & 8 $\times$ NVIDIA H100 \\
    Wall-clock time & 17.0\,h (with in-loop validation) \\
    GPU-hours (H100) & 136 \\
    Checkpoint used & Last (step 8k), EMA weights (decay 0.995); later steps are worse \\
    Ensemble size at evaluation & 64 members, 50 Euler steps each (25 on the 2026 test set) \\
    \bottomrule
  \end{tabular}
  }%
  \caption{CorrDiff implementation details.}
  \label{tab:supp_corrdiff_impl}
\end{table}

\paragraph{Ablations.}
\Cref{tab:supp_corrdiff_ablation} tests two changes to the generative stage on the 2017--2025 test stations.
The first replaces the zero off-station target with a kriged one: a Gaussian radial-basis interpolation of the station residuals (simple kriging, 2\,km length scale), exact at the stations and decaying to zero, the prior, away from them.
It is worse than the baseline for every variable, most of all $T$ and $T_d$ (fCRPS 1.074 against 1.007 and 1.289 against 1.199), so we keep the zero fill.
The second adds a loss on the wind speed derived from the generated $u$ and $v$, a fifth scored quantity beside the four generated variables (L2, weight 0.04).
It stays close to the baseline but is not better for any variable (wind speed 0.661 against 0.659, mean 0.826 against 0.819), so it is not used.
A wind spectral ceiling as in Vernier (\cref{supp:spectral}) raised the in-loop wind-speed fCRPS (0.71 against 0.66) and was not pursued.

\begin{table}[t]
  \centering
  \footnotesize
  \setlength{\tabcolsep}{3pt}
  \begin{tabular}{@{}l rrrrr r@{}}
    \toprule
    & \multicolumn{6}{c}{fCRPS $\downarrow$} \\
    \cmidrule(lr){2-7}
    & ws & $T$ & $T_d$ & $u$ & $v$ & Mean \\
    \midrule
    Baseline & \textbf{0.659} & \textbf{1.007} & \textbf{1.199} & \textbf{0.602} & \textbf{0.627} & \textbf{0.819} \\
    + kriged fill & 0.669 & 1.074 & 1.289 & \underline{0.607} & \underline{0.630} & 0.854 \\
    + speed loss & \underline{0.661} & \underline{1.015} & \underline{1.209} & 0.609 & 0.634 & \underline{0.826} \\
    \bottomrule
  \end{tabular}
  \caption{\textbf{CorrDiff ablations} on the 2017--2025 test stations (50k samples, 184,595 station-hours; 64 members, 50 Euler steps). fCRPS in $^\circ$C and m\,s$^{-1}$; Mean averages the five. The baseline is the CorrDiff run of \cref{tab:results}. Bold: best; underline: second best.}
  \label{tab:supp_corrdiff_ablation}
\end{table}

\subsection{WeatherNext 3}
\label{supp:wn3}
WeatherNext~3 (WN3)~\cite{rasp2026weathernext} is a global 64-member probabilistic forecast. Its 2\,m temperature and dew point come from a station head trained on station observations and released on a 0.05$^\circ$ grid; its 10\,m wind comes from the 0.1$^\circ$ model fields. We use the 2026 forecasts initialized at 00, 06, 12 and 18\,UTC, cropped to CONUS, at leads 3--48\,h. The deterministic row uses the ensemble mean ($T$ and $T_d$ from the station-head fields, $u$ and $v$ from the 0.1$^\circ$ mean); the probabilistic row uses the 64 members. Fields are interpolated to the 100\,m pixel centers of each tile with cubic convolution (as we lift HRRR) and scored with the same validator, tiles and forecast times as our models.

\subsection{UW-WRF}
\label{supp:uwwrf}
UW-WRF is the University of Washington's operational WRF forecast for the Pacific Northwest~\cite{mass2003regional,uwwrf}. We use its innermost domain (d04, 1.33\,km), initialized at 00 and 12\,UTC, at leads 12--24\,h. Wind components are rotated from the model grid to the earth frame, and dew point is derived from the 2\,m mixing ratio and surface pressure. Fields are interpolated to the 100\,m pixel centers with cubic convolution on the model's own Lambert grid and scored on tiles inside d04, on the same tiles and forecast times as Vernier.

\subsection{WindNinja}
\label{supp:windninja}
WindNinja~\cite{windninja2025} is a diagnostic mass-conserving wind model used in wildland fire operations. We run it for every test tile from the HRRR forecast at the scored lead (10\,m wind, 2\,m temperature and cloud cover), on a 100\,m domain that extends the tile by a 4.8\,km halo, with GLO-30 terrain and land-cover-derived vegetation, and crop the solution back to the tile. It predicts wind only, so only its wind scores are reported. One tile takes about 30\,s on one CPU core.

\subsection{MOS and EMOS}
\label{supp:mos_config}
Model Output Statistics (MOS) and its probabilistic form, ensemble MOS (EMOS), correct HRRR at each station and serve as references only in the operational comparison (\cref{tab:operational}), where every test station has a 2017--2025 record; \cref{supp:mos} analyzes their results.

\paragraph{Station calibration.}
MOS~\cite{glahn1972mos} regresses the observation $y$ on the forecast $x$ at each station over a training archive.
For station $s$ and six-hour block $h$ of local solar time, we fit
\begin{equation}
  \begin{aligned}
    \hat y_{s,h}(x) &= a_{s,h} + b_{s,h}\,x,\\
    (a_{s,h}, b_{s,h}) &= \operatorname*{arg\,min}_{a,b} \sum_{n \in \mathcal{T}_{s,h}} \left(y_n - a - b\,x_n\right)^2,
  \end{aligned}
  \label{eq:mos}
\end{equation}
where $\mathcal{T}_{s,h}$ holds the station's 2017--2025 forecast--observation pairs in that block and $x$ is HRRR at the station's nearest 3\,km grid cell.
We report three variants: one line for all stations (no site information), one line per station (no blocks), and one line per station and block (station $\times$ time of day, ToD).
A fit with fewer than 50 pairs falls back to the next coarser one.
As a probabilistic counterpart, we use ensemble MOS (EMOS)~\cite{gneiting2005emos} in its simplest form, a Gaussian around the MOS line:
\begin{equation}
  F_{s,h}(\cdot \mid x) = \mathcal{N}\!\left(a_{s,h} + b_{s,h}\,x,\ \sigma_{s,h}^2\right),
  \label{eq:emos}
\end{equation}
where $\sigma_{s,h}^2 = \sum_{n \in \mathcal{T}_{s,h}} (y_n - a_{s,h} - b_{s,h}\,x_n)^2 / (|\mathcal{T}_{s,h}| - 2)$ is the residual variance of the fit; for wind speed, the Gaussian is truncated at zero~\citesupp{thorarinsdottir2010wind}.
Wind-speed MOS uses every valid hour of 2017--2025 at all 48 lead times (a median of $3.6\times10^5$ pairs per station).
Temperature and dew-point MOS use one HRRR initialization per day in 2017--2025 (00 or 12\,UTC, every lead hour in the archive; a median of $8.8\times10^4$ pairs per station) against the quality-controlled hourly observations.
No 2026 data enter any fit.

\paragraph{Scores.}
Deterministic forecasts are scored by the MAE and probabilistic ones by the CRPS, $\mathrm{CRPS}(F, y) = \mathbb{E}_F |X - y| - \tfrac{1}{2}\, \mathbb{E}_F |X - X'|$, which reduces to the absolute error for a deterministic forecast, so MAE and CRPS share units and can be compared directly.
For the Gaussian EMOS, with $z = (y-\mu)/\sigma$ and $\phi$ and $\Phi$ the standard normal density and distribution function~\cite{gneiting2005emos},
\begin{equation}
  \mathrm{CRPS}\!\left(\mathcal{N}(\mu,\sigma^2), y\right) = \sigma\left[z\left(2\Phi(z) - 1\right) + 2\phi(z) - \tfrac{1}{\sqrt{\pi}}\right],
  \label{eq:crps_gauss}
\end{equation}
and for the wind-speed EMOS truncated at zero, with $p = \Phi(\mu/\sigma)$~\citesupp{thorarinsdottir2010wind},
\begin{multline}
  \mathrm{CRPS}\!\left(\mathcal{N}_{0}(\mu,\sigma^2), y\right) = \frac{\sigma}{p^{2}}\Big[z\,p\left(2\Phi(z) + p - 2\right) \\
    + 2\phi(z)\,p - \tfrac{1}{\sqrt{\pi}}\,\Phi\!\left(\sqrt{2}\,\mu/\sigma\right)\Big].
  \label{eq:crps_tn}
\end{multline}
Vernier-Prob's $K = 64$ members are scored with the fair CRPS (\cref{eq:fair_score})~\cite{ferro2014fair}, as in \cref{tab:operational}.
The fair CRPS is an unbiased estimate of the CRPS of the distribution the members are drawn from, so it can be compared directly with the exact CRPS of EMOS: scoring 64 EMOS draws with the fair CRPS reproduces the exact values to within 0.001, whereas the standard ensemble CRPS would penalize them by 1.4--1.6\%.
$R_{\mathrm{pair}}$ is computed as in \cref{tab:operational}, on the same station pairs. Member $R_{\mathrm{pair}}$ averages it over the individual members, and for EMOS over 64 draws sampled independently at each station-hour.

\section{Additional Results}
\label{supp:results}

\subsection{Stratified results}
\label{supp:strat_bias}
\begin{figure}[!t]
  \centering
  \includegraphics[width=\columnwidth]{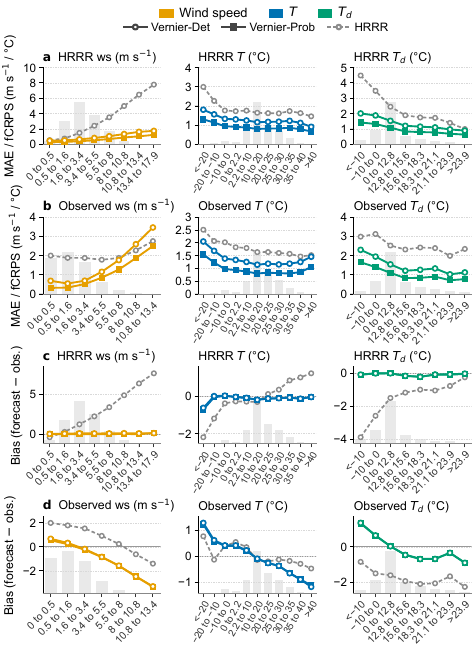}
  \caption{\textbf{Stratified errors and bias on the HRRR domain in 2026} (the 50k samples of \cref{tab:operational}). \textbf{a--b}: errors in bins of the HRRR forecast (\textbf{a}) and of the station observation (\textbf{b}): Vernier-Det MAE, Vernier-Prob fCRPS and HRRR MAE. \textbf{c--d}: bias (forecast minus observation) in the same bins: Vernier-Det, the Vernier-Prob ensemble mean and HRRR. Open circles show Vernier-Det, filled squares Vernier-Prob and gray dashed lines HRRR. Gray bars show each bin's share of station-hours; bins with fewer than 100 station-hours are excluded. In HRRR-forecast bins (\textbf{c}), Vernier's bias stays within 0.2\,m\,s$^{-1}$ for wind and 0.21\,$^\circ$C for $T$ and $T_d$ in every bin except $T$ below $-20\,^\circ$C ($-0.6\,^\circ$C for Vernier-Det, $-0.7\,^\circ$C for Vernier-Prob), while HRRR's conditional bias reaches $+7.6$\,m\,s$^{-1}$, $-2.2$/$+1.2\,^\circ$C and $-3.9\,^\circ$C. In observed bins (\textbf{d}) every forecast is too low in the highest bin and too high in the lowest (regression to the mean), except HRRR's dew point, which is too low in every bin; Vernier underestimates observed winds above 8\,m\,s$^{-1}$ by 2.5--3.3\,m\,s$^{-1}$.}
  \label{fig:strat_detail}
\end{figure}
\Cref{fig:strat_detail} breaks down the errors of \cref{tab:operational} (HRRR domain, 2026) in bins of the HRRR forecast and of the station observation. Binned by the HRRR forecast, both Vernier models improve on HRRR in every bin, most where HRRR is most wrong: for HRRR wind forecasts of 13.4--17.9\,m\,s$^{-1}$, HRRR's MAE is 7.8\,m\,s$^{-1}$ against 1.8\,m\,s$^{-1}$ for Vernier-Det. Binned by the observation, Vernier-Det's MAE exceeds HRRR's only for observed winds above 8\,m\,s$^{-1}$, where it underestimates the speed, while Vernier-Prob's fCRPS stays below HRRR's MAE in every bin. Panels c and d of \cref{fig:strat_detail} show the bias in the same bins. Binned by the HRRR forecast, Vernier removes HRRR's conditional bias: HRRR's wind bias grows with its forecast speed to +7.6\,m\,s$^{-1}$, while Vernier's stays within 0.2\,m\,s$^{-1}$ in every bin. Binned by the observation, Vernier regresses toward the mean, as expected of a forecast of the conditional mean: its wind is 0.6--0.7\,m\,s$^{-1}$ too high for observed calms below 0.5\,m\,s$^{-1}$ and 2.5--3.3\,m\,s$^{-1}$ too low above 8\,m\,s$^{-1}$. HRRR's wind is instead too high for every observed speed below 8\,m\,s$^{-1}$, by up to 2.0\,m\,s$^{-1}$, and its dew point is too low in every bin.

\Cref{tab:strat_network} splits the same errors by station network. Vernier-Det improves on HRRR in every network for all three variables. The wind gain is smallest at airports, whose anemometers follow the 10\,m open-exposure standard (MAE 1.35 to 1.09\,m\,s$^{-1}$), and largest at CWOP home stations, which are often sheltered and where HRRR's wind MAE is 2.48\,m\,s$^{-1}$ (0.68\,m\,s$^{-1}$ for Vernier-Det). CWOP and MesoWest stations together carry 89\% of the wind station-hours.

\begin{table}[h]
  \centering
  \small
  \setlength{\tabcolsep}{4pt}
  \supptabfit{\columnwidth}{%
  \begin{tabular}{lrcccc}
    \toprule
    & & \multicolumn{2}{c}{Wind (m\,s$^{-1}$)} & $T$ ($^\circ$C) & $T_d$ ($^\circ$C) \\
    \cmidrule(lr){3-4}
    Network & Share & MAE & Bias & MAE & MAE \\
    \midrule
    Airports & 4.7\% & 1.09\,/\,1.35 & $-0.06$ & 1.27\,/\,1.58 & 1.48\,/\,1.92 \\
    RAWS & 4.9\% & 0.89\,/\,1.77 & $-0.03$ & 1.45\,/\,1.76 & 1.73\,/\,2.25 \\
    CWOP & 26.8\% & 0.68\,/\,2.48 & $+0.29$ & 1.28\,/\,1.64 & 1.41\,/\,2.60 \\
    MesoWest & 62.2\% & 0.71\,/\,1.72 & $+0.06$ & 1.12\,/\,1.68 & 1.56\,/\,2.64 \\
    DOT & 0.9\% & 0.84\,/\,1.87 & $-0.06$ & 1.33\,/\,1.66 & 1.56\,/\,2.14 \\
    Other & 0.5\% & 0.80\,/\,2.16 & $+0.04$ & 1.20\,/\,1.58 & 1.51\,/\,2.05 \\
    \bottomrule
  \end{tabular}
  }%
  \caption{\textbf{Errors by station network on the CONUS 2026 set} (Vernier-Det, the 50k samples of \cref{tab:operational}). The network comes from the station's data provider: airports (ASOS/AWOS), RAWS (fire-weather stations), CWOP (citizen home stations), MesoWest (other mesonets relayed by MesoWest), DOT (road-weather stations) and other. Share is the network's share of wind station-hours. MAE is given as Vernier-Det\,/\,HRRR; bias is the Vernier-Det forecast minus the observation.}
  \label{tab:strat_network}
\end{table}

\subsection{Additional holdout results}
\label{supp:ecoregion_splits_results}
We report two further holdouts. The \emph{spatial holdout} scores the models at the same held-out test stations as \cref{tab:results} but over 2017--2025, the training years. The \emph{ecoregion holdouts} withhold an entire region from training: the Great Plains and the Cold Deserts of the Intermountain West (holdout region maps in \cref{fig:supp_ecoregion_splits}). Each region is split into non-overlapping validation and test portions, a 20\,km buffer separates it from the training stations, and the models are retrained without it. \Cref{tab:extreme} reports the metrics of \cref{tab:results} except the de-meaned $R_{\mathrm{pair}}$.

\ifdefined\supptabbox\else\newsavebox{\supptabbox}\fi
\begin{table}[!t]
\centering
\footnotesize
\aitwotable{tab:extreme}%
\definecolor{ai2pinkrow}{HTML}{F0529C}%
\setlength{\tabcolsep}{2pt}
\sbox{\supptabbox}{%
\begin{tabular}{l cccc ccc ccc}
\toprule
\emph{Deterministic} & \multicolumn{4}{c}{MAE $\downarrow$} & \multicolumn{3}{c}{Bias $\to 0$} & \multicolumn{3}{c}{$R_{\mathrm{pair}}$ $\uparrow$} \\
\cmidrule(lr){2-5}\cmidrule(lr){6-8}\cmidrule(lr){9-11}
Model & ws & dir & $T$ & $T_d$ & ws & $T$ & $T_d$ & ws & $T$ & $T_d$ \\
 & {\scriptsize m\,s$^{-1}$} & {\scriptsize $^\circ$} & {\scriptsize $^\circ$C} & {\scriptsize $^\circ$C} & {\scriptsize m\,s$^{-1}$} & {\scriptsize $^\circ$C} & {\scriptsize $^\circ$C} & {\scriptsize pairs} & {\scriptsize pairs} & {\scriptsize pairs} \\
\midrule
\multicolumn{8}{l}{\emph{Spatial holdout, 2017--2025}} & {\scriptsize 99k} & {\scriptsize 108k} & {\scriptsize 106k} \\
\quad HRRR interp. & 2.05 & 35.85 & 1.72 & 2.39 & +1.75 & +0.13 & $-$1.24 & 0.07 & 0.26 & 0.18 \\
\quad HRRR stn.\ corr. & 1.03 & 34.11 & 1.43 & 1.64 & +0.08 & +0.10 & \underline{$-$0.03} & 0.11 & 0.29 & 0.21 \\
\quad U-Net & 0.87 & 33.80 & 1.31 & 1.58 & \textbf{+0.01} & \underline{+0.03} & \textbf{$-$0.02} & 0.41 & 0.52 & 0.39 \\
\quad ConvCNP$^{\dagger}$ & 0.89 & \underline{32.25} & 1.35 & 1.60 & +0.15 & \textbf{0.00} & $-$0.10 & 0.43 & 0.47 & 0.39 \\
\quad CorrDiff$^{\dagger}$ & 0.87 & 34.45 & 1.36 & 1.63 & $-$0.26 & +0.19 & $-$0.08 & 0.45 & 0.48 & 0.36 \\
\rowcolor{ai2pinkrow!10}\quad\textbf{Vernier-Det} & \underline{0.84} & \textbf{32.24} & \underline{1.27} & 1.52 & +0.04 & +0.06 & +0.05 & \textbf{0.48} & 0.53 & \underline{0.43} \\
\rowcolor{ai2pinkrow!10}\quad\quad\emph{+ Spectral} & 0.87 & \underline{32.25} & \underline{1.27} & 1.52 & \underline{+0.02} & +0.06 & +0.05 & 0.45 & 0.53 & \textbf{0.44} \\
\rowcolor{ai2pinkrow!10}\quad\textbf{Vernier-Prob}$^{\dagger}$ & \textbf{0.82} & 32.52 & \underline{1.27} & \textbf{1.51} & $-$0.05 & +0.05 & +0.05 & \textbf{0.48} & \textbf{0.54} & \underline{0.43} \\
\rowcolor{ai2pinkrow!10}\quad\quad\emph{+ Spectral} & 0.85 & 32.68 & \textbf{1.26} & \textbf{1.51} & $-$0.07 & +0.05 & +0.04 & 0.44 & \textbf{0.54} & \underline{0.43} \\
\addlinespace
\multicolumn{8}{l}{\emph{Great Plains (held out)}} & {\scriptsize 76k} & {\scriptsize 80k} & {\scriptsize 74k} \\
\quad HRRR interp. & 2.06 & 35.20 & 1.66 & 2.64 & +1.66 & +0.05 & $-$1.59 & 0.03 & 0.04 & 0.00 \\
\quad HRRR stn.\ corr. & 1.33 & 33.65 & 1.60 & 1.79 & +0.12 & $-$0.08 & $-$0.19 & 0.03 & 0.02 & 0.00 \\
\quad U-Net & 0.99 & 34.14 & 1.50 & 1.69 & $-$0.05 & $-$0.09 & +0.19 & 0.32 & 0.16 & 0.18 \\
\quad ConvCNP$^{\dagger}$ & 0.99 & \textbf{32.71} & 1.49 & 1.65 & \underline{$-$0.01} & \underline{$-$0.03} & +0.15 & 0.41 & 0.21 & \textbf{0.22} \\
\quad CorrDiff$^{\dagger}$ & 1.02 & 34.85 & 1.58 & 1.72 & $-$0.36 & $-$0.19 & +0.23 & 0.35 & 0.20 & 0.16 \\
\rowcolor{ai2pinkrow!10}\quad\textbf{Vernier-Det} & \underline{0.98} & 33.25 & \underline{1.45} & \underline{1.64} & \textbf{+0.00} & \underline{$-$0.03} & \underline{+0.10} & \textbf{0.44} & 0.28 & 0.21 \\
\rowcolor{ai2pinkrow!10}\quad\quad\emph{+ Spectral} & 1.03 & \underline{33.12} & \underline{1.45} & \textbf{1.63} & $-$0.02 & +0.04 & +0.11 & 0.33 & \textbf{0.30} & 0.21 \\
\rowcolor{ai2pinkrow!10}\quad\textbf{Vernier-Prob}$^{\dagger}$ & \textbf{0.95} & 33.65 & \underline{1.45} & 1.65 & $-$0.11 & \textbf{+0.02} & +0.19 & \textbf{0.44} & 0.28 & \textbf{0.22} \\
\rowcolor{ai2pinkrow!10}\quad\quad\emph{+ Spectral} & 1.01 & 33.43 & \textbf{1.44} & \underline{1.64} & $-$0.13 & +0.08 & \textbf{+0.06} & 0.34 & \underline{0.29} & \textbf{0.22} \\
\addlinespace
\multicolumn{8}{l}{\emph{Cold Deserts (held out)}} & {\scriptsize 150k} & {\scriptsize 171k} & {\scriptsize 167k} \\
\quad HRRR interp. & 2.03 & 45.46 & 1.95 & 3.08 & +1.46 & +0.33 & $-$1.97 & 0.07 & 0.36 & 0.16 \\
\quad HRRR stn.\ corr. & 1.26 & 42.94 & 1.83 & 2.01 & $-$0.22 & $-$0.48 & +0.41 & 0.11 & 0.39 & 0.21 \\
\quad U-Net & 1.12 & 42.54 & 1.57 & 1.93 & $-$0.31 & \textbf{$-$0.01} & \textbf{+0.06} & 0.41 & 0.52 & 0.39 \\
\quad ConvCNP$^{\dagger}$ & \underline{1.09} & \textbf{39.79} & 1.63 & 1.91 & \textbf{$-$0.11} & $-$0.31 & +0.40 & \underline{0.48} & 0.51 & 0.43 \\
\quad CorrDiff$^{\dagger}$ & 1.16 & 43.44 & 1.60 & 2.00 & $-$0.66 & +0.07 & +0.37 & 0.45 & 0.51 & 0.35 \\
\rowcolor{ai2pinkrow!10}\quad\textbf{Vernier-Det} & \underline{1.09} & \underline{39.92} & \underline{1.53} & \textbf{1.88} & \underline{$-$0.15} & $-$0.14 & \underline{+0.28} & 0.47 & \underline{0.55} & 0.44 \\
\rowcolor{ai2pinkrow!10}\quad\quad\emph{+ Spectral} & 1.11 & 40.10 & \underline{1.53} & \underline{1.90} & $-$0.23 & $-$0.11 & +0.44 & 0.45 & 0.54 & 0.43 \\
\rowcolor{ai2pinkrow!10}\quad\textbf{Vernier-Prob}$^{\dagger}$ & \textbf{1.08} & 40.46 & \textbf{1.51} & \underline{1.90} & $-$0.20 & \underline{$-$0.03} & +0.37 & \textbf{0.49} & \textbf{0.56} & \textbf{0.45} \\
\rowcolor{ai2pinkrow!10}\quad\quad\emph{+ Spectral} & 1.11 & 40.36 & \underline{1.53} & 1.91 & $-$0.25 & $-$0.05 & +0.50 & 0.43 & \underline{0.55} & \textbf{0.45} \\
\midrule
\emph{Probabilistic} & \multicolumn{4}{c}{fCRPS $\downarrow$} & \multicolumn{3}{c}{SSR $\to 1$} & \multicolumn{3}{c}{Member $R_{\mathrm{pair}}$ $\uparrow$} \\
\cmidrule(lr){2-5}\cmidrule(lr){6-8}\cmidrule(lr){9-11}
\multicolumn{11}{l}{\emph{Spatial holdout, 2017--2025}} \\
\quad ConvCNP & 0.62 & 23.66 & 0.97 & 1.15 & \underline{0.85} & \textbf{0.97} & \textbf{1.00} & 0.21 & 0.15 & 0.08 \\
\quad CorrDiff & 0.66 & 24.65 & 1.01 & 1.20 & 0.55 & 0.67 & 0.66 & 0.30 & 0.27 & 0.15 \\
\rowcolor{ai2pinkrow!10}\quad\textbf{Vernier-Prob} & \textbf{0.58} & \textbf{23.34} & \textbf{0.90} & \underline{1.08} & \textbf{0.90} & \textbf{0.97} & \underline{0.97} & \textbf{0.38} & \underline{0.46} & \underline{0.35} \\
\rowcolor{ai2pinkrow!10}\quad\quad\emph{+ Spectral} & \underline{0.60} & \underline{23.43} & \textbf{0.90} & \textbf{1.07} & 0.84 & 0.96 & \underline{0.97} & \underline{0.37} & \textbf{0.48} & \textbf{0.37} \\
\multicolumn{11}{l}{\emph{Great Plains (held out)}} \\
\quad ConvCNP & \underline{0.71} & \textbf{23.58} & 1.07 & 1.19 & \underline{0.81} & \textbf{0.94} & \textbf{0.96} & 0.18 & 0.04 & 0.04 \\
\quad CorrDiff & 0.78 & 24.92 & 1.20 & 1.30 & 0.50 & 0.53 & 0.54 & 0.19 & 0.07 & 0.07 \\
\rowcolor{ai2pinkrow!10}\quad\textbf{Vernier-Prob} & \textbf{0.67} & 23.91 & \textbf{1.04} & \textbf{1.18} & \textbf{0.86} & \underline{0.85} & \underline{0.86} & \textbf{0.36} & \underline{0.23} & \underline{0.18} \\
\rowcolor{ai2pinkrow!10}\quad\quad\emph{+ Spectral} & 0.72 & \underline{23.66} & \textbf{1.04} & \textbf{1.18} & 0.78 & \underline{0.85} & \underline{0.86} & \underline{0.29} & \textbf{0.25} & \textbf{0.20} \\
\multicolumn{11}{l}{\emph{Cold Deserts (held out)}} \\
\quad ConvCNP & \underline{0.79} & 28.78 & 1.16 & 1.37 & 0.79 & 1.02 & 1.03 & 0.24 & 0.16 & 0.09 \\
\quad CorrDiff & 0.91 & 30.10 & 1.21 & 1.50 & 0.46 & 0.57 & 0.56 & 0.30 & 0.32 & 0.16 \\
\rowcolor{ai2pinkrow!10}\quad\textbf{Vernier-Prob} & \textbf{0.76} & \textbf{28.18} & \textbf{1.08} & \textbf{1.35} & \textbf{0.87} & \textbf{0.99} & \textbf{1.00} & \textbf{0.38} & \underline{0.48} & \underline{0.37} \\
\rowcolor{ai2pinkrow!10}\quad\quad\emph{+ Spectral} & \underline{0.79} & \underline{28.30} & \textbf{1.08} & \underline{1.36} & \underline{0.81} & \textbf{0.99} & \underline{1.01} & \underline{0.34} & \textbf{0.50} & \textbf{0.39} \\
\bottomrule
\end{tabular}}%
\ifdim\wd\supptabbox>\columnwidth\resizebox{\columnwidth}{!}{\usebox{\supptabbox}}\else\usebox{\supptabbox}\fi
\caption{Additional holdouts, with the metrics of \cref{tab:results} (without de-meaned $R_{\mathrm{pair}}$). \emph{Spatial holdout}: the held-out test stations of \cref{tab:results} over 2017--2025 (50k samples, 184k station-hours). \emph{Ecoregion holdouts}: models retrained without the Great Plains or the Cold Deserts (and a 20\,km buffer) and scored on the region's test part (50k samples each; 145k and 231k station-hours), 2017--2025. \emph{+ Spectral}: trained with the spectral loss. $^{\dagger}$Ensemble mean. The row under the column names gives units; the counts beside each holdout are the station pairs 0.1--3\,km apart. Bold: best; underline: second best, none when the best is tied, within each holdout.}
\label{tab:extreme}
\end{table}

\paragraph{Ecoregion holdouts.}
Withholding a whole region narrows every learned model's lead over HRRR for wind speed and $T$, most for $T$ in the Great Plains (Vernier's MAE is 12--13\% below HRRR's, against about 26\% in the spatial holdout), but not for $T_d$, where much of the gain comes from removing HRRR's dry bias.
Vernier stays ahead in both regions, with the lowest wind-speed, $T$ and $T_d$ MAE and the highest $R_{\mathrm{pair}}$ (shared with ConvCNP for $T_d$ in the Great Plains), while ConvCNP has the lowest wind-direction error.
Station-pair structure transfers well to the Cold Deserts, where Vernier's $R_{\mathrm{pair}}$ matches the spatial holdout, but less to the flat Great Plains, where HRRR's $R_{\mathrm{pair}}$ is near zero and Vernier's falls to 0.28--0.30 for $T$ and 0.21--0.22 for $T_d$, still at or above the baselines.
In both regions Vernier-Prob has the lowest fCRPS for wind speed, $T$ and $T_d$, and CorrDiff remains underdispersed. Vernier-Prob's spread--skill ratio is closest to one in the Cold Deserts, but in the Great Plains ConvCNP's is closer for $T$ and $T_d$ (0.94--0.96 against 0.85--0.86), and ConvCNP has the lowest direction fCRPS there. Member $R_{\mathrm{pair}}$, however, is far lower for ConvCNP and CorrDiff than for Vernier-Prob for every variable in both regions (for $T$ and $T_d$ in the Great Plains, 0.04--0.07 against 0.18--0.23): their individual members are much less coherent than Vernier-Prob's.

The spectral loss costs more wind-speed skill than in \cref{tab:results}: in the Great Plains it lowers wind-speed $R_{\mathrm{pair}}$ from 0.44 to 0.33--0.34 and raises MAE by 0.05--0.06\,m\,s$^{-1}$, while $T$ and $T_d$ change by at most 0.02. This is especially relevant for the Great Plains, because it is not a terrain-dominated region, and our prior has very little information here to downscale winds. 
This highlights one of the shortcomings of the spectral regularization method, and a \textit{distinct failure mode}.

\vspace{-3mm}
\paragraph{Overall.}
The ranking of \cref{tab:results} holds on every holdout: Vernier has the lowest wind-speed, $T$ and $T_d$ errors and the most station-pair structure, then the learned baselines, then station-corrected HRRR and HRRR.
Vernier's MAE lead over the learned baselines is small, but its lead in structure and fCRPS is larger and persists out of region.
Holding out a region is harder than holding out stations: the gains over HRRR shrink for wind speed and $T$, and structure transfers better to complex terrain than to flat terrain.
The Vernier-Prob ensemble mean has the lowest wind-speed MAE on every holdout and is close to Vernier-Det otherwise.
The spectral loss leaves $T$ and $T_d$ nearly unchanged and costs some wind-speed skill, more in held-out regions where the prior can provide less information.

\subsection{Probabilistic analysis}
\label{supp:prob_analysis}

\begin{table}[t]
  \centering
  \scriptsize
  \definecolor{ai2pinkrow}{HTML}{F0529C}%
  \definecolor{abBetter}{HTML}{4C9BE8}%
  \definecolor{abWorse}{HTML}{F0529C}
  \setlength{\tabcolsep}{2.6pt}
  \supptabfit{\columnwidth}{%
  \begin{tabular}{@{}l ccccccc@{}}
    \toprule
    & \multicolumn{4}{c}{fCRPS $\downarrow$} & \multicolumn{3}{c}{SSR $\to 1$} \\
    \cmidrule(lr){2-5}\cmidrule(lr){6-8}
    Ablation & ws & dir & $T$ & $T_d$ & ws & $T$ & $T_d$ \\
    & {\tiny m\,s$^{-1}$} & {\tiny $^\circ$} & {\tiny $^\circ$C} & {\tiny $^\circ$C} & & & \\
    \midrule
    \rowcolor{ai2pinkrow!10}\textbf{Forecast*} & 0.600 & 22.19 & 0.881 & 1.099 & 0.787 & 0.947 & 0.970 \\
    Comp.\ only & \cellcolor{abWorse!30}+0.005 & \cellcolor{abWorse!40}+0.15 & \cellcolor{abBetter!12}$-$0.001 & \cellcolor{abWorse!26}+0.004 & \cellcolor{abBetter!32}+0.016 & 0.000 & \cellcolor{abWorse!43}$-$0.013 \\
    + twCRPS & \cellcolor{abBetter!23}$-$0.003 & \cellcolor{abBetter!30}$-$0.08 & 0.000 & \cellcolor{abWorse!26}+0.004 & $-$0.003 & +0.001 & \cellcolor{abWorse!36}$-$0.009 \\
    + Surface SpAdaLN & \cellcolor{abBetter!14}$-$0.001 & \cellcolor{abBetter!34}$-$0.11 & \cellcolor{abWorse!25}+0.005 & \cellcolor{abWorse!26}+0.004 & \cellcolor{abWorse!32}$-$0.016 & $-$0.002 & \cellcolor{abWorse!38}$-$0.010 \\
    \midrule
    & \multicolumn{3}{c}{Mean $R_{\mathrm{pair}}$ $\uparrow$} & & & & \\
    \cmidrule(lr){2-4}
    Ablation & ws & $T$ & $T_d$ & & & & \\
    \midrule
    \rowcolor{ai2pinkrow!10}\textbf{Forecast*} & 0.501 & 0.566 & 0.328 & & & & \\
    Comp.\ only & +0.002 & +0.001 & $-$0.006 & & & & \\
    + twCRPS & \cellcolor{abBetter!31}+0.012 & +0.002 & \cellcolor{abWorse!38}$-$0.015 & & & & \\
    + Surface SpAdaLN & \cellcolor{abWorse!26}$-$0.008 & $-$0.002 & +0.005 & & & & \\
    \bottomrule
  \end{tabular}
  }%
  \caption{\textbf{Further probabilistic ablations} on the validation samples of \cref{tab:ablation} (64 members). The first row of each block is the reference, Forecast*; other rows show the change from it, shaded \colorbox{abBetter!40}{blue} when better and \colorbox{abWorse!40}{pink} when worse if the change exceeds the seed standard deviation, on the scale of \cref{tab:ablation}. Comp.\ only: fair CRPS of $u$ and $v$ separately instead of the energy score and the speed term. twCRPS: an added threshold-weighted CRPS of wind speed above 1.6\,m\,s$^{-1}$ (weight 0.5). + Surface SpAdaLN: the surface descriptors (Earth embedding and terrain) added to the forecast in the SpAdaLN field.}
  \label{tab:supp_prob_ablation}
\end{table}

\paragraph{Further ablations.}
\Cref{tab:supp_prob_ablation} adds three changes of Forecast* to those of \cref{tab:ablation}, on the same validation samples.
Scoring $u$ and $v$ with separate fair CRPS terms instead of the energy score and the speed term worsens wind speed and direction (\cref{supp:probabilistic}).
A threshold-weighted CRPS on wind speeds above 1.6\,m\,s$^{-1}$ slightly improves the wind-speed and direction fCRPS and the wind-speed $R_{\mathrm{pair}}$, but worsens dew point and lowers its spread--skill ratio, so it is not used.
Surface SpAdaLN modulates the noise with the surface descriptors, the Earth embedding and the terrain, together with the HRRR forecast, instead of the forecast alone (\cref{supp:gen-ens}); it slightly improves the wind-direction fCRPS but worsens temperature and dew point and lowers the wind-speed spread--skill ratio.
Across these variants, a lower fCRPS tends to come with a lower spread--skill ratio, so the two trade off.
Threshold-weighted terms remain underexplored: we tried one threshold and one weight, on wind speed only.
Tuning them and extending them to the other variables is future work, especially to improve the tails.

\begin{figure*}[p]
  \centering
  \begin{minipage}{0.64\textwidth}
    \small\textbf{a}\enspace Vernier-Prob\par\vspace{1pt}
    \includegraphics[width=\linewidth]{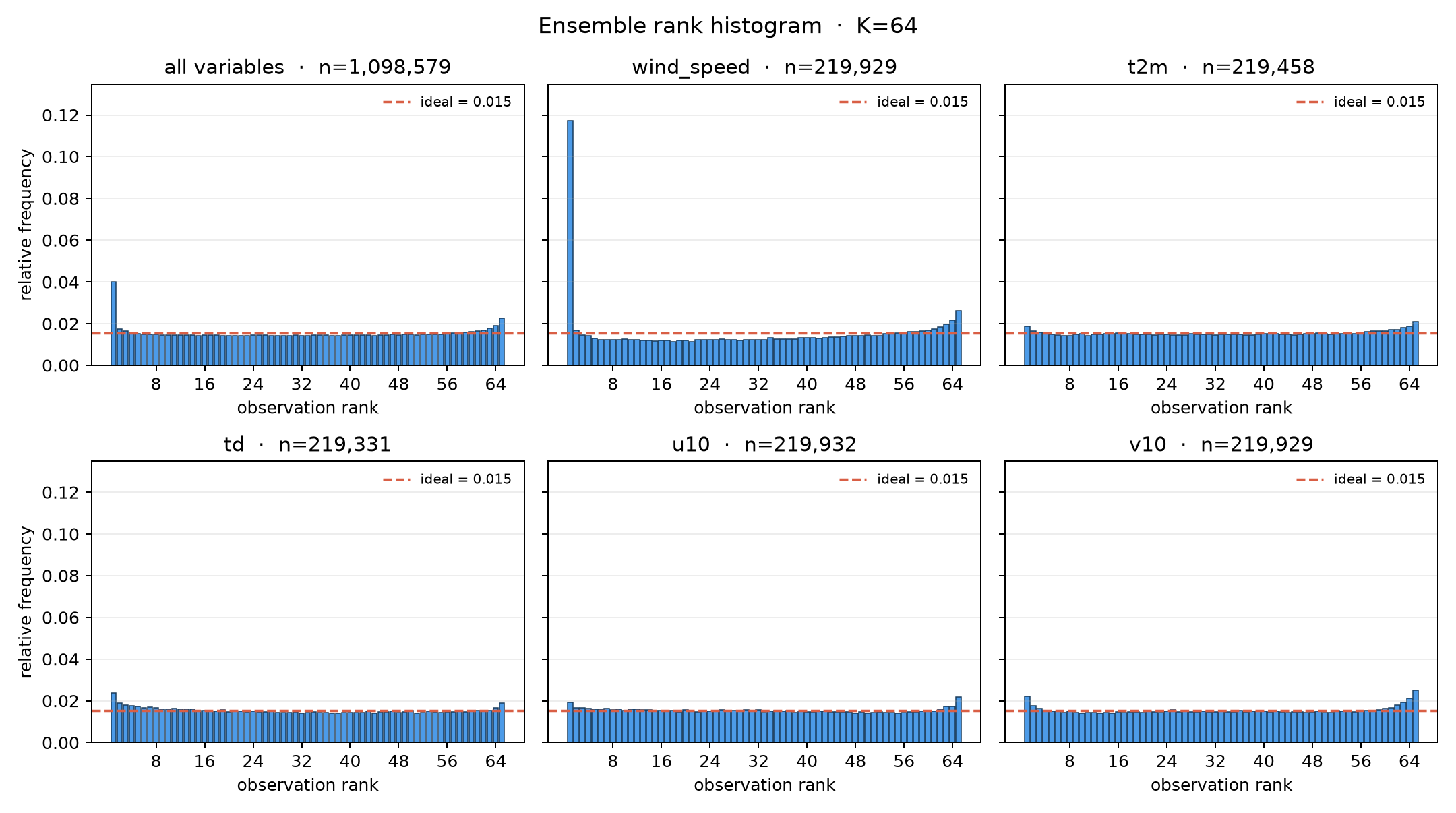}\par\medskip
    \textbf{b}\enspace Vernier-Prob + Spectral\par\vspace{1pt}
    \includegraphics[width=\linewidth]{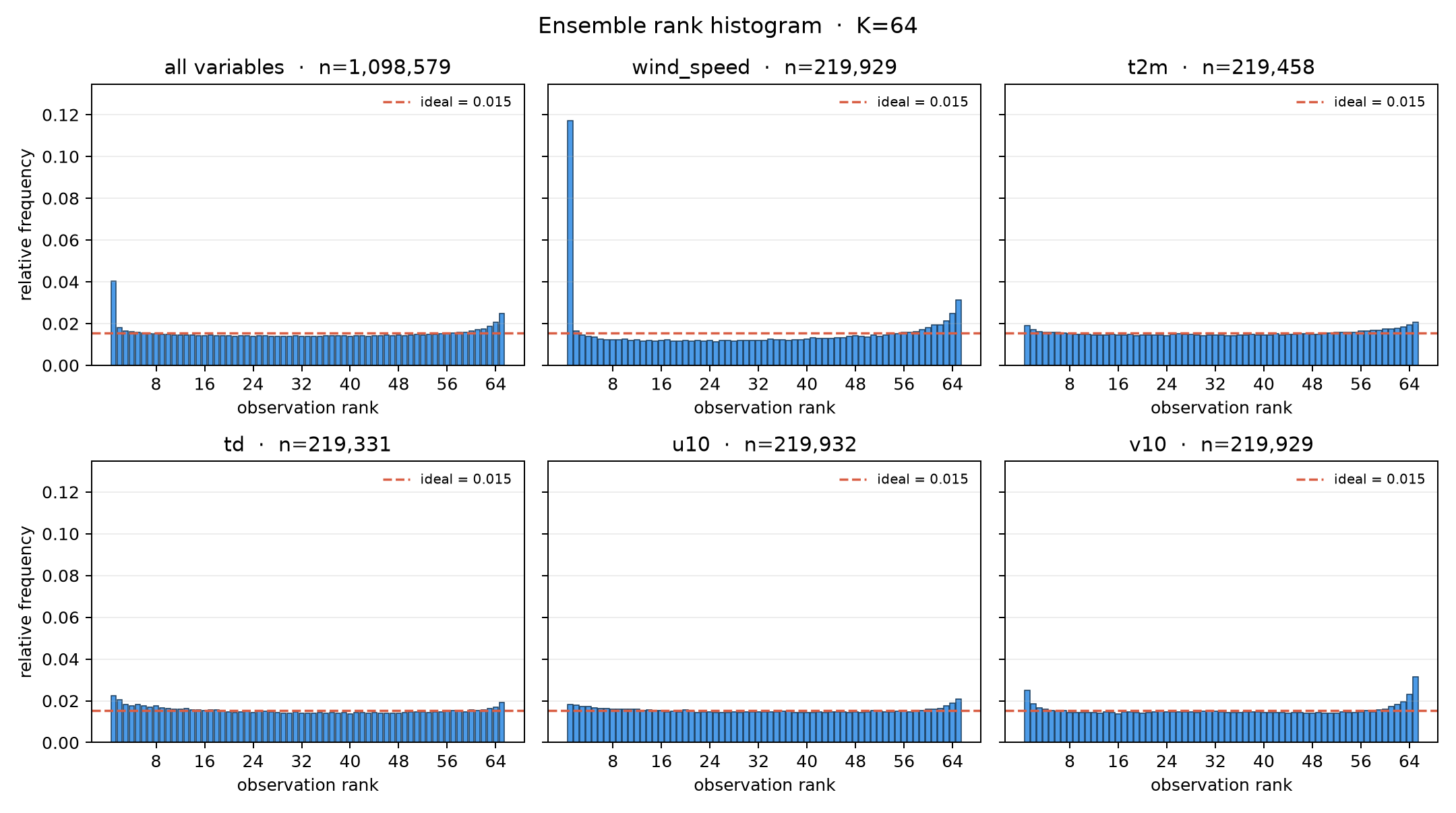}\par\medskip
    \textbf{c}\enspace Operational Vernier-Prob (+ Spectral)\par\vspace{1pt}
    \includegraphics[width=\linewidth]{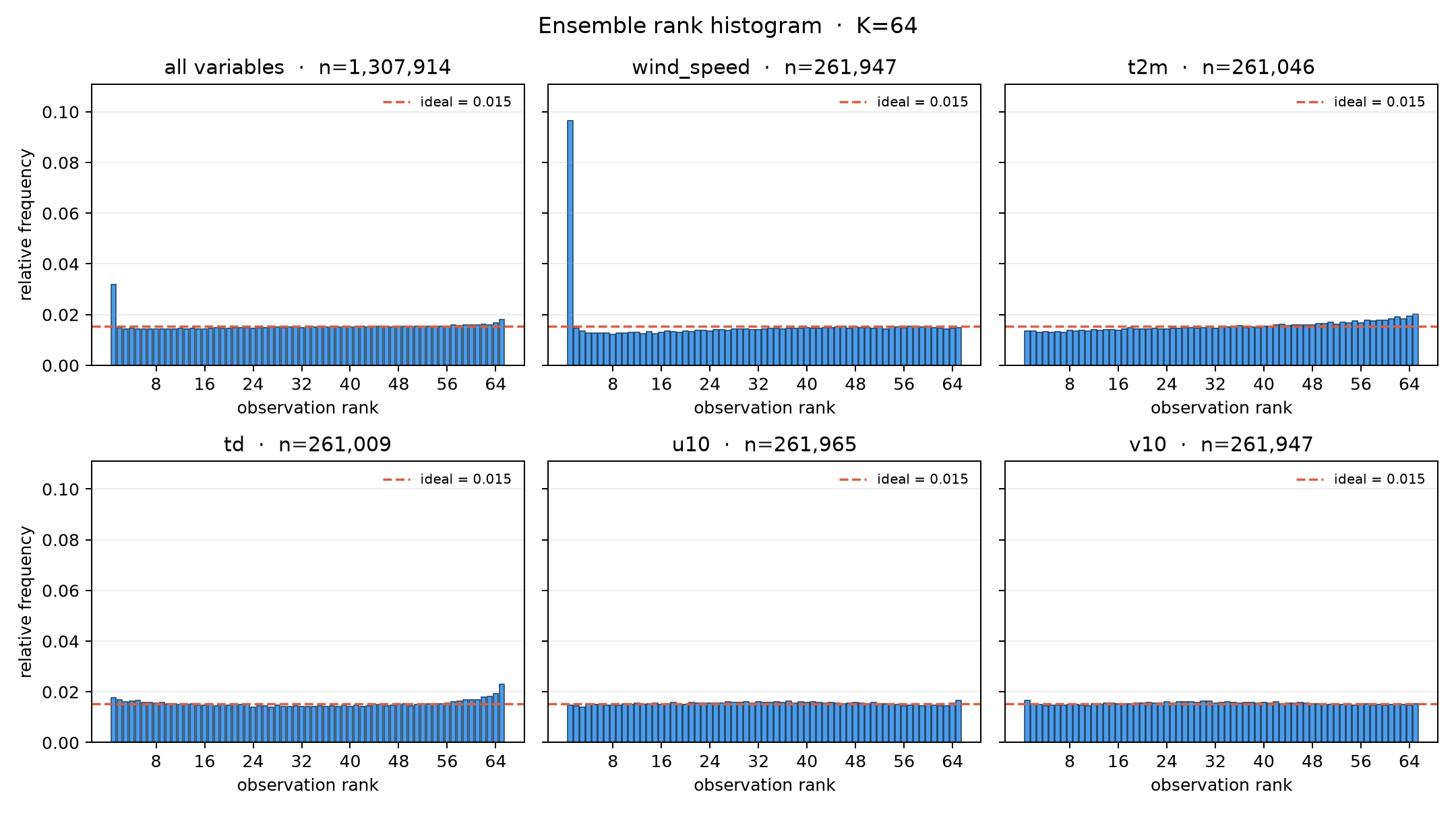}
  \end{minipage}
  \caption{\textbf{Rank histograms of the 64-member Vernier-Prob ensembles.} Each panel shows how often the observation falls at each rank among the sorted members (1: below every member; 65: above every member). Dashed: the uniform frequency $1/65$ of a calibrated ensemble; $n$: station-hours. \emph{all variables} pools the other five panels; t2m, td, u10 and v10 are $T$, $T_d$, $u$ and $v$. \textbf{a}, \textbf{b}: the spatio-temporal holdout of \cref{tab:results}, on the same samples. \textbf{c}: the 2026 HRRR-domain test of \cref{tab:operational}.}
  \label{fig:supp_rank_hist}
\end{figure*}

\vspace{-3mm}
\paragraph{Calibration.}
A rank histogram counts where each observation falls among the sorted members; for a calibrated 64-member ensemble, each of the 65 ranks is equally likely. \Cref{fig:supp_rank_hist} shows the final 32\,+\,32 ensembles: Vernier-Prob without (a) and with (b) the spectral loss on the spatio-temporal holdout of \cref{tab:results}, and the operational model (c) on the 2026 HRRR-domain test of \cref{tab:operational}. For $u$, $v$, $T$ and $T_d$ the histograms are nearly flat, with slightly raised outer ranks: the mild underdispersion that an SSR below one indicates. Wind speed has one large departure: in 11.6\% of station-hours the observation lies below all 64 members, against 1.5\% for a flat histogram. These are calm hours. Of the observed speeds, 30\% are below 0.5\,m\,s$^{-1}$, but a member's speed, the magnitude of its $u$ and $v$, is rarely that low. The same hours fall inside the spread of $u$ and $v$, whose lowest ranks are raised far less (1.2--1.6 times uniform), and at observed speeds of at least 1.6\,m\,s$^{-1}$ the observation almost never falls below every member. At the other end, the observation exceeds every member in 2.6\% of hours, so the ensemble slightly underrepresents the strongest winds.

The spectral loss makes every histogram slightly less flat. The L1 distance from uniform rises from 0.27 to 0.30 for wind speed, 0.05 to 0.07 for $T$, 0.06 to 0.08 for $T_d$ and 0.07 to 0.09 for $v$, mostly through the upper tail of wind speed and $v$: the observation exceeds every member in 3.1\% and 3.2\% of hours, against 2.6\% and 2.5\% without the loss, in line with its lower wind-speed SSR (0.85 against 0.89). The operational model, whose test stations are mostly seen in training, is the flattest for $u$ and $v$ and has no upper wind-speed tail. Its $T$ histogram slopes upward: observations fall in the upper ranks more often than in the lower, the signature of an ensemble that is too cold ($-0.13\,^\circ$C, \cref{tab:operational}) rather than too narrow.

\subsection{Is the model just station de-biasing?}
\label{supp:mos}
\begin{table}[t]
  \centering
  \footnotesize
  \definecolor{ai2pinkrow}{HTML}{F0529C}%
  \setlength{\tabcolsep}{5pt}
  \begin{tabular}{l ccc ccc ccc}
  \toprule
  \emph{Deterministic} & \multicolumn{3}{c}{MAE $\downarrow$} & \multicolumn{3}{c}{$R_{\mathrm{pair}}$ $\uparrow$}
    & \multicolumn{3}{c}{De-meaned $R_{\mathrm{pair}}$ $\uparrow$} \\
  \cmidrule(lr){2-4}\cmidrule(lr){5-7}\cmidrule(lr){8-10}
  Model & ws & $T$ & $T_d$ & ws & $T$ & $T_d$ & ws & $T$ & $T_d$ \\
   & {\scriptsize m\,s$^{-1}$} & {\scriptsize $^\circ$C} & {\scriptsize $^\circ$C} & & & & & & \\
  \midrule
  HRRR & 1.91 & 1.67 & 2.57 & 0.08 & 0.25 & 0.20 & 0.08 & 0.20 & 0.13 \\
  HRRR + MOS, one line & 1.05 & 1.71 & 2.26 & 0.06 & 0.16 & 0.12 & 0.05 & 0.12 & 0.08 \\
  HRRR + MOS, per station & 0.72 & 1.55 & 2.09 & 0.68 & 0.42 & 0.46 & 0.29 & 0.17 & 0.14 \\
  HRRR + MOS, station $\times$ ToD & \underline{0.68} & 1.43 & 2.03 & \textbf{0.71} & 0.56 & \textbf{0.51} & 0.31 & 0.22 & 0.15 \\
  \rowcolor{ai2pinkrow!10}\textbf{Vernier-Det} & 0.73 & \underline{1.19} & \textbf{1.52} & 0.62 & \underline{0.67} & 0.44 & \underline{0.38} & \underline{0.61} & \underline{0.33} \\
  \rowcolor{ai2pinkrow!10}\textbf{Vernier-Prob}$^{\dagger}$ & \textbf{0.67} & \textbf{1.18} & \textbf{1.52} & \underline{0.69} & \textbf{0.69} & \underline{0.48} & \textbf{0.45} & \textbf{0.63} & \textbf{0.35} \\
  \midrule
  \emph{Probabilistic} & \multicolumn{3}{c}{fCRPS / CRPS $\downarrow$} & \multicolumn{3}{c}{Member $R_{\mathrm{pair}}$ $\uparrow$}
    & \multicolumn{3}{c}{\shortstack{Member\\de-meaned $R_{\mathrm{pair}}$ $\uparrow$}} \\
  \cmidrule(lr){2-4}\cmidrule(lr){5-7}\cmidrule(lr){8-10}
  HRRR + EMOS, one line & 0.83 & 1.23 & 1.66 & 0.01 & 0.05 & 0.03 & 0.01 & 0.03 & 0.02 \\
  HRRR + EMOS, per station & 0.54 & 1.11 & 1.51 & 0.43 & 0.14 & 0.14 & \underline{0.09} & 0.04 & \underline{0.03} \\
  HRRR + EMOS, station $\times$ ToD & \underline{0.51} & \underline{1.03} & \underline{1.47} & \underline{0.46} & \underline{0.24} & \underline{0.16} & \underline{0.09} & \underline{0.06} & \underline{0.03} \\
  \rowcolor{ai2pinkrow!10}\textbf{Vernier-Prob} & \textbf{0.46} & \textbf{0.83} & \textbf{1.08} & \textbf{0.60} & \textbf{0.63} & \textbf{0.41} & \textbf{0.31} & \textbf{0.53} & \textbf{0.26} \\
  \bottomrule
  \end{tabular}
  \caption{\textbf{Station-calibrated HRRR on the operational test} of \cref{tab:operational} (HRRR domain, January--September 2026: 50k samples, 264,960 station-hours at 23,667 stations). MOS (\cref{eq:mos}) and its probabilistic form EMOS (\cref{eq:emos}) are fit on each station's own 2017--2025 record, with one line for all stations, one per station, or one per station and six-hour block of local solar time (station $\times$ ToD). They use no 2026 data; Vernier saw the same stations' history in training (as target, not inputs). MOS and EMOS rows are scored on the 98.5--98.9\% of station-hours where every forecast exists; stations without enough history for a per-station fit (about 2\%) use the one-line fit. De-meaned $R_{\mathrm{pair}}$: $R_{\mathrm{pair}}$ after removing each station pair's mean difference (\cref{supp:structure}), here separately in each six-hour block of local solar time, over the whole test period, which cancels any per-station calibration. Member $R_{\mathrm{pair}}$ of EMOS uses 64 draws sampled independently at each station-hour. $^{\dagger}$Ensemble mean (wind speed: mean of the member speeds). Bold: best; underline: second best. Paired differences with 95\% intervals, for (ws, $T$, $T_d$), against the station $\times$ ToD line: Vernier-Det MAE $+7.0\%$ $[+5.3,+8.9]$, $-17.0\%$ $[-18.6,-15.4]$ and $-24.9\%$ $[-27.0,-23.0]$ (wind speed against the per-station line: $+1.1\%$ $[-0.7,+3.0]$); Vernier-Prob ensemble-mean wind-speed MAE $-1.7\%$ $[-3.3,+0.1]$; Vernier-Prob fCRPS against the EMOS CRPS $-9.0\%$ $[-10.2,-7.6]$, $-19.3\%$ $[-21.0,-17.7]$ and $-26.8\%$ $[-28.8,-25.0]$; Vernier-Det de-meaned $R_{\mathrm{pair}}$ $+0.07$ $[0.03,0.10]$, $+0.38$ $[0.33,0.43]$ and $+0.18$ $[0.15,0.21]$; Vernier-Prob member de-meaned $R_{\mathrm{pair}}$ against the EMOS draws $+0.22$ $[0.18,0.25]$, $+0.47$ $[0.43,0.51]$ and $+0.23$ $[0.20,0.26]$.}\label{tab:mos}
\end{table}

A downscaler trained on station observations could score well simply by learning the systematic errors of its input at each site: HRRR's 10\,m wind is too fast, most of all at slower and forested sites and at night~\citesupp{fovell2020hrrr,fovell2022hrrrwind} (\cref{fig:strat_detail}c), its 2\,m temperature errors depend on station elevation~\citesupp{fovell2020hrrr}, and at our stations its dew point is too dry (\cref{tab:operational}).
Operational forecasting removes such errors with statistical post-processing at every station that has a record~\cite{glahn1972mos,gneiting2005emos}, at a fraction of the cost of a neural downscaler.
If Vernier's gains were mostly this calibration, a much simpler method would suffice wherever station history exists.
We therefore compare against post-processing where it is strongest: the operational test of \cref{tab:operational}, whose 2026 stations all have 2017--2025 observations.
Every station can thus be calibrated on its own history, using only data that Vernier also saw in training.
Unlike the HRRR station-correction baseline (\cref{supp:unet}), a U-Net that must generalize to unseen stations, MOS and EMOS fit each station separately (configuration and scoring in \cref{supp:mos_config}).

\paragraph{De-meaned $R_{\mathrm{pair}}$.}
We demean the station-pair differences as in \cref{supp:structure} (\emph{Separating persistent and time-varying structure}), with one change: each pair's mean is removed separately in each six-hour block of local solar time, the blocks of \cref{eq:mos}.
Any per-station offset then cancels, even one that differs by block, so a forecast scores only by following how the contrast between stations changes with the weather.
We keep pair-blocks with at least five station-hours, which hold 90\% of the pair-hours below 3\,km (about 6,500 pairs at 4,500 stations, a median of 14 hours per pair and block).
Intervals are 95\% intervals of paired differences from the two-way cluster bootstrap of \cref{supp:structure}, here with 1,000 replicates (Poisson(1) weights per 50\,km cell and per day).

\paragraph{Calibration and error.}
Calibration removes most of HRRR's wind-speed error, because a station's exposure (sheltering, siting, local roughness) acts as a fixed multiplier on its wind that years of its own record pin down: per-station MOS lowers the wind-speed MAE from 1.91 to 0.68--0.72\,m\,s$^{-1}$ (\cref{tab:operational}).
Vernier-Det's MAE is close to the per-station line (0.73 vs.\ 0.72\,m\,s$^{-1}$) but trails the per-station $\times$ ToD line by 7\%, which the Vernier-Prob ensemble mean matches (0.67 vs.\ 0.68\,m\,s$^{-1}$; $-1.7\%$, 95\% interval $[-3.3,+0.1]$).
Temperature and dew-point errors depend more on the weather than on the site, and there Vernier-Det lowers the MAE of the best MOS by a further 17\% and 25\%.
Probabilistically, Vernier-Prob's fCRPS is 9--27\% below the CRPS of the best EMOS for all three variables, with 95\% intervals well clear of zero (\cref{tab:mos}).

\paragraph{Why MOS reaches a high $R_{\mathrm{pair}}$ but a low de-meaned $R_{\mathrm{pair}}$.}
Nearby stations usually share an HRRR cell, so HRRR alone barely tells them apart.
MOS gives each station its own line, fit to years of its record, and so builds each site's long-run contrast with its neighbors into the forecast: a sheltered station is forecast lower than an exposed one every hour.
This fixed contrast explains much of the observed difference between nearby stations, so MOS reaches a high $R_{\mathrm{pair}}$, matching or beating Vernier's for wind speed and dew point (\cref{tab:mos}).
Once each pair's mean contrast is removed, MOS falls well behind Vernier for every variable, most of all for temperature.
This follows from its form: MOS can only rescale HRRR's value at each station, while the contrast between neighbors changes with the weather, with the wind direction relative to ridges and buildings, a calm valley under a windy ridge at night, cold air pooling on a valley floor, or moist air crossing a coast.
Vernier sees the HRRR fields around the site together with 100\,m terrain and surface descriptors, and follows these changes.
MOS fit per time of day does learn the regular day--night swing of the wind contrast, so for wind speed it keeps up with Vernier-Det when only each pair's overall mean is removed, as in \cref{supp:structure} (de-meaned $R_{\mathrm{pair}}$ 0.42 for both).
In short, station calibration copies the fixed differences between neighboring stations, but only Vernier captures how those differences change with the weather.

\begin{table*}[t]
\centering
\footnotesize
\aitwotable{tab:station_networks}%
\setlength{\tabcolsep}{3pt}
\supptabfit{\textwidth}{%
\begin{tabular}{l cccccc cccc}
\toprule
 & \multicolumn{6}{c}{Wind-speed MAE (bias), m\,s$^{-1}$} & Dir & $R_{\mathrm{pair}}$ & $T$ & $T_d$ \\
\cmidrule(lr){2-7}
Station wind used in training & Airports & RAWS & CWOP & MesoWest & DOT, other & All & $^\circ$ & ws & $^\circ$C & $^\circ$C \\
 & {\scriptsize 47k} & {\scriptsize 42k} & {\scriptsize 285k} & {\scriptsize 499k} & {\scriptsize 10k} & {\scriptsize 882k} & & {\scriptsize 684k} & & \\
\midrule
HRRR interp. & 1.39 & 1.78 & 2.49 & 1.77 & 1.99 & 1.99 & 36.91 & 0.06 & 1.72 & 2.53 \\
\midrule
Every network (Vernier-Det) & 1.14 ($-$0.05) & 0.95 ($-$0.14) & 0.72 (+0.20) & 0.85 (+0.06) & 0.94 ($-$0.16) & 0.83 (+0.09) & 33.41 & 0.49 & 1.28 & 1.56 \\
No CWOP & 1.13 (+0.01) & 0.97 ($-$0.03) & 1.00 (+0.70) & 0.85 (+0.12) & 0.98 (+0.05) & 0.92 (+0.29) & 33.62 & 0.48 & 1.28 & 1.56 \\
Airports + RAWS & 1.13 (+0.13) & 1.02 (+0.04) & 1.30 (+1.10) & 0.96 (+0.31) & 1.15 (+0.45) & 1.08 (+0.55) & 34.90 & 0.41 & 1.30 & 1.56 \\
Airports only & 1.12 (+0.16) & 1.20 (+0.40) & 1.65 (+1.50) & 1.13 (+0.59) & 1.32 (+0.75) & 1.30 (+0.86) & 35.95 & 0.32 & 1.30 & 1.58 \\
\bottomrule
\end{tabular}
}%
\caption{\textbf{Training on different station networks.} Wind-speed MAE and bias (in parentheses) per station network for Vernier-Det trained with the station wind of different networks, scored at every station on a 200k-sample draw of the test set of \cref{tab:results} (held-out test stations, January--September 2026). Networks as in \cref{tab:strat_network}, with DOT and other combined. \emph{All}: every station. Dir: wind-direction MAE at observed speeds $\geq 1.6$\,m\,s$^{-1}$. $R_{\mathrm{pair}}$: wind speed, station pairs 0.1--3\,km apart. $T$, $T_d$: MAE at every station. The row below the headings gives each network's wind station-hours and the number of station pairs.}
\label{tab:station_networks}
\end{table*}

\paragraph{Structure in ensemble members.}
A probabilistic forecast must also give realistic scenarios, not just good values at each station.
EMOS draws are sampled independently at each station, so they keep the fixed contrasts of the per-station fit but almost none of how the contrast changes with the weather, and the members of ConvCNP and CorrDiff keep much less structure than Vernier's (\cref{tab:extreme}).
Each Vernier-Prob member is a spatially coherent 100\,m field that keeps this structure, also after each pair's mean contrast is removed (\cref{tab:mos}).
In short, Vernier's members keep far more of the structure between stations than the EMOS draws or the other downscalers' members.

\paragraph{Vernier is not just station de-biasing.} Against HRRR calibrated on each station's own nine-year history, Vernier-Prob's CRPS is lower for every variable (by 9--27\%), its ensemble mean is better for temperature and dew point and ties the best calibration for wind speed, and Vernier-Det reduces temperature and dew-point errors well beyond what calibration achieves.
Only for wind-speed MAE does per-station calibration match or beat Vernier-Det.
Calibration can reproduce the persistent contrasts between neighboring stations but not how those contrasts change with the weather: once each pair's mean contrast per block is removed, Vernier leads for every variable.
Vernier takes no station observations as input, and yet it provides realistic structure between stations, not just at them: its members are coherent 100\,m fields whose station-pair differences track the observations far better than those of per-station post-processing or the other probabilistic downscalers.

\subsection{Training on different station networks}
\label{supp:station_networks}
The networks of \cref{tab:strat_network} site their anemometers differently: airports follow the 10\,m open-exposure standard, while CWOP home stations and many mesonet stations are sheltered and read lower winds. To test whether the wind should be trained on the standard-exposure networks only, we retrain Vernier-Det with the station wind loss restricted to subsets of the networks; temperature and dew point always use every station. \emph{No CWOP} drops the CWOP stations, and \emph{Airports + RAWS} and \emph{Airports only} keep those networks. \Cref{tab:station_networks} scores every model at every station on a new 200k-sample draw of the test set of \cref{tab:results}, held out in space and time (test stations unseen in training, forecasts initialized January--September 2026; 2,513 stations; 882k wind station-hours, 47k of them at airports and 42k at RAWS stations).

Restricting the wind to the standard-exposure networks hardly helps where it is meant to: at airports, the wind-speed MAE is 1.12--1.14\,m\,s$^{-1}$ for every model. At a network left out of training the wind becomes too fast, at CWOP stations by +0.70 to +1.50\,m\,s$^{-1}$. With fewer stations, the station-pair structure weakens ($R_{\mathrm{pair}}$ 0.49 with every network, 0.32 with airports only) and the direction error grows (33.4$^\circ$ to 36.0$^\circ$). Dropping only CWOP keeps the structure ($R_{\mathrm{pair}}$ 0.48) and changes the wind-speed MAE at the other networks by at most 0.04\,m\,s$^{-1}$. Temperature and dew point change by at most 0.02\,$^\circ$C at any network. Vernier-Det is therefore trained on every network.

\section{Searchlight}
\label{supp:searchlight}

\begin{figure*}[t]
  \centering
  \includegraphics[width=\textwidth]{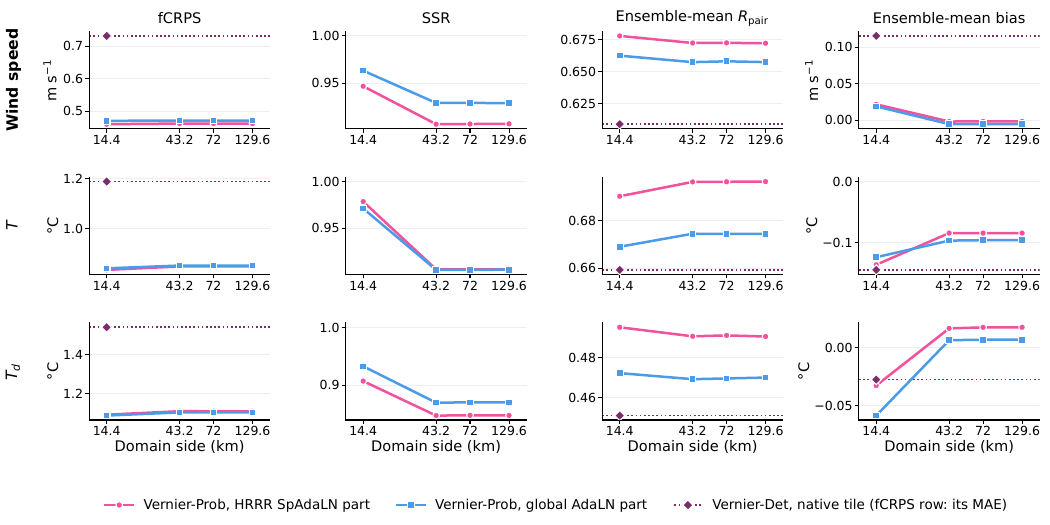}
  \caption{\textbf{Searchlight domain-size scaling (CONUS 2026 test).} The central $14.4\,\mathrm{km}$ tile scored after native inference or as the center of a $43$, $72$ or $130\,\mathrm{km}$ Searchlight domain, for the two parts of the operational Vernier-Prob ensemble: spatial AdaLN driven by HRRR (pink) and global AdaLN (blue). Same $5{,}000$ draws, $K=16$, one noise realization. Rows: wind speed, $T$ and $T_d$. Columns: fCRPS, SSR, and the $R_{\mathrm{pair}}$ and bias of the ensemble mean (wind speed: mean of the member speeds). Dark pink: the operational Vernier-Det on its native tile, scored on the same draws; in the fCRPS column its MAE, which is the CRPS of a deterministic forecast. Gray dashed lines: SSR of one and zero bias.}
  \label{fig:supp_searchlight_scaling}
\end{figure*}

Vernier is trained on $14.4\,\mathrm{km}$ tiles ($144\times144$ pixels at $100\,\mathrm{m}$).
Searchlight is how we run it over a larger domain, and was created to solve the noise coordination problem: an ensemble member must be one realization of the fine-scale weather over the whole domain. Climate in a Bottle (cBottle~\cite{brenowitz2025cbottle}) uses MultiDiffusion~\citesupp{bartal2023multidiffusion} iteratively over sampling to average neighbors' fields to achieve this coordination. Single-step ensemble methods cannot do this, so we choose to run neighborhood attention with a kernel size equal to the tile size (in tokens)---this keeps the train-test distribution shift small. However, we see a small degradation in performance as a one-time scaling cost; see~\cref{fig:supp_searchlight_scaling}.

Vernier-Prob draws one latent vector per member and the network turns it into a spatial noise field (\cref{supp:gen-ens}).
If a large domain were covered with tiles, each tile would be a separate forward pass: the same latent would be shaped by a different tile context in each pass, the noise fields would not agree at tile edges, and the member would be a patchwork rather than a sample.
Searchlight runs the whole domain in one forward pass per member, so every member is one noise field from one latent, coordinated across the domain.
Seamless output was a side effect: the deterministic model gets the same treatment, and no tile edge exists in the output and no retraining is required.

\paragraph{Windowed attention.}
Vernier's backbone is a joint-attention transformer (\cref{supp:architecture}): the tokens of every stream (the prior stream, whose tokens feed the decoder, the terrain, the surface embedding and the registered HRRR cells) attend to each other in one sequence. 
For an $N\times N$ pixel domain we build the same network with every input at the domain size, so each stream's tokens lie at their true positions, and we restrict attention with a mask: a query attends to the keys of each stream that fall within a window the size of a training tile centered on the query, that is, within $\pm R_s$ along each axis, where $2R_s$ is the number of that stream's tokens across a tile times their spacing ($18$ tokens of $8$ pixels for the $100\,\mathrm{m}$ streams, $5$ cells for HRRR).
The window slides with the query and shifts inward at the domain edge instead of shrinking, so every token sees exactly a tile's worth of context, as in training, while neighboring tokens see overlapping windows and no boundary is ever drawn.
This is neighborhood attention~\cite{hassani2023neighborhood} applied at inference only, over all streams at once; \textit{we therefore extend neighborhood attention to multi-stream transformers}.

The mask is sparse: tokens are permuted into $64$-pixel spatial groups and the window becomes a block mask for flex attention, so key blocks outside the window are never computed.
On a $1000\,\mathrm{km}$ domain ($10{,}008$ pixels a side, $4.9$ million tokens) only $0.09\%$ of the key blocks are visited.
Positions enter through the physical rotary embedding with the true token-center coordinates, so a training-size domain is the training geometry up to a translation; a training-size domain run through Searchlight reproduces the ordinary tile forward to $10^{-4}$.
Conditioning is unchanged: the global AdaLN vector (time, lead, member latent) is one per forward, and the spatial AdaLN of the HRRR-driven part of Vernier-Prob reads the domain's own HRRR embedding token by token.

\paragraph{The physics prior is composited rather than solved on the domain.}
The physics prior (\cref{supp:prior}) could be solved over the whole domain in one mass-consistent problem, and we tried it.
The result was a large loss of skill (\cref{tab:supp_helene_prior}): on a $1000\,\mathrm{km}$ domain during Hurricane Helene (2024-09-27 04\,UTC, $1{,}419$ stations), Vernier-Det's wind-speed MAE rose from $1.96$ to $2.17\,\mathrm{m\,s^{-1}}$, its $u_{10}$ MAE from $1.35$ to $2.81\,\mathrm{m\,s^{-1}}$, and its station correlation fell from $0.79$ to $0.57$; an intermediate solve on $504\,\mathrm{km}$ tiles landed in between ($u_{10}$ MAE $1.64\,\mathrm{m\,s^{-1}}$).
We attribute this to a train-test shift: the network learned to correct priors solved on tile-sized problems, whose large-scale flow is HRRR's, and a prior solved over hundreds of kilometers couples the whole domain and no longer looks like what the network saw.
Searchlight therefore solves the prior on $24\,\mathrm{km}$ tiles (the training tile rounded up to whole HRRR cells and patches) at a $12\,\mathrm{km}$ stride and composites them by center crop: every pixel takes the solve it is most interior to, overlaps are cut at the midpoint between the two solve centers, and nothing is blended.
The composited prior enters the Searchlight forward as the one domain-wide base field, and the network's correction, computed in a single pass, runs continuously across the prior's cuts.
With this prior fixed, the whole domain can be inferred at once.
\begin{table}[t]
  \centering
  \footnotesize
  \setlength{\tabcolsep}{5pt}
  \begin{tabular}{@{}l rrr@{}}
    \toprule
     & \multicolumn{3}{c}{Prior solved on} \\
    \cmidrule(l){2-4}
     & 24\,km tiles & 504\,km tiles & Whole domain \\
    \midrule
    \multicolumn{4}{@{}l}{\emph{MAE}} \\
    Wind speed (m\,s$^{-1}$) & \textbf{1.96} & 2.04 & 2.17 \\
    $u_{10}$ (m\,s$^{-1}$) & \textbf{1.35} & 1.64 & 2.81 \\
    $v_{10}$ (m\,s$^{-1}$) & \textbf{1.72} & 1.89 & 2.37 \\
    $T$ ($^\circ$C) & 0.78 & 0.77 & \textbf{0.72} \\
    $T_d$ ($^\circ$C) & \textbf{0.74} & 0.76 & 0.80 \\
    \midrule
    \multicolumn{4}{@{}l}{\emph{Station correlation}} \\
    Wind speed & \textbf{0.79} & 0.66 & 0.57 \\
    $u_{10}$ & \textbf{0.83} & 0.69 & 0.30 \\
    $v_{10}$ & \textbf{0.89} & 0.83 & 0.76 \\
    \bottomrule
  \end{tabular}
  \caption{\textbf{Solving the physics prior over a large domain.} Vernier-Det run with Searchlight over one $1000\,\mathrm{km}$ domain during Hurricane Helene (2024-09-27 04\,UTC, $1{,}419$ stations), with the physics prior solved on $24\,\mathrm{km}$ tiles and composited (as in Searchlight), on $504\,\mathrm{km}$ tiles, or over the whole domain in one solve. Same checkpoint and HRRR forecast in all three. The $T$ and $T_d$ correlations are $0.96$ in all three.}
  \label{tab:supp_helene_prior}
\end{table}

\paragraph{Cost.}
Memory grows with the number of tokens per forward.
On an NVIDIA B200 (179\,GiB), a $1000\,\mathrm{km}$ cell ($10{,}008$ pixels, $4.9$ million tokens, one member) runs with attention in bf16 at about $128\,\mathrm{GiB}$ of GPU memory in use, and $163\,\mathrm{GiB}$ with the attention in fp32; solving the prior over the whole domain instead of on tiles needs $169\,\mathrm{GiB}$.
One such cell-hour takes about $7$ minutes, $2{,}304$ prior tiles included.
On the test-tile protocol (K=16 members per forward, fp32, \cref{fig:supp_searchlight_scaling}) the peak allocated memory is $9.6$, $23.3$ and $64.6\,\mathrm{GiB}$ for $43$, $72$ and $130\,\mathrm{km}$ domains, and one forward takes $1.4\,\mathrm{s}$ (H100, $43\,\mathrm{km}$), $3.7$ and $12.9\,\mathrm{s}$ (B200, $72$ and $130\,\mathrm{km}$).
Against the native tile the $43\,\mathrm{km}$ domain costs $9\times$ more per draw.
The $100.8\,\mathrm{km}$ domains of \cref{fig:supp_searchlight_det} ($1{,}008$ pixels, $50{,}078$ tokens) run on one NVIDIA L40S: the deterministic Searchlight forward takes $14$--$20\,\mathrm{s}$ at a peak of $3.7\,\mathrm{GiB}$, the $64$ prior tiles about $3$ minutes, and the tile composites $169$ forwards.

\paragraph{Examples.}
\Cref{sec:searchlight_results} compares Searchlight with tile composites and shows its ensemble members on two $100.8\,\mathrm{km}$ domains (\cref{fig:supp_searchlight_det,fig:supp_searchlight_prob}).

\paragraph{Domain-size scaling.}
Does the size of the domain used during inference affect performance?
We inferred $5{,}000$ of the 50k test draws of \cref{tab:operational} natively or as the center of a $43$, $72$ or $130\,\mathrm{km}$ Searchlight domain and scored only the central tile (exactly as the test set), for both parts of the operational Vernier-Prob ensemble (\cref{fig:supp_searchlight_scaling}).
Both parts pay the same small, one-time cost: Searchlight narrows the spread by about $5\%$ and raises the fCRPS by about $1\%$, mostly for $T$ and $T_d$, while $R_{\mathrm{pair}}$ and the bias of the ensemble mean barely change, and the margin over Vernier-Det stays large.
The cost appears entirely between the native tile and the first Searchlight domain; context beyond $43\,\mathrm{km}$ changes nothing measurable.

The cost is therefore not caused by the weather-dependent modulation: both forms of noise give a slightly narrower ensemble when the tile is inferred inside a larger domain. Our hypothesis as to the causation is that even though neighborhood attention fixes the field of view of tokens as if they were tile-limited, during full-domain forward passes this field of view grows with every block ($8\times$ in all) as tokens exchange information beyond their field of view. This might drive a small train-test distribution shift, causing the modest drop in performance, as a one-time bump. It is unclear why this affects SSR more than other metrics; this requires more investigation.

\ifnum\suppreflist=1
\clearpage%
\section*{Supplementary references}
{\small
\begin{list}{[S\arabic{suppref}]}{\usecounter{suppref}\setlength{\labelwidth}{2.2em}\setlength{\leftmargin}{2.7em}\setlength{\labelsep}{0.5em}\setlength{\itemsep}{1pt}\setlength{\parsep}{0pt}}
\item\label{suppref:bartal2023multidiffusion} Omer Bar-Tal, Lior Yariv, Yaron Lipman, and Tali Dekel. MultiDiffusion: Fusing diffusion paths for controlled image generation. In \emph{Proceedings of the 40th International Conference on Machine Learning}, pages 1737--1752. PMLR, 2023.
\item\label{suppref:bazlen2026windbench} Kyla Bazlen, Grant Buster, Brandon Benton, Lauren North, Ansley Baring, David~D. Turner, Emily Wells, and Laura Vimmerstedt. WIND-Bench: A benchmark dataset for in-situ near-surface wind speed observations across the conterminous United States. \emph{arXiv preprint arXiv:2609.12228}, 2026.
\item\label{suppref:demuzere2022lcz} Matthias Demuzere, Jonas Kittner, Alberto Martilli, Gerald Mills, Christian Moede, Iain~D. Stewart, Jasper van Vliet, and Benjamin Bechtel. A global map of local climate zones to support earth system modelling and urban-scale environmental science. \emph{Earth System Science Data}, 14(8):3835--3873, 2022.
\item\label{suppref:demuzere2023cglcmodislcz} Matthias Demuzere, Cenlin He, Alberto Martilli, and Andrea Zonato. A hybrid 100-m global land cover dataset with Local Climate Zones for WRF. Zenodo, version 1.0.0, 2023. Dataset, \url{https://doi.org/10.5281/zenodo.7670653}.
\item\label{suppref:fovell2020hrrr} Robert~G. Fovell and Alex Gallagher. Boundary layer and surface verification of the High-Resolution Rapid Refresh, version 3. \emph{Weather and Forecasting}, 35(6):2255--2278, 2020.
\item\label{suppref:fovell2022hrrrwind} Robert~G. Fovell and Alex Gallagher. An evaluation of surface wind and gust forecasts from the High-Resolution Rapid Refresh model. \emph{Weather and Forecasting}, 37(6):1049--1068, 2022.
\item\label{suppref:karras2022edm} Tero Karras, Miika Aittala, Timo Aila, and Samuli Laine. Elucidating the design space of diffusion-based generative models. In \emph{Advances in Neural Information Processing Systems}, pages 26565--26577, 2022.
\item\label{suppref:lipman2023flowmatching} Yaron Lipman, Ricky T.~Q. Chen, Heli Ben-Hamu, Maximilian Nickel, and Matt Le. Flow matching for generative modeling. In \emph{International Conference on Learning Representations}, 2023.
\item\label{suppref:thorarinsdottir2010wind} Thordis~L. Thorarinsdottir and Tilmann Gneiting. Probabilistic forecasts of wind speed: Ensemble model output statistics by using heteroscedastic censored regression. \emph{Journal of the Royal Statistical Society: Series A (Statistics in Society)}, 173(2):371--388, 2010.
\end{list}
}
\fi

\end{document}